\documentclass[a4paper,fleqn]{cas-dc}

\newif\ifblindreview
\blindreviewfalse
\newif\ifpublicpreprint
\publicpreprinttrue

\ifdefined\pdftrailerid
  \pdftrailerid{AIJIM-CSI-1786615200-\jobname}
\fi

\usepackage[utf8]{inputenc}
\usepackage{microtype}
\usepackage{hyphenat}
\usepackage{needspace}
\usepackage{mathtools}
\usepackage[inline]{enumitem}
\usepackage{tabularx}
\usepackage{ragged2e}
\usepackage{xurl}
\usepackage{seqsplit}
\usepackage{placeins}
\usepackage[capitalise,nameinlink]{cleveref}
\usepackage{bookmark}
\usepackage[numbers,sort&compress]{natbib}

\setlist{noitemsep,topsep=2pt,parsep=0pt,partopsep=0pt}
\makeatletter
\def\input@path{{./}{./tex/}{./tex/tables/}{./}}
\makeatother
\graphicspath{{./}}

\definecolor{rowlight}{gray}{0.96}
\newcolumntype{Y}{>{\RaggedRight\arraybackslash\hspace{0pt}}X}
\newcolumntype{P}[1]{>{\RaggedRight\arraybackslash\hspace{0pt}}p{#1}}

\providecolor{rowlight}{gray}{0.96}

\makeatletter
\@ifundefined{NC@find@P}{\newcolumntype{P}[1]{>{\RaggedRight\arraybackslash\hspace{0pt}}p{#1}}}{}\@ifundefined{NC@find@T}{\newcolumntype{T}[1]{>{\ttfamily\small\RaggedRight\arraybackslash\hspace{0pt}}p{#1}}}{}\makeatother

\providecommand{\TableSmallZebra}{\begingroup
  \footnotesize
  \renewcommand{\arraystretch}{1.14}\setlength{\tabcolsep}{3.2pt}}
\providecommand{\EndTableSmallZebra}{\endgroup}

\makeatletter
\@ifundefined{TableSmallZebraEnv}{}{}\makeatother

\newenvironment{SpecHeroTable}[1][t]{\begin{table*}[#1]
    \captionsetup{width=\textwidth,position=top}\centering
    \footnotesize
    \renewcommand{\arraystretch}{1.14}\setlength{\tabcolsep}{3.2pt}}{\end{table*}}
 
\ExplSyntaxOn
\RenewDocumentCommand \printorcid { } { }

\cs_set:Npn \__first_footerline:
{
  \group_begin:
  \small\sffamily
  \ifnum\theblind>0\relax
  \else
    \__short_authors: :~
  \fi
  { \rmfamily\itshape Preprint~---~not~peer-reviewed }
  \group_end:
}

\RenewDocumentEnvironment { Abstract } { o }
{
  \group_begin:
  \IfNoValueTF { #1 } { }
  { \tex_gdef:D \abstractname { #1 } }
  \parindent 0pt
  \box_if_empty:NTF \g_stm_key_box
  { \leftskip = .35 \textwidth }
  {
    \dim_gset:Nn \l_tmpa_dim { \box_ht:N \g_stm_key_box }
    \dim_gadd:Nn \l_tmpa_dim { \box_dp:N \g_stm_key_box }
    \leftskip .35\textwidth
    \hspace*{-.35 \textwidth }
    \noindent\rlap{\box_use_drop:N \g_stm_key_box}
    \skip_vertical:n { - \l_tmpa_dim }
  }
  \noindent \abstractname \par
  \skip_vertical:n { -4pt}
  \noindent \rule{.65\textwidth}{.2pt}\par \footnotesize
  \ignorespaces \everypar { \parindent=1.5em }
}
{ \par \group_end: }
\ExplSyntaxOff

\hypersetup{
  colorlinks=true,
  linkcolor=black,
  citecolor=black,
  urlcolor=black,
  pdfstartview=FitH,
  pdfpagemode=UseNone
}
\providecommand{\doi}[1]{\href{https://doi.org/#1}{doi:\,\nolinkurl{#1}}}
\providecommand{\arxiv}[1]{\href{https://arxiv.org/abs/#1}{arXiv:\,\nolinkurl{#1}}}

\crefname{figure}{Fig.}{Figs.}
\Crefname{figure}{Figure}{Figures}
\crefname{table}{Table}{Tables}
\Crefname{table}{Table}{Tables}
\crefname{section}{Section}{Sections}
\Crefname{section}{Section}{Sections}

\providecommand{\captionsetup}[1]{}
\providecommand{\Description}[1]{}

\newenvironment{HeroTable*}[1][]{\begin{table*}[t]
  \centering
}{\end{table*}}

\newcommand{\AIJIM}{\mbox{AIJIM}}
\newcommand{\AIJIMLongForm}{Artificial Intelligence Justification and Integrity Model}

\newcommand{\FlowAEvidenceDOI}{10.5281/zenodo.21339360}
\newcommand{\FlowAVerifierDOI}{10.5281/zenodo.21329825}
\newcommand{\FlowAReviewerInputDOI}{10.5281/zenodo.21333201}

\newcommand{\PaperTitle}{Making AI-Assisted Claims Independently Challengeable:
  Publication Authority and a Protocol for Falsifiable Publication Records}
\newcommand{\PaperShortTitle}{Publication Authority for AI-Assisted Claims}
\newcommand{\PaperRunAuthor}{Tiltack et al.}
\newcommand{\PaperPDFAuthors}{Torsten Olivi Tiltack; Yifei Dong; Kun Yu; Xu Wang; Wei Liu; Jianlong Zhou; Ren Ping Liu; Fang Chen}
\newcommand{\PaperPDFSubject}{A Design Science Research article presenting and evaluating publication-accountability design knowledge and PAC-2026 SF-4}

\newcommand{\ApplyPaperPDFMetadata}[1]{\ifblindreview
    \hypersetup{pdftitle={#1},pdfauthor={Anonymous author(s)},pdfsubject={\PaperPDFSubject}}\else
    \hypersetup{pdftitle={#1},pdfauthor={\PaperPDFAuthors},pdfsubject={\PaperPDFSubject}}\fi
}
 
\begin{document}
\let\WriteBookmarks\relax
\def\floatpagepagefraction{1}
\def\textpagefraction{.001}

\shorttitle{\PaperShortTitle}
\shortauthors{\PaperRunAuthor}
\title[mode=title]{\PaperTitle}

\ifblindreview
  \author[1]{Anonymous author(s)}
\else
  \author[1]{Torsten Olivi Tiltack}
  \author[1]{Yifei Dong}
  \cormark[1]
  \author[1]{Kun Yu}
  \author[2]{Xu Wang}
  \author[3]{Wei Liu}
  \author[1]{Jianlong Zhou}
  \author[2]{Ren Ping Liu}
  \author[1]{Fang Chen}
\fi

\affiliation[1]{organization={Data Science Institute, University of Technology Sydney},
  city={Ultimo},
  state={NSW},
  country={Australia}}
\affiliation[2]{organization={Global Big Data Technologies Centre, University of Technology Sydney},
  city={Ultimo},
  state={NSW},
  country={Australia}}
\affiliation[3]{organization={School of Computer Science, University of Technology Sydney},
  city={Ultimo},
  state={NSW},
  country={Australia}}

\ifblindreview\else
  \cortext[cor1]{Corresponding author: Yifei.Dong@uts.edu.au}
\fi

\begin{abstract}
AI-assisted claims can appear authoritative when evidence, analysis,
human authorization, presentation, and correction history refer
to different states. We develop Publication
Authority as an exact-state, non-transferable, single-use publication
capability, instantiated in the Publication-Accountability Calculus
(PAC-2026), a candidate within the Artificial Intelligence Justification and
Integrity Model (AIJIM) Protocol. Its fourth bounded semantic freeze (SF-4) specifies six
obligations governing evidence, runs and artifacts, measurement disclosure,
authorization, surface correspondence, and lifecycle continuity. Each yields
a target-bound witness, localized counterexample, or localized
unverifiability; none can compensate for another. Only a
fresh, complete all-pass record derives the permit consumed by one atomic
publication transition.

Ten models explored 110,764 distinct safe reachable states; 76 unsafe
configurations produced the expected violation or observer countermodel. A
surface passing its own correspondence check cannot authorize
publication unless the accepted record admits it. SF-4
separates evidence horizon from verification time and rejects an authentic but
causally invalid authorization.
Instance-blind G7-B matched 183 scored expectations in-house after shared
integration.

Three author-led native comparisons identify protection gaps in a constructed
in-toto/OPA profile: measurement-state admission, proof-expiry evaluation,
and required-run identity. Repairs restore parity on the tested
cases. An exploratory extension constructs exact claim-edit regions under
existing e-closure semantics: a minimum retained core qualifies every
permitted selection; for the full hypothesis universe (\(U=H\)), a quadratic
frontier computes all strongest-core sizes. Enumeration and native controls distinguish reusable statistical
qualification from the fresh publication authority each candidate requires.
These results connect exact-state semantics, reusable qualification, and
transferable protection rules; independent conformance and field evaluation
remain open.
\end{abstract}
 \begin{keywords}
AI-assisted claims \sep publication accountability \sep Publication Authority
\sep protocol conformance \sep falsifiability \sep design science research
\end{keywords}
 
\maketitle
\ApplyPaperPDFMetadata{\PaperTitle}

\section{Introduction}\label{sec:introduction}

AI-assisted publication creates a coordination problem that is not reducible
to model accuracy. Consider a claim approved as version \(c_1\). A
meaning-bearing edit creates version \(c_2\), but the evidence, analytical run,
and human approval still refer to \(c_1\), while the reader-facing page omits a
limitation. Every retained item may be authentic, yet the items no longer
describe the same state. The resulting claim can appear
more strongly supported, authorized, measured, current, or reviewable than the
joint record warrants. We call this failure \emph{silent epistemic promotion}.
``Stronger'' denotes the status asserted by the publication record; it is not
a numerical truth scale and does not mean that the factual subject matter is
true or scientifically falsifiable.

The component problems are independently established. Long-form generated
text can contain support differences at the level of individual claims, and
the presence of citations is distinct from citation completeness and support
\citep{min2023factscore,gao2023citations}. Computational and reporting details
can materially affect reproducibility and even substantive conclusions
\citep{pineau2020reproducibility,kapoor2023leakage}. Accountability research
likewise motivates explicit rules, inspectable use, and later verification
rather than reliance on access control or opaque assurance alone
\citep{weitzner2008information,kroll2017accountable}. These studies do not
demonstrate the prevalence of the complete multi-state failure in publishing.
They establish why keeping the constituent relations recoverable is a
consequential design problem.

Here, \emph{publication} denotes a governed transition rather than a document
genre: one concrete, versioned candidate is intentionally made available to a
declared audience under a claimed recorded status. The audience may be public
or access-controlled, and the candidate may be a scientific paper, news
report, dataset description, or another claim-bearing object.

AI-assisted publication motivates the design. The admitted profile and its
bound policy determine which model or tool
contributions are material; examples include claim wording, evidence
synthesis, analysis code, figures, tables, or translation. The unit of
authorization is the complete
candidate placed before the human authorizer. Claim-level evidence relations
remain possible, but the protocol does not require a separate permit for every
sentence.
The laws are therefore not logically AI-exclusive, although this article does
not evaluate a non-AI domain. Here, ``independently challengeable'' means that a
relying verifier can reconstruct and contest the recorded basis and later
status without an unrecorded producer-side join. It does not mean that
independent specification-only re-implementation and conformance (G7), general
interoperability, or field utility has already been demonstrated.

Existing standards solve much of the mechanism layer. Provenance models,
research-object packages, signed attestations, content credentials,
transparency services, receipts, and update notices can record and transport
relevant facts. The unresolved scientific question is narrower: what
publication-specific semantics should govern the transition into a stronger
recorded state when those mechanisms must refer to the same exact target and
success on one dimension must not compensate for failure on another? An
existing profile or defined composition that already supplies equivalent
semantics is a direct falsifier of the residual contribution.

We address that question through a Design Science Research (DSR) artifact
\citep{hevner2004design}. The
proposed design knowledge is \emph{Publication Authority}: a
protocol-derived, non-transferable capability to perform one atomic
publication transition for one exact candidate and context.

In plain terms, Publication Authority answers one question: may this exact
version---rather than an earlier or materially different one---be published
under the claimed recorded status now?

\begin{quote}
\textbf{Definition 1 (Publication Authority).}
For a fixed admitted profile, Publication Authority exists only when every
mandatory publication obligation has a fresh, exact-target passing result and
promotion witness. A valid, unconsumed \texttt{PublicationPermit} is its
per-permit single-use executable representation: one permit can authorize at
most one publication act. An approval, signature, receipt,
provenance trace, or passing result on any proper subset is not Publication
Authority.
\end{quote}

This is a protocol term, not legal, editorial, institutional, or standards
authority. A material non-equivalent change does not inherit it. A valid
negative verification record remains portable evidence, but cannot authorize
publication.

We call the versioned, machine-readable artifact introduced here the
\emph{Publication-Accountability Calculus} (PAC-2026), a candidate within the
\AIJIMLongForm{} (\AIJIM{}) Protocol. The primary protocol object evaluated here
is its fourth bounded semantic freeze (SF-4), which we refer to as PAC-2026
SF-4. The year identifies the candidate series; SF-4 denotes its fourth
preserved research freeze. PAC-2026 is a research specification candidate,
not a published standard. Its fixed Core Profile---the bundle of rules admitted for
this evaluation---closes six coordinates: evidence; runs and artifacts;
measurement disclosure; exact human authorization; registered reader-surface
correspondence; and lifecycle continuity. Every applicable obligation returns
a target-bound witness, a localized counterexample, or localized
unverifiability. The latter two fail closed, and no coordinate can promote
another. ``Substrate-neutral'' describes the design objective of separating
these semantics from databases, envelopes, repositories, and transparency
services; binding-specific evidence is reported in \Cref{sec:evaluation}.

The study has one research question (RQ) and two subquestions (SQ1--SQ2):

\begin{description}
  \item[RQ:] How can a fixed-profile publication-accountability semantics
  govern the transition of an AI-assisted claim into a stronger recorded
  publication state so that evidence, analytical work, authorization,
  reader-visible meaning, and later status changes remain bound to the exact
  state being published?

  \item[SQ1:] Which literature- and standards-grounded failure classes and
  design requirements define that transition within the declared profile?

  \item[SQ2:] To what extent does PAC-2026 SF-4 satisfy the stated criteria
  for semantic coherence, bounded safety, targeted fault sensitivity, and
  constructibility within the declared evaluation scope?
\end{description}

The paper makes three ordered contributions.

\begin{enumerate}[label=(C\arabic*)]
  \item \textbf{Publication Authority and non-compensating design laws.}
  We frame exact-state authority as the theoretical object and state five
  falsifiable laws for its derivation and preservation: Accountability
  Closure, Authorization Freshness, Epistemic Non-Amplification, Semantic
  Surface Preservation, and Continuing Challengeability.

  \item \textbf{PAC-2026 SF-4 as an executable specification artifact.}
  The fixed profile turns those laws into six obligations, separate result
  maps, canonical identities, a temporal authorization-to-permit chain, an
  at-most-once permit, registered surfaces, and authenticated lifecycle events.
  Here, ``executable'' denotes the machine-checkable specification corpus and
  its validation and model artifacts, not an end-to-end SF-4 implementation.

  \item \textbf{Criterion-led bounded evaluation.}
  Executable controls distinguish locally valid surfaces from authorized
  publication configurations, witnessed absence from performed measurement,
  and authentic authorization from valid temporal ordering. Finite-state
  models and separately identified implementation layers test the four
  evaluation criteria.
\end{enumerate}

Targeted native comparisons isolate three protections supplied by
the archived binding but missing from a constructed in-toto/OPA profile:
state admission, proof-expiry evaluation, and required-run identity.
Targeted repairs restore parity and identify the integration rules that
produce these protections
(\Cref{ssec:native-protection-comparisons}).
An exploratory follow-on contributes an exact construction for reusing
statistical assurance across claim selections. It determines a minimum
retained core whose permitted edit region remains qualified under a fixed
evidence snapshot. When the upper set is the full hypothesis universe
(\(U=H\)), it computes the complete strongest-core frontier in quadratic
arithmetic work. Mathematical derivation, exact enumeration, and
native controls distinguish reusable analytical qualification from the fresh
Publication Authority still required for each candidate
(\Cref{sec:edit-regions}). This adds a computational result without changing
the SF-4 specification or its original evaluation.

The separately versioned PAC corpus defines the executable semantics; this
article develops their scientific rationale, abstraction, and evaluation.
Historical and later implementation paths supply separately identified
constructibility evidence while preserving the frozen SF-4 study baseline.
The compact
claim--evidence--falsifier ledger is introduced here and reported in
\Cref{tab:claim-evidence-ledger}; it makes each principal conclusion and its
defeater inspectable without turning the ledger into a substitute for the
argument.

\Cref{sec:literature} establishes the literature and standards context.
\Cref{sec:method} derives the bounded requirements and evaluation criteria.
\Cref{sec:design} presents the design knowledge and PAC instantiation.
\Cref{sec:evaluation} reports results by criterion.
\Cref{sec:edit-regions} presents the exploratory edit-region construction,
followed by interpretation
and limits in \Cref{sec:discussion} and the conclusion in
\Cref{sec:conclusion}.

\section{Literature and Standards Context}\label{sec:literature}

Publication accountability builds on established methods for provenance,
evidence binding, authorization, and verification. The design problem is to
specify their dependencies when one concrete candidate acquires a recorded
publication status. The comparison below identifies what existing standards
and systems already provide and how PAC-2026 couples those capabilities at
the publication boundary.

\subsection{Evidence, computation, and accountable claims}

Claim-level evaluation shows why a document-level score is insufficient:
support can differ among atomic factual units, and a citation may be present
without adequately or completely supporting the associated assertion
\citep{min2023factscore,gao2023citations}. Reproducibility work similarly
shows that code, data, execution conditions, and reporting choices can be
material to a computational claim \citep{pineau2020reproducibility,
kapoor2023leakage}. These findings support explicit evidence and run closure
for claims to which those dependencies apply. They do not imply that every
claim requires the same artifacts or that a protocol can determine factual
truth.

Information-accountability scholarship adds the temporal and procedural
dimension. Inspectable rules, ex ante commitments, and ex post verification
can make consequential information use challengeable even when full internal
disclosure is impossible \citep{weitzner2008information,
kroll2017accountable}. This supports challengeability as a design objective,
but does not validate PAC-2026 or prescribe its fields.

PAC can consequently be understood as one technical control within broader AI
risk governance, not as an AI risk-management framework. The NIST AI Risk
Management Framework organizes organizational activity across Govern, Map,
Measure, and Manage and treats context and lifecycle responsibilities as
material \citep{nist2023airmf}. PAC addresses the publication transition by
making its decision basis inspectable. Its records can supply structured,
authenticated decision evidence to risk-governance processes; risk
identification, likelihood and impact assessment, risk tolerance, and the
adequacy of organizational governance remain outside this control.

\subsection{Strong adjacent standards and systems}

\paragraph{Provenance and research objects.}
The World Wide Web Consortium's PROV model represents entities, activities,
agents, derivations, bundles, and
temporal constraints; OpenLineage represents process lineage; RO-Crate
packages research objects; and nanopublications bind atomic assertions to
provenance and publication information
\citep{w3c2013provdm,w3c2013provconstraints,openlineage2026spec,
soilandreyes2022rocrate,kuhn2016nanopublications}. Together they provide
established ways to represent dependencies, preserve claim identity, and make
research objects retrievable beyond the originating system.

\paragraph{Attestation, authenticity, and transparency.}
in-toto and Supply-chain Levels for Software Artifacts (SLSA) record and
verify declared software-supply-chain steps; World Wide Web Consortium (W3C)
Verifiable Credentials (VC) represent signed claims; and the Coalition for
Content Provenance and Authenticity (C2PA) specification, version 2.4,
provides signed manifests, hard and soft bindings, AI-related assertions,
document embedding, repository receipts, and version-history data
\citep{torresarias2019intoto,slsa2025v12,w3c2025vcdm,c2pa2026spec}.
The Remote Attestation Procedures (RATS) architecture separates Attester,
Verifier, Relying Party, Evidence, policy, and Attestation Results
\citep{birkholz2023rats}. These approaches establish precedents for signing,
integrity, freshness, role separation, and origin-independent checking.

Supply Chain Integrity, Transparency, and Trust (SCITT) is the closest
standards substrate. RFC~9943 defines registration-policy checks and
append-only transparency for signed statements; RFC~9942 defines portable
CBOR Object Signing and Encryption (COSE) receipts
\citep{birkholz2026rfc9943,steele2026rfc9942}. SCITT is deliberately
content-agnostic and leaves statement
management and application meaning to profiles and relying systems. It can
transport parts of a publication record, but its general architecture does not
by itself determine whether evidence is sufficient, an approval still targets
the current wording, a registered surface preserves the authorized state, or
a challenge has changed current status. Conversely, PAC-2026 does not replace
SCITT's transparency service or require it as a substrate.

\paragraph{Authorization, assurance, and lifecycle.}
Proof-carrying authorization attaches checkable evidence to a request, while
runtime verification distinguishes satisfaction, violation, and inconclusive
observation \citep{appel1999pca,bauer2001pcasystem,bauer2011runtime}.
The Structured Assurance Case Metamodel (SACM), Resolute, and the Open Security
Controls Assessment Language (OSCAL) provide structured assurance and
assessment artifacts
\citep{omg2023sacm,gacek2014resolute,nist2025oscalassessment}.
Registered Reports establish precommitment, and Crossmark links corrections,
retractions, withdrawals, and other updates
\citep{chambers2022registered,crossref2026crossmarkupdates}. Type-and-effect,
typestate, substructural, refinement, and proof-carrying-code traditions supply
the formal vocabulary used below \citep{lucassen1988effects,
strom1986typestate,walker2005substructural,abadi1991refinement,
necula1997pcc}. PAC applies this vocabulary to the derivation and use of
publication authority.

Scholarly-agent protocols are closer comparators than generic provenance
alone. Agent-Native Research Artifacts (ARA) connect claims, execution, and
exploration histories to a staged verification and human-review process;
Knows represents claims, evidence, review relations, and version history in
structured research records \citep{liu2026ara,yu2026knows}. The Agentic
Publication Protocol (APP) explicitly binds a publication release to a
repository commit, tree, validation result, and human approval
\citep{lu2026app}. ScientistOne's Chain-of-Evidence checks claims against
papers, code, and execution records; the Hypothesis Evolution Protocol (HEP)
makes evidence-linked belief updates and hypothesis lifecycles explicit
\citep{meng2026scientistone,takahara2026hep}. Evidence-Graded Decision
Authorization (EGDA) restricts clinical assertions according to claim-specific
evidence thresholds, including qualification or refusal
\citep{lin2026egda}. Together, these works already specify research-object
verification, evidence-gated claims, and versioned human-approved publication.
The question for PAC is how those capabilities are coupled to the complete
reader-facing transition, not whether those components exist.

Recent work brings evidence and authorization together. The Calibration Turn formalizes
evidence-licensed assertion; the AI Deployment Authorisation proposal (ADAS)
couples multidimensional evidence to a deployment authorization certificate;
and current
agent-accountability drafts
bind authorization, receipts, protected objects, and presentation evidence to
high-risk actions \citep{li2026calibration,saparning2026adas,
schrock2026authorizationreceipts}. They establish evidence-bound gating,
multidimensional non-compensation, revocation, or exact-action display evidence
in their respective domains. Their units of analysis differ from the complete
reader-facing publication transition evaluated here. That distinction is a
bounded comparison, not proof of non-substitution.

Agent-security proposals address several of the same mechanisms.
Context-to-Execution Integrity (CXI) and CapLease bind exact actions to
consumable capabilities; the Action Evidence Boundary (AEB) Internet-Draft and
AID-Guard add evidence checks, commit-time revalidation, uncertain outcomes,
and recovery rules
\citep{santosgrueiro2026cxi,xu2026caplease,
schrock2026actionevidenceboundary,tong2026aidguard}. These proposals inform the
comparison through exact-target authorization, consumable authority,
freshness, and recovery-safe uniqueness. In the inspected versions, the
governed object is an agent action or provider effect rather than the complete
reader-facing publication transition. Whether these mechanisms can be
composed into an equivalent publication profile remains open.

A later targeted source audit identified four relevant preprints that had
already been submitted before the frozen comparison closed but were absent
from its corpus. Together, Auditable Autonomous Research (AAR), Pramana,
Traxia, and Proof-or-Stop cover claim-level auditability, conflict-preserving
reporting, three-valued or offline verification, human publication review and
living status, and exact-state conjunctive gates
\citep{rasheed2026fluent,kadaboina2026pramana,dogah2026traxia,
huang2026prooforstop}. Their governed objects differ: research synthesis,
agent claims, living publication, and
software-merge readiness. No single inspected paper specifies the complete
reader-facing publication transition claimed here.

The revision-stage audit also identified SIP-RC, published on 14 July 2026.
This candidate release profile defines artifact relationships, hard gates,
hash-bound human approval, view agreement, and atomic publication
\citep{liu2026siprc}. Its relational model includes a conditional conformance
guarantee. It reports executed checks of selected mechanisms, conformance
vectors, and execution logs; the planned comparison of the complete profile
has not yet been performed. Its three case systems share an author and are
not independent replications. Watts's individual Internet-Draft of
1 September 2026, already retained in the follow-on comparison's source
record, specifies registered predicates, append-only events, and claim
versions for research provenance, without assigning decision authority to
automated systems \citep{watts2026admissibility}. It postdates the historical
27 July comparison cut-off but predates this article's submission.
ClaimReceipt, submitted on 2 September, defines claim-relative evidence
sufficiency with a first-class inconclusive outcome
\citep{zhu2026claimreceipt}. The SATO composition draft of 16 August binds
human authorization to an exact action, requires current appraisal before
an effect, and preserves indeterminate checks separately from passed and
failed checks. Its vector-freeze rule requires agreement between
implementations maintained by different parties; the reported cross-slot
pack remains a candidate awaiting that condition \citep{mih2026sato}.
\Cref{tab:residual-falsifier} compares the documented properties of these
profiles; it is not an execution or a performance ranking.

LEDGER appeared after the comparison cut-off and belongs only to the living
comparison. It turns observed agent sessions into layered claim-to-evidence
trace graphs for human review while preserving the underlying records and
explicitly treating the inferred graph as an audit aid rather than a source of
truth \citep{kim2026ledger}. This is a strong upstream complement for
producing inspectable evidence and run records, not a publication-admission
semantics. The living comparison also records a rapid convergence of
agent-effect proposals on exact-state, consumable authority: single-use effect
permits consumed at effect linearization \citep{zhu2026runtimeauth},
parameter-bound grants with limited uses \citep{zhang2026agate}, and a
passkey-signed canonical effect digest with trusted rendering whose binding
and single-use properties are machine-checked in Tamarin
\citep{kodathala2026approve}. The first of these was submitted two days before
this article. PAC therefore claims no priority for single-use or
exact-state authority as mechanisms; it claims their instantiation for the
publication act together with the obligations below. A concurrent empirical
study of publication enforcement reports that a production response-act
checker accepted 291 of 302 unsupported-labelled answers; a direct-grounding
baseline accepted 41, with lower retention of supported answers
\citep{alpay2026approval}. This separates the quality of a semantic decision
from the integrity of its use at publication: exact binding preserves an
approval's association with content, including errors in that approval.
Both dimensions matter; this external study does not validate PAC-2026.

\begin{SpecHeroTable}[!b]
\caption{Living capability map, not the frozen substitution corpus. Existing
work already supplies every component class in the middle column. The final
column states the narrower publication-transition question posed here.}
\label{tab:standards}
\begin{tabularx}{\textwidth}{@{}P{0.20\textwidth}P{0.36\textwidth}Y@{}}
\toprule
\textbf{Capability family} & \textbf{What prior work already provides} &
\textbf{Remaining composition question} \\
\midrule
Provenance and research objects &
Claim-, entity-, activity-, and process-level lineage; portable research
packages; persistent assertion identity
\citep{w3c2013provdm,openlineage2026spec,soilandreyes2022rocrate,
kuhn2016nanopublications} &
Do the evidence, analytical runs, disclosures, intended surface, and
predecessor state all refer to one exact reader-facing candidate? \\

Authenticity, attestation, and transparency &
Signed manifests, software-supply-chain attestations, content bindings,
portable receipts, and append-only transparency
\citep{torresarias2019intoto,slsa2025v12,c2pa2026spec,
birkholz2026rfc9943,steele2026rfc9942} &
Can a valid signature, receipt, or attestation remain evidence for its own
coordinate without promoting the publication decision as a whole? \\

Authorization and effect control &
Evidence-carrying decisions, exact-action authorization, consumable authority,
durable replay protection, and indeterminate-outcome recovery
\citep{appel1999pca,santosgrueiro2026cxi,xu2026caplease,
schrock2026actionevidenceboundary,tong2026aidguard} &
Can these controls be composed with claim and run closure, the registered
reader surface, and the later publication lifecycle? \\

Scientific claim governance &
Evidence-licensed assertion, claim-level auditability, three-valued
verification, conflict preservation, and final human review
\citep{li2026calibration,rasheed2026fluent,kadaboina2026pramana,
dogah2026traxia,huang2026prooforstop} &
Does a complete fresh passing set authorize only this revision and this
publication transition, with honest absence retained? \\

Assurance, presentation, and lifecycle &
Structured assurance arguments, assessment records, reader-visible content
bindings, and correction or withdrawal notices
\citep{omg2023sacm,nist2025oscalassessment,c2pa2026spec,
crossref2026crossmarkupdates} &
Are the authorized reader observables and ordered postpublication changes
bound to the same candidate and evidence horizon? \\

Prospective composed rival &
Any source-documented profile or coherent composition of the capabilities
above; the unexecuted successor challenge permits newly written integration
rules &
The rival wins if it jointly supplies exact candidate identity, separate
mandatory outcomes, non-compensating promotion, fresh authorization, surface
correspondence, and lifecycle continuity. This is not the historical
27 July 2026 procedure, which withheld credit from new coordination rules. \\
\bottomrule
\end{tabularx}
\end{SpecHeroTable}

\begin{SpecHeroTable}[!t]
\caption{Selected publication-relevant properties in five inspected profiles.
Y: documented for the source's governed object; P: partial match;
N: not identified in the inspected passages. These entries describe
capabilities, not scores or evidence of inexpressibility. A substitute would
need to preserve the combined publication behavior, not PAC's data structures.
Inspection covers SIP-RC Sections III--VII, SATO Sections 4--8, Watts
Sections 4--7, ClaimReceipt Sections 2--3, and APP Sections 2.1--2.3.}
\label{tab:residual-falsifier}
\begin{tabularx}{\textwidth}{@{}P{0.21\textwidth}P{0.055\textwidth}P{0.055\textwidth}P{0.055\textwidth}P{0.065\textwidth}P{0.055\textwidth}Y@{}}
\toprule
\textbf{Residual property} &
\textbf{SIP-RC} \citep{liu2026siprc} &
\textbf{SATO} \citep{mih2026sato} &
\textbf{Watts} \citep{watts2026admissibility} &
\textbf{Claim\-Receipt} \citep{zhu2026claimreceipt} &
\textbf{APP} \citep{lu2026app} &
\textbf{Documented scope and distinction} \\
\midrule
One exact candidate admitted & Y & Y & P & P & Y &
The bound objects are a release graph, action, registered procedure, claim,
or repository release; their scopes differ. \\

Separate outcomes, including undetermined checks & P & Y & P & Y & N &
SIP-RC separates hard gates; Watts records binding-gate results. SATO and
ClaimReceipt explicitly preserve undetermined checks. \\

Authorization freshness on candidate change & Y & Y & N & N & P &
SIP-RC invalidates hash-mismatched approvals after mutation; SATO rejects
reuse under a changed action; APP records approval beside the tree hash. \\

Surface correspondence & Y & P & N & N & N &
SIP-RC checks views against authoritative artifacts and binds published
bytes. SATO binds action projections rather than a publication surface. \\

Non-compensation & Y & Y & P & P & N &
SIP-RC states the conjunction explicitly; Watts forbids passing while a
binding calibration or power gate fails. \\

Unverifiability distinct from violation & P & Y & N & Y & N &
SATO and ClaimReceipt distinguish undetermined results. SIP-RC represents
unknown nodes, but its release predicate is binary. \\

Lifecycle effects on related records & P & P & P & N & P &
SIP-RC distinguishes revoked from never-conforming releases; Watts derives
state from events. The inspected rules differ from PAC's challenge,
correction, supersession, withdrawal, and status-recovery coupling. \\
\bottomrule
\end{tabularx}
\end{SpecHeroTable}
 
\subsection{Publication-transition coordination}

The living comparison in \Cref{tab:standards} makes the coordination question
concrete. A
composed stack can represent evidence and claims, package research objects,
attest production, authenticate human decisions, register signed statements,
issue receipts, preserve status updates, and support external verification.
The residual question is whether the composition itself has an explicit
publication meaning: one exact candidate; separate mandatory outcomes;
non-compensating promotion; revision-exact authorization; correspondence with
the registered reader surface; and authenticated challenge, correction,
supersession, or withdrawal without rewriting the predecessor.
No item in that list is claimed novel in isolation. The candidate contribution
is their coupling at one publication transition, including honest absence, a
single-use permit, and distinct evidence-horizon and verification-instant
semantics. SATO and ClaimReceipt already document undetermined outcomes
separately from violations, so that distinction is not itself the residual
claim. PAC makes its publication-specific consequences executable: a
substantive unverifiable obligation blocks promotion; a witnessed
\texttt{NOT\_MEASURED} can satisfy measurement disclosure without asserting
a measurement; and authorization checks distinguish authentic-but-stale
approval, relabeled approval, and an unavailable relation. Missing required
disclosure structure is rejected at admission rather than automatically
becoming a substantive checker \texttt{FAIL}. The scientific comparison
concerns the behavior and coupling of these rules, not whether another
profile uses PAC's field names or architecture. The later WIMSE evaluation
draft also distinguishes an evaluation instant from fetched-input age and
validity horizons; these temporal roles differ from PAC's evidence-scope
horizon \citep{jackson2026wimse}.

The independent literature record supports two failure classes directly:
unbound or incomplete evidence, and missing or misrepresented computational
work. Stale authorization and reader-surface drift are analytically grounded
failure hypotheses whose prevalence and field effect remain open. Lifecycle
opacity is partly grounded by established update mechanisms, but PAC's exact
state vocabulary is a design choice under test. The six-coordinate coupling is
therefore neither asserted universally minimal nor empirically prevalent.

PAC-2026 proposes \emph{No Silent Epistemic Promotion} as one answer. A
pre-existing profile defeats the residual novelty claim if it preserves
equivalent candidate admission, result separation, authorization freshness,
surface correspondence, non-compensation, honest absence, and lifecycle
effects. A later equivalent profile can narrow necessity, minimality, or
distinctiveness claims without retrospectively deciding priority. The
historical rival procedure excluded newly written coordination rules and
thus did not fairly test composed substitutes. Four omissions above plus
SIP-RC predate its 27 July cut-off; Watts belongs to the later source record.
That historical result cannot support current novelty. The targeted
inspection in \Cref{tab:residual-falsifier} credits overlap, including
SIP-RC's rendered-view checks and revocation, but establishes neither a
complete substitute nor its inexpressibility. The preregistered
strongest-composition challenge remains unexecuted. The contribution is
publication-specific design knowledge with explicit falsification conditions,
not a selected-criteria ranking or global-priority claim.

\section{Design Science Method and Requirements}\label{sec:method}

\subsection{Research design and unit of analysis}

PAC-2026 was designed before we assembled the literature matrix. That timing
creates a methodological risk: the requirements might simply restate the
artifact. We therefore use the matrix as a retrospective audit, not as evidence
that the requirements were derived independently of PAC. To expose that risk
rather than hide it, we traced each requirement from an external source to a
failure class, then to an observable protocol obligation, evaluation criterion,
and falsifier. We also tested PAC-2026 and its predecessor against that chain.
Existing AIJIM artifacts did not supply the literature base or establish the
problem's prevalence. The eight design requirements (DR1--DR8) still reflect
an artifact-first design, so some
retrofit risk remains.

Within that constraint, DSR requires a recoverable path from a relevant problem
and prior knowledge to objectives, construction, demonstration, and evaluation
\citep{hevner2004design,peffers2007dsrm,gregor2013positioning}. The unit of
design in this study is one fixed-profile publication-accountability
transition, not a deployed product. The unit of evaluation is a named claim
about the frozen protocol or a historically identified implementation layer.
This separation prevents a working implementation from standing in for a
protocol theorem and prevents a finite model from standing in for field
utility.

\subsection{Failure classes and requirement admission}

Five failure classes organize the design input. FC1 is unbound or incomplete
claim evidence; FC2 is missing or overstated run, artifact, or measurement
state. Both have direct external grounding. FC3 is authorization that targets
an earlier materially different state; FC4 is drift between the authorized
record and the registered reader surface. FC3 and FC4 are analytic,
adversarial failure hypotheses because direct prevalence evidence is absent.
FC5 is lifecycle opacity after challenge, correction, supersession, or
withdrawal; its general need is supported by existing update systems, while
the exact transition vocabulary remains an artifact choice.

A candidate requirement was admitted only when omitting it left one of these
declared failures possible and when it could be expressed as an observable
obligation with a direct falsifier. The resulting requirements are not claimed
to be universally minimal, jointly sufficient, or novel. A simpler rule or an
existing profile that preserves the same distinctions can narrow or defeat
them.

\begin{SpecHeroTable}[t]
\caption{Candidate design requirements, protocol realization, and direct
falsifier. ``Profile'' means the fixed admitted PAC-2026 Core Profile.}
\label{tab:design-requirements}
\begin{tabularx}{\textwidth}{@{}P{0.07\textwidth}P{0.25\textwidth}
P{0.31\textwidth}Y@{}}
\toprule
\textbf{DR} & \textbf{Requirement} & \textbf{PAC realization} &
\textbf{Direct falsifier} \\
\midrule
DR1 & Policy-declared evidence closure & Typed claim--evidence loci and E1 &
Required support is absent, mutable, ambiguous, or bound to another target. \\
DR2 & Run, artifact, and measurement closure & Profile-required run IDs,
states, targets, artifact digests or typed absence (R1); typed measured,
not-measured, or unverifiable state (M1) &
A material artifact is replaced or missing analysis is rendered as neutral,
zero, or pass. \\
DR3 & Canonical candidate identity & Typed digest over profile and complete
candidate input closure & Materially different candidates share authority, or
the target cannot be reconstructed. \\
DR4 & Revision-exact human authorization & Authenticated A1 record over exact
candidate, authority, scope, limitations, trust, and time & Approval is forged,
stale, expired, revoked, out of scope, or inherited after material change. \\
DR5 & Registered surface preservation & Candidate-specific intent, manifest,
renderer/extractor, and Core projection (S1) & A registered surface omits,
promotes, substitutes, or cannot reconstruct a required observable. \\
DR6 & Separate results and honest absence & Disjoint applicability,
substantive, and promotion maps; no scalar aggregation & One result strengthens
another, silence becomes success, or unverifiability is relabeled as failure. \\
DR7 & Origin-independent verification & Portable canonical records, versioned
vectors, and binding-refinement obligation & Verification requires hidden
producer state or an unearned substrate assumption. \\
DR8 & Continuing challengeability & Authenticated append-only publication and
post-publication events (L1) & A predecessor is overwritten, a successor is
unlinked, or stale current status survives an unverifiable prefix. \\
\bottomrule
\end{tabularx}
\end{SpecHeroTable}

DR1--DR2 close evidence and computation; DR3--DR4 make the authorized target
exact; DR5--DR6 protect what can be inferred from the reader-facing state;
DR7 removes dependence on one live origin; and DR8 carries accountability past
the signing event. They prescribe no database, signature suite, transparency
service, model provider, or interface topology.

These levels are related but not one-to-one. FC1--FC5 motivate DR1--DR8; the
five laws abstract the properties those requirements seek to preserve;
E1--L1 instantiate them for the fixed profile; Propositions~1--4 expose
protocol-level falsifiers; and the four evaluation criteria test the resulting
artifact. \Cref{tab:design-requirements,tab:claim-evidence-ledger} provide
compact traceability views; neither is a normative one-to-one crosswalk.

\subsection{Build--evaluate learning cycles}

The artifact developed through linked but retrospectively reported cycles. The
2025 AIJIM Reference Model line (RM-2025), published as a conceptual
architecture in 2026~\citep{tiltack2025refmodel}, represented evidence, claims,
runs, artifacts, governance policies, and separated human decision authority.
The published model does not supply PAC's executable publication-transition
semantics. PAC adds exact candidate identities, separate result maps,
freshness-checked authorization, registered surface correspondence,
authenticated lifecycle transitions, and an at-most-once permit. The controlled
precursor cases reported below are separate historical evidence, not results
established by the reference-model publication.

Subsequent implementation attempts changed the semantic object only when a
reproducible counterexample demanded it. SF-2 failed closed but could not
express both an authentic predecessor authorization and a distinct current
target while returning a portable localized stale result. SF-3 introduced a
target-separated A1 input and a non-authorizing relation assessment. A later
counterexample showed that SF-3 reused one time as both evidence cutoff and
verification instant, permitting proof issuance before the recorded decision.
SF-4 preserved both predecessors and separated evidence horizon \(h_e\) from
verification instant \(t_v\), with an explicit causal order. These cycles are
design-learning evidence, not a preregistered empirical experiment.

\subsection{Evaluation strategy and strongest rival}

Following the Framework for Evaluation in Design Science (FEDS)
\citep{venable2016feds}, the current evaluation is primarily artificial and
organizes the reported build--evaluate cycles through four criteria:

\begin{enumerate}
  \item \emph{semantic coherence}: cross-schema, registry, identity, and
  transition consistency;
  \item \emph{bounded safety}: invariants in configured finite models;
  \item \emph{targeted fault sensitivity}: predefined invalid cases,
  mutants, and expected localized outcomes; and
  \item \emph{constructibility}: one local fixed-profile reference path and
  historically separate precursor cases.
\end{enumerate}

Naturalistic utility, organizational effect, adoption, and market value are
not evaluated.

Novelty is treated as a claim to attack, but the historical comparison does not
resolve it. The 27 July 2026 procedure mapped documented source capabilities
and an input-restricted, model-assisted reconstruction function by function.
It did not credit the additional coordination rules needed by the composed
profile. Its conditional result therefore cannot establish that a coherent
composition is unable to substitute. The second mapping also lacks the retained
model, prompt, and sampling configuration needed for an exact rerun. We retain
it as method history, not current novelty evidence. The separately
preregistered successor challenge allows newly written integration rules to
succeed, but remains unexecuted. Later source inspection informs the analytical
comparison without relabeling the historical corpus or supplying that missing
experiment.

The RQ is answered by the design knowledge and transition semantics; SQ1 by
the independently grounded failure classes and candidate requirements; and
SQ2 by the four criteria above. \Cref{sec:evaluation} reports the semantic
freeze, historical SF-3 reference path, and RM-2025 precursor as separate
evidence layers so that no result silently upgrades another.

\section{Publication Authority Design Knowledge and PAC-2026}
\label{sec:design}

The design problem is not how to attach more metadata to a document. It is
how to govern a change in recorded status without allowing evidence about one
state, or success on one dimension, to authorize a different state. The
generalizable design response is
\emph{Publication Authority}: a derived capability for one exact publication
act. PAC-2026 SF-4 is the fixed-profile specification artifact that
operationalizes this response. The governed schemas, registry, vectors,
validators, and models---not this explanatory article---control its executable
semantics.

\subsection{Five laws for an accountable publication transition}
\label{ssec:invariants}

The five laws below are the primary design knowledge. They state what must be
preserved even if a later implementation changes its storage system,
signature envelope, transparency service, model provider, or interface. They
do not assert that every such implementation already preserves the laws.

\paragraph{Law 1: Accountability Closure.}
Every profile-required locus has an explicit typed input state. Unknown,
unavailable, and not measured are declared conditions, not empty fields that
can be interpreted as success. The closure contains the evidence on which
checks operate, but neither verifier outcomes nor witnesses generated during
verification; including either would make success self-justifying. Closure is
profile-relative and does not imply that every relevant fact in the world has
been captured. A verifier can check that every locus required by the admitted
profile has a typed state; it cannot discover an undeclared causal dependency.
Identifying material dependencies---including mutable services, hosted models,
dynamic data, or ephemeral runtime state---therefore remains a profile-design
and governance responsibility. If a required dependency cannot be
authenticated, its state is not silently upgraded to success.

\paragraph{Law 2: Authorization Freshness.}
Human authorization applies to one exact candidate and context. A material
mutation, changed trust snapshot or evidence horizon, revocation, expiry, or
wrong target prevents reuse of the prior A1 result. Advancing the verification
instant \(t_v\) likewise invalidates that result and requires a new complete
A1 evaluation bound to the new instant. It does not by itself require a new
human decision or a rewrite of the immutable \texttt{AuthorizationRecord}; the
existing record can pass only if its proof, target, scope, limitations, trust,
revocation, expiry, and causal-order checks still pass in the new context. The
later verification must still occur no later than the record's activation
time, as specified below. The record remains durable evidence of a past
decision, but is not a transferable
publication capability. A relation assessment may explain why a predecessor
decision is stale or why that determination is unavailable; it cannot
preserve, revive, or create authorization.

\paragraph{Law 3: Epistemic Non-Amplification.}
Each accountability coordinate must earn its own stronger state. Integrity
does not establish truth; provenance does not establish evidential quality;
authorization does not establish measurement; reproducibility does not
establish scientific validity; and visual polish does not establish claim
support. For any \(i\neq j\), a witness \(W_j\) for coordinate \(j\)
cannot, under the fixed profile, establish a stronger state
\(\sigma_i\prec_{\pi_0,i}\sigma'_i\) for coordinate \(i\). This
non-conversion rule rejects aggregate scores in which strength on one
dimension compensates for failure or absence on another.

\paragraph{Law 4: Semantic Surface Preservation.}
Here, semantic preservation is profile-relative registered-observable
correspondence: a registered reader surface must reconstruct the
profile-defined observables for its exact candidate, phase, evidence horizon
\(h_e\), and verification instant \(t_v\). For registered surface \(j\),
renderer \(r_j\), extractor
\(e_j\), and lifecycle notice \(\lambda_{\phi,h_e,t_v}\), the design requires

\begin{equation}
  \operatorname{ext}_{e_j}
  \!\left(\operatorname{rend}_{r_j}
  (c,\lambda_{\phi,h_e,t_v})\right)
  =P_{0,\phi,h_e,t_v}(c).
  \label{eq:surface-preservation}
\end{equation}

Here, \(P_0\) is the fixed profile's Core projection and \(\phi\) selects
the prepublication or later lifecycle phase. In plain terms, reading back the
registered surface must recover exactly the
protocol-defined observables for that candidate, phase, evidence horizon, and
verification instant. The manifest, renderer, extractor, and projection
identities are bound. This is exact correspondence over registered
machine-readable observables within the registered extractor's observation
scope, not a theorem about unrestricted natural-language equivalence or reader
comprehension. A change outside that scope need not be detected by S1.
The protected observables include the claim revision, disposition,
measurement state, limitations, and lifecycle notice. Recovering an embedded
field is not, by itself, evidence that a reader can see or understand it:
if an extractor reads a hidden field, a visibility change outside its
registered observation scope can escape detection. The renderer--extractor
registration therefore defines what correspondence is actually checked.

\paragraph{Law 5: Continuing Challengeability.}
Publication is not terminal. Challenges, resolutions, corrections,
supersessions, and withdrawals append authenticated, target-bound events
rather than rewriting their predecessors. If current status cannot be
reconstructed from the admitted lifecycle view at the requested verification
instant, the registered surface
must expose that unverifiability instead of presenting a stale status as
current. This requirement governs an executed current-status check; it does
not automatically update an unattended static copy or establish that the
supplied history contains every event. The observation boundary is made
explicit below in Section~\ref{ssec:publication-authority}.

Together, the laws make Publication Authority exact and non-compensating,
and keep the recorded publication status surface-visible and challengeable
after its permit has been consumed. They are candidate design
knowledge: a simpler semantics or an existing profile that preserves the same
distinctions would narrow or defeat the contribution.

\subsection{Fixed-profile instantiation}
\label{ssec:fixed-profile}

PAC-2026 SF-4 fixes one admitted Core Profile, \(\pi_0\). It distinguishes a
candidate constructor, human authorizer, registered obligation checkers,
verification engine, renderer--extractor pair, lifecycle custodian,
challenger, and relying verifier. One organization may perform several roles,
but their records and decisions remain typed separately. Verification takes
the digest-bound Core Registry and Profile, canonicalization and hash suite,
checker and reason-code registries, declared trust snapshot, authority, time,
and lifecycle inputs as explicit inputs rather than ambient assumptions.

\paragraph{Trust and fault boundary.}
Within the declared inputs, the profile is designed to catch omissions or
substitutions, stale or revoked authorization, cross-coordinate promotion,
misbound results, reader-surface drift, corrupted lifecycle history, replay,
and transport tampering. It either rejects a structurally invalid record or
localizes the problem as \texttt{FAIL} or \texttt{CANNOT\_VERIFY}; the gate
then blocks promotion. This protection ends at the trust and observation
boundary. It does not cover an authority that approves a false claim,
collusion, an invalid rule admitted as a checker, common-mode false inputs, an
observer that never runs, or denial of service.

A signature has an equally narrow role: it establishes only the integrity and
signer relationship admitted by its binding. It cannot establish evidence
quality or substitute for candidate-specific authorization and Core results.
Exact identities and append-only history may also expose sensitive
relationships. PAC-2026 does not provide encryption, selective disclosure,
access control, retention law, or availability; deployments must handle those
controls separately. Its claim is to preserve declared status and
challengeability, not to determine truth, scientific validity, legal
admissibility, or reader comprehension.

The verifier reconstructs one context \(\chi_{h_e,t_v}\) from the Core
Registry and Profile, policy and trust snapshots, exact candidate,
authorization, surface and lifecycle inputs, and publication target. The
profile maps that context to its six-coordinate state:

\begin{equation}
  \tau_{\pi}(\chi_{h_e,t_v})
    =\langle E,R,M,A,S,L\rangle .
\label{eq:typed-publication-state}
\end{equation}

This context notation is article shorthand, not a new frozen object type. The
six coordinates are evidence closure \(E\), run--artifact closure \(R\),
measurement disclosure \(M\), exact authorization \(A\), surface
correspondence \(S\), and lifecycle continuity \(L\). They are not reduced to
a scalar score. Nor are they necessarily separate files or causal stages.
They are separate judgments and typed witness roles: one content-addressed
artifact may be referenced as a run output, measurement evidence, and claim
evidence without copying its bytes. The verdicts remain distinct because the
existence of a run artifact does not establish that a required measurement was
performed or disclosed, and neither fact establishes evidential support for a
claim. The supplementary responsibility-separation table gives one profile-relative
analytic responsibility-separation scenario for each coordinate; these are
not executed single-coordinate ablations.

The two temporal coordinates serve different evidential functions. The
evidence horizon \(h_e\) is the immutable cutoff for the candidate's evidence;
the verification instant \(t_v\) is when the complete evaluation and
its admitted authentication and trust inputs are checked. Neither denotes the
authorization decision, permit-validity interval, publication time, or time
of a later lifecycle event. Advancing \(h_e\) admits a different evidence
set, while advancing \(t_v\) may change freshness, expiry, revocation, trust,
or reconstructed current status. For a new publication act, either change
requires a complete new evaluation, not a rewritten timestamp on an old record.
Three uses must therefore be distinguished. Historical reconstruction checks
the recorded publication against its original bound inputs. A later status
check freshly evaluates the lifecycle and registered surface at its own
\(t_v\), without reviving the consumed permit or requiring the original A1
result to pass at that later instant. A new publication act instead requires
all six obligations to pass in one fresh context, including A1's
\(t_v\leq t_f\) restriction. These are uses of the existing frozen objects,
not new record types or exceptions to their admission rules. A verifier that
cannot authenticate a required time returns the registered non-passing
classification rather than inventing order from document position or local
file metadata.

The fixed profile realizes these coordinates through six always-applicable
obligations:

\begin{description}[style=sameline,leftmargin=2em]
  \item[E1 Evidence closure.] Required claim--evidence relations resolve to
  authenticated loci under the candidate policy.
  \item[R1 Run--artifact closure.] Every run required by the profile is named.
  Each entry records its state and exact target; a completed run binds one or
  more authoritative artifact digests, while a not-run or no-run-required state
  binds a typed absence witness.
  \item[M1 Measurement disclosure.] Measurement is explicitly
  \texttt{MEASURED}, \texttt{NOT\_MEASURED}, or
  \texttt{CANNOT\_VERIFY}, with the required supporting record.
  In frozen SF-4's \texttt{MEASURED} branch, HC-0024 exposes an unresolved operational
  requirement concerning the relationship between the evidence digest,
  the candidate's commitment, and resolvable measurement bytes. This can
  affect M1 acceptance; Section~\ref{ssec:eval-synthesis} and the supplementary
  adjudication report the finding, which records no incorrectly issued permit.
  A later correction candidate addresses the binding requirement without
  amending SF-4 or claiming byte availability.
  \item[A1 Exact authorization.] An authenticated human decision binds the
  exact candidate, actor, authority, scope, disposition, limitations, trust,
  and time.
  \item[S1 Surface correspondence.] Every registered surface preserves the
  candidate's exact Core projection under its declared renderer and extractor.
  \item[L1 Lifecycle continuity.] Publication and subsequent events preserve
  authenticated order and immutable target identity within the supplied
  lifecycle view.
\end{description}

R1 binds the declared run state and artifact identities within the
fixed profile; it does not by itself guarantee exact replay. Those are
different evidential claims. Exact replay requires a profile to capture a
sufficient execution envelope for the relevant system. A stochastic or closed
system may instead support a separately defined distributional reproduction
test. Where neither condition is met, the defensible result is audit-only
preservation of the original frozen output. SF-4 does not retroactively require
model or provider revision, complete prompts, sampling settings, tool traces,
retrieval snapshots, dependencies, post-edit history, or implementation
identity through the R1 schema.

Objects receive typed, context-separated identities. For each complete
object, the frozen operation forms an envelope that binds its domain, object
type, required context, and body; canonicalizes that complete envelope under
a restricted RFC~8785 profile; and hashes the canonical UTF-8 bytes with
SHA-256~\citep{rundgren2020rfc8785}. The compact formal construction appears
as Supplementary Eq.~(S1); the machine-readable freeze remains controlling.

Registry and Profile identities are recomputed from their complete bodies,
not accepted from caller labels. A \texttt{PublicationCandidate} binds exact
claim bytes and dispositions, evidence relations, required runs and
artifacts, measurement states, policy and profile, requested authority,
predecessors, and a pre-candidate \texttt{SurfaceIntent}. A later
\texttt{SurfaceManifest} binds the already existing candidate digest without
creating a circular identity. A different candidate identity inherits no
authority. Not every presentation change creates a new identity, however:
the frozen candidate construction determines that boundary, and S1 separately
checks the registered surface.

The resulting act has six ordered stages. First, a \texttt{SurfaceIntent}
records the meaning-bearing inputs. Second, those inputs form one canonical
\texttt{PublicationCandidate}, the exact target of every later decision.
Third, after that target exists, candidate-specific authorization, evaluation,
surface, and lifecycle-plan records bind the human decision, intended
presentation, and available post-publication actions. At this prepublication
phase, S1 checks the declared renderer--extractor path, while L1 checks the
target-bound plan and required event-capability domain rather than demanding an
already existing \texttt{PUBLISH} event. The lifecycle input calls this phase
\texttt{PREPUBLISH}; \texttt{DRAFT}, \texttt{REVIEWABLE}, and
\texttt{AUTHORIZED} are preparation labels, not additional lifecycle-input
phases. Fourth, the verification engine
assembles one complete, ordered \texttt{VerificationRecord}. Fifth, only a
fresh record whose substantive and binding checks all pass can support a
single-use permit. Sixth, one atomic act consumes that permit and creates the
immutable \texttt{PublicationRecord} and first authenticated \texttt{PUBLISH}
event. L1 thereafter checks the authenticated ordered prefix, and registered
surfaces expose its current-status projection.

Three distinctions are essential. Candidate identity freezes what is being
evaluated; publication identity exists only after authorization,
verification, surface, and lifecycle obligations close. A complete negative
\texttt{VerificationRecord} is valid portable evidence of a diagnosis, not a
publishable record. Finally, a technical binding transports typed semantics
but does not define them: a repository, file, signature envelope, or receipt
cannot create a PAC verdict merely by containing similarly named fields.

Consider an editor who changes a meaning-bearing sentence after candidate
\(c_1\) has been authorized, thereby creating \(c_2\). The authentic record
still targets \(c_1\); it does not become false merely because a successor
exists. A1 instead evaluates that immutable record against the separately
bound current target. Established material non-equivalence yields localized
stale failure; an unavailable relation yields localized unverifiability; and
relabeling the old record as \(c_2\) yields target mismatch or structural
invalidity. Every path blocks permit derivation for \(c_2\). The example
illustrates the distinction between preserving historical evidence and
transferring operative authority: the former is required, whereas the latter
is prohibited without a new exact-target decision and complete evaluation.

The version history is deliberately not relabeled. SF-2 failed closed but
could not represent both an authentic predecessor authorization and a
distinct current target while returning the required portable stale result.
SF-3 introduced \texttt{AuthorizationEvaluationInput:1} and a separate,
non-authorizing \texttt{CandidateRelationAssessment}; its latter object can
report only material non-equivalence or unverifiability, never
authority-preserving equivalence. A later counterexample showed that SF-3
used one horizon as both evidence cutoff and verification instant, permitting
a proof to predate the decision it purported to authenticate. SF-4 retains
the immutable evidence horizon \(h_e\), introduces the verification instant
\(t_v\), and mints exactly four incompatible \texttt{:2} schemas:
\texttt{AuthorizationEvaluationInput:2}, \texttt{VerificationRecord:2},
\texttt{PublicationPermit:2}, and \texttt{LifecycleContinuityInput:2}. Their
\texttt{:1} predecessors remain historical SF-3 objects.

\begin{figure*}[!t]
  \centering
  \captionsetup{width=\textwidth}
  \includegraphics[width=0.995\textwidth]{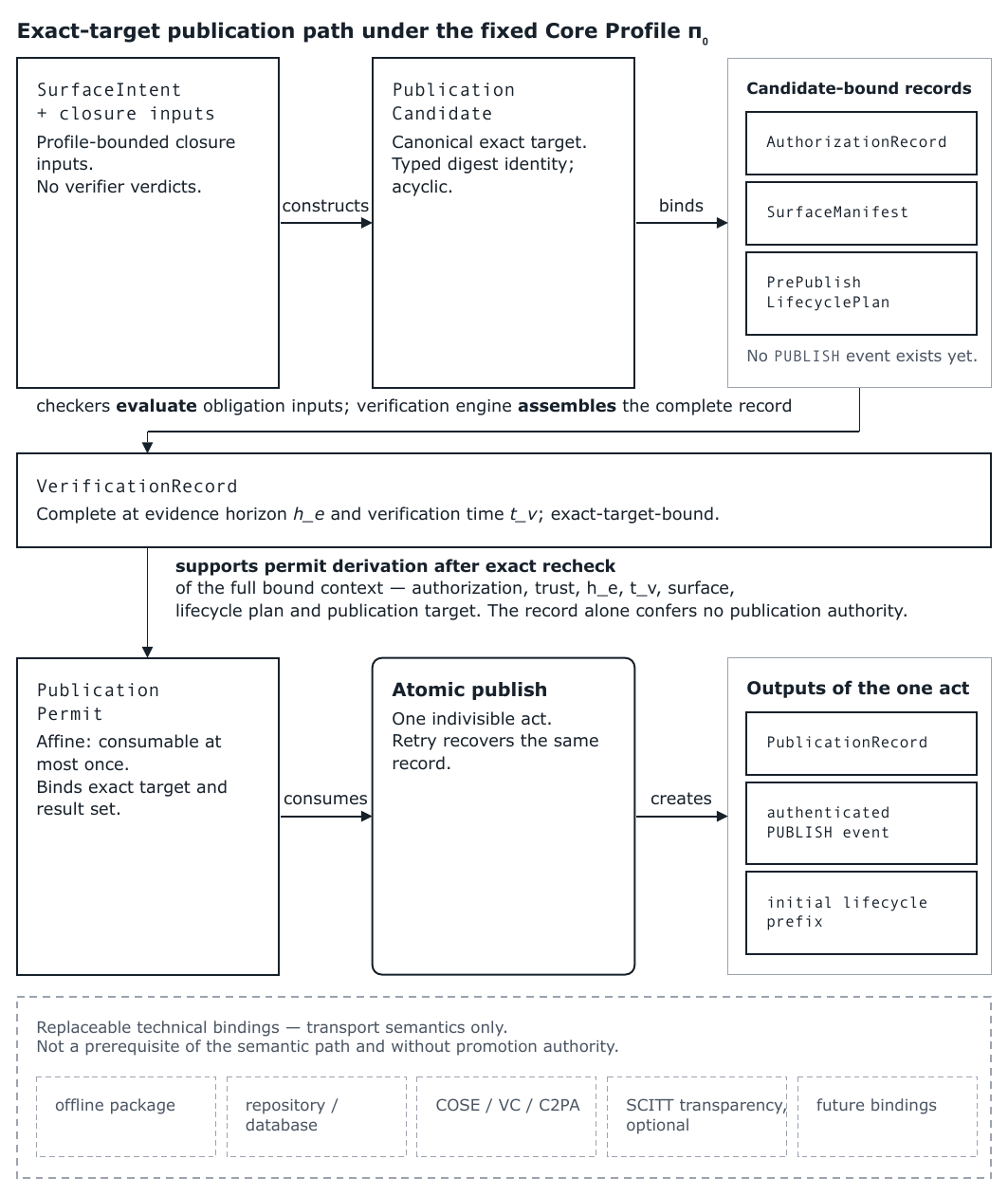}
  \caption{\textbf{Exact-target publication path under the fixed profile.}
  A \texttt{SurfaceIntent} forms one canonical candidate. Candidate-specific
  authorization, surface, and lifecycle-plan records feed a complete
  \texttt{VerificationRecord} at \(h_e,t_v\). Only an exact, fresh all-pass
  record supports a single-use \texttt{PublicationPermit}; one atomic act
  consumes it and creates the immutable publication record and first
  lifecycle event. The dashed transport band creates no semantic verdict.}
  \Description{An exact-target flow from SurfaceIntent and candidate through
  separate authorization, surface, lifecycle, and verification records to a
  single-use permit and atomic publication act. A disconnected band lists
  replaceable technical transports with no promotion authority.}
  \label{fig:pac-publication-path}
\end{figure*}

\subsection{Separate outcomes and no silent promotion}
\label{ssec:result-algebra}

For every applicable obligation, the substantive result is exactly

\begin{equation}
\begin{aligned}
\mathsf{Substantive}\in\{&
  \operatorname{PASS}(W^+),\operatorname{FAIL}(C^-),\\
  &\operatorname{CANNOT\_VERIFY}(U^?)\}.
\end{aligned}
\label{eq:substantive-result}
\end{equation}

\texttt{PASS} carries a target-bound positive witness. \texttt{FAIL} carries
a localized, machine-checkable counterexample to the named rule.
\texttt{CANNOT\_VERIFY} records that an admitted check cannot determine the
named obligation under its registered reason-code rule; it does not assert
that the rule is false. Schema and admission validity are separate: a missing
required checker input or authenticated execution record, an unknown reason
code, or a mismatched constructor can make the record invalid. Invalidity is
not a fourth substantive outcome. Within an admitted check, an authenticated
omission of a required evidence locus yields the registered \texttt{FAIL}; an
evidence input that cannot be authenticated yields the registered
\texttt{CANNOT\_VERIFY}. These examples do not define a universal classification
for every missing or unauthenticated object.
A separate promotion gate returns \texttt{GATE\_PASS},
\texttt{GATE\_BLOCKED}, or \texttt{GATE\_UNVERIFIABLE}. Thus a substantively
correct result still cannot promote a state if its execution, identity, or
freshness binding is inadequate.

At the executable layer, Law 3 becomes a target-bound admission rule. For
coordinate \(i\), profile \(\pi\) defines a strict local promotion relation
\(\prec_{\pi,i}\) and its required obligations \(\mathcal R_{\pi,i}\). No
Silent Epistemic Promotion requires

\begin{equation}
\begin{aligned}
&\sigma_i\prec_{\pi,i}\sigma'_i,\quad
  o\in\mathcal R_{\pi,i}(\chi'_{h_e,t_v})\\
&\quad\Longrightarrow
  \operatorname{Sub}_o(x_o)=\operatorname{PASS}(W_o^+),\\
&\qquad\quad
  \operatorname{Gate}_o(W_o^+)=\operatorname{GATE\_PASS}.
\end{aligned}
\label{eq:no-silent-promotion}
\end{equation}

Here, \(\chi'\) is the successor target context, \(\operatorname{Sub}_o\)
the substantive check, and \(\operatorname{Gate}_o\) its separate binding
check. Each \(x_o\) is reconstructed from authenticated inputs for the target
context, never from the predecessor or desired status. The witness binds the
obligation, coordinate, rule and checker versions, schema and input digest,
locus, target identity, candidate, profile, trust snapshot, verifier,
evidence horizon, and verification instant. Equation
\eqref{eq:no-silent-promotion} is a necessary admission condition, not a
score or a sufficient rule for publication. Put plainly, a stronger state on
one coordinate must be earned by that coordinate's own passing check and
binding witness.

\FloatBarrier

\begin{SpecHeroTable}[!t]
\caption{PAC-2026 coordinate-local promotions. The machine-readable Core
Registry is controlling. Every listed promotion requires fresh substantive
\texttt{PASS} and separate \texttt{GATE\_PASS} records for the named hard
obligation over the exact target state; no row contributes to a global score.
The supplementary material expands the short obligation labels to their exact
registered identifiers.}
\label{tab:promotion-witness-matrix}
\begin{tabularx}{\textwidth}{@{}P{0.18\textwidth}P{0.31\textwidth}Y@{}}
\toprule
\textbf{Coordinate} & \textbf{Concrete promotion} &
\textbf{What must be freshly verified} \\
\midrule
Evidence closure (E1) &
\texttt{UNASSESSED}, \texttt{NONCONFORMING}, or
\texttt{CANNOT\_VERIFY} $\rightarrow$ \texttt{CLOSED} &
Authenticated evidence loci and exact claim-revision relations. \\
Run/artifact closure (R1) &
\texttt{UNASSESSED}, \texttt{NONCONFORMING}, or
\texttt{CANNOT\_VERIFY} $\rightarrow$ \texttt{CLOSED} &
Profile-required run IDs, per-run state and exact target, and authoritative
artifact digests or typed absence. \\
Measurement disclosure (M1) &
\texttt{NOT\_MEASURED} or \texttt{CANNOT\_VERIFY}
$\rightarrow$ \texttt{MEASURED} &
New exact-target measurement evidence on a later fresh evaluation; predecessor
absence is neither zero nor pass. \\
Authorization (A1) &
\texttt{UNAUTHORIZED}, \texttt{STALE}, \texttt{REVOKED}, or
\texttt{EXPIRED} $\rightarrow$ \texttt{AUTHORIZED} &
A new target-bound human \texttt{AuthorizationRecord} and
\texttt{AuthenticationProof}; no old record is revived. \\
Surface correspondence (S1) &
\texttt{UNASSESSED}, \texttt{MISMATCH}, or
\texttt{CANNOT\_VERIFY} $\rightarrow$ \texttt{CORRESPONDENT} &
The registered renderer and extractor reconstruct the exact Core projection. \\
Lifecycle continuity (L1) &
After an authenticated event and fresh lifecycle reconstruction,
\texttt{PREPUBLISH} $\rightarrow$ \texttt{PUBLISHED} or
\texttt{CHALLENGED} $\rightarrow$ \texttt{RESOLVED}; also
\texttt{CANNOT\_VERIFY\_CURRENT\_STATUS} $\rightarrow$ any freshly verified
published-state notice listed below &
Authenticated typed events and predecessor links through the check's \(t_v\),
with candidate evidence fixed at \(h_e\); recovery also requires fresh L1
\texttt{PASS} and \texttt{GATE\_PASS}.
\\
\bottomrule
\end{tabularx}
\par\smallskip
\noindent\textit{Current-status recovery is an L-coordinate promotion, not a
new lifecycle event.} Its target notice is \texttt{PUBLISHED},
\texttt{CHALLENGED}, \texttt{RESOLVED}, \texttt{CORRECTED},
\texttt{SUPERSEDED}, or \texttt{WITHDRAWN}. Fresh prefix reconstruction
appends no event and rewrites no authenticated lifecycle state. The A1 row's
new-record requirement applies to its listed promotion edges, not to every
later evaluation of a still-valid decision record.
\end{SpecHeroTable}
 
Honest absence follows from separating measurement state from checker result.
A target-bound \texttt{NOT\_MEASURED} declaration may yield M1
\texttt{PASS/GATE\_PASS} if the checker verifies the required absence
evidence. That pass establishes faithful disclosure only; it neither promotes
\(M\) to \texttt{MEASURED} nor establishes factual truth or scientific
sufficiency. By contrast, substantive \texttt{CANNOT\_VERIFY} means that M1
itself could not be checked and blocks authority derivation. A structurally
valid negative verification record remains portable diagnostic evidence but
is not publishable.

\begin{figure*}[!t]
  \centering
  \captionsetup{width=\textwidth}
  \includegraphics[width=\textwidth]{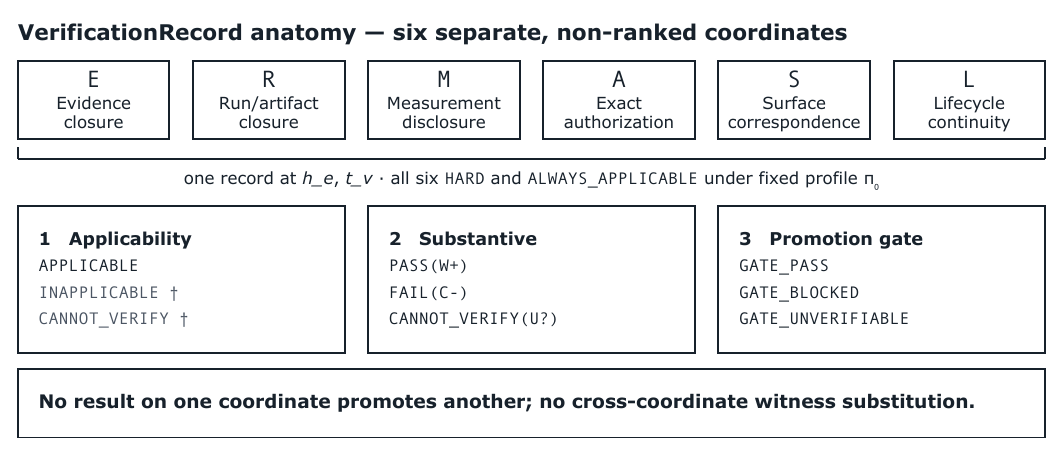}
  \caption{\textbf{Anatomy of a complete \texttt{VerificationRecord}.}
  The six coordinates are separate and non-ranked. Applicability, substantive
  outcome, and promotion authority are disjoint maps. The displayed values
  are the permitted algebra, not one observed run. The fixed profile admits
  all six Core obligations as applicable; no coordinate can promote another,
  and no global truth or quality score is formed. \(\dagger\) marks
  applicability values belonging only to the general extension algebra;
  either value makes a record invalid for the six fixed Core obligations.}
  \Description{Six independent coordinate cells---E for evidence closure, R
  for run--artifact closure, M for measurement disclosure, A for exact
  authorization, S for surface correspondence, and L for lifecycle
  continuity---sit above three distinct result maps. Publication requires PASS
  and GATE PASS on all six applicable obligations without compensating
  aggregation. Dagger-marked applicability values are excluded for these fixed
  Core obligations.}
  \label{fig:pac-verification-record}
\end{figure*}

\subsection{Deriving and consuming Publication Authority}
\label{ssec:publication-authority}

An \texttt{AuthorizationRecord} persistently records a decision attributed
to an admitted human authority about one exact candidate. A1 authenticates
that recorded decision; assurance that the person actually made it also
depends on identity assurance, credential custody, and governance. The human
role is required, but a valid proof is not evidence of personal presence.
Its proof covers
a canonical, target-bound authorization statement; it cannot cover a
caller-selected digest or incomplete projection. A1 evaluates the immutable
record through an \texttt{AuthorizationEvaluationInput:2} that separately
binds the current target and \(h_e,t_v\). Exact-current authorization can pass
only if the proof, trust, target, scope, limitations, revocation, expiry, and
causal order all pass. A material predecessor returns localized stale failure;
unavailable relation evidence returns localized unverifiability. Neither path
transfers authority.

Let \(V_{h_e,t_v}\) be the complete record, \(\mathcal O_0\) the six Core
obligations, and \(\nu\) an unused nonce. Readiness requires an admitted exact
context, complete and fresh record, exact active authorization, accepted
surface and lifecycle plan, and both substantive and gate pass for every Core
obligation. In the following compact notation,
\(\chi=\chi_{h_e,t_v}\) and \(V=V_{h_e,t_v}\). Here \(\alpha\),
\(\varsigma\), \(\ell\), and \(y\) denote the authorization, surface,
lifecycle-plan, and publication-target components of \(\chi\), respectively:

\begin{equation}
\begin{aligned}
\operatorname{Ready}_{\pi_0}(\chi,V,\nu)
&\Longleftrightarrow \operatorname{Admit}(\chi)\\
&\land\operatorname{CompleteFresh}(V,\chi)\\
&\land\operatorname{ExactActiveAuth}(\alpha)\\
&\land\operatorname{SurfaceReady}(\varsigma)\\
&\land\operatorname{PlanAccepted}(\ell,y)
 \land\operatorname{Unused}(\nu)\\
&\land\forall o\in\mathcal O_0:
  \operatorname{Sub}_o=\operatorname{PASS}(W_o^+)\\
&\land\forall o\in\mathcal O_0:
  \operatorname{Gate}_o=\operatorname{GATE\_PASS}.
\\[3pt]
\operatorname{Ready}_{\pi_0}(\chi,V,\nu)
&\Longrightarrow (\chi,V,\nu)\vdash\\
&\qquad
  \operatorname{mint}_{\pi_0}(\chi,V,\nu):\\
&\qquad\qquad
  \texttt{PublicationPermit:2}.
\end{aligned}
\label{eq:publication-readiness}
\end{equation}

The unargumented \(\operatorname{Sub}_o\) and \(\operatorname{Gate}_o\)
refer to those check results in \(V\). The symbols \(\land\), \(\forall\),
\(\Longleftrightarrow\), \(\Longrightarrow\), and \(\vdash\) mean
``and,'' ``for every,'' ``if and only if,'' ``implies,'' and ``permits the
derivation of,'' respectively.

The frozen temporal rules additionally require
\(h_e\leq t_d\leq t_a\leq t_v\leq t_f<t_u\), where \(t_d\) is decision time,
\(t_a\) proof issuance, and \([t_f,t_u)\) the authorization-validity interval
bound into the permit. In plain terms, verification finishes no later than
activation; the resulting permit can be consumed within that interval. The
preactivation condition is a restriction of this fixed profile, not a general
consequence of causal ordering. It excludes a fresh all-pass A1 evaluation
after that decision's activation time, even if its validity interval has not
ended. The
authorization proof must be unexpired at \(t_v\), but need not remain
unexpired at publication time \(t_p\). At that time,
\(t_v\leq t_p\) and \(t_f\leq t_p<t_u\) must hold. The publisher rechecks
the exact record, A1, permit, and lifecycle bindings, the permit's validity
and single-consumption state, and the new \texttt{PUBLISH} event's own
authentication and trust requirements at \(t_p\). The same declared
trust snapshot is checked at proof issuance and \(t_v\); changing it requires
reevaluation. Revocation conclusions are limited to the inputs admitted at
\(t_v\); these checks do not automatically discover later external revocations
or establish clock authenticity or trust continuity beyond those inputs.
Here \(\operatorname{ExactActiveAuth}\) abbreviates the exact-target, proof,
trust, scope, revocation, expiry, and temporal checks just stated; it does not
mean that the publication interval has already opened at verification.
Equation~\eqref{eq:publication-readiness} defines admission, rather than proving
every implementation correct: completeness and freshness are checked against
the bound record and authenticated executions; surface and plan readiness
against S1/L1 inputs; and nonce consumption at the atomic publication boundary.
The corresponding vectors and finite models are reported in
Section~\ref{sec:evaluation}.

The permit implication in
Equation~\eqref{eq:publication-readiness} is the only executable
introduction rule for Publication Authority.

The permit binds the exact candidate, Registry, Profile, trust state,
authorization and verification record, surface, lifecycle plan, target,
times, and nonce. The lifecycle label \texttt{AUTHORIZED} is deliberately
narrower: it records A1 pass only and is neither a permit nor Publication
Authority. No user-interface state, signature, receipt, aggregate score, or
partial record provides an alternative derivation.

\begin{figure*}[!t]
  \centering
  \captionsetup{width=\textwidth}
  \includegraphics[width=\textwidth]{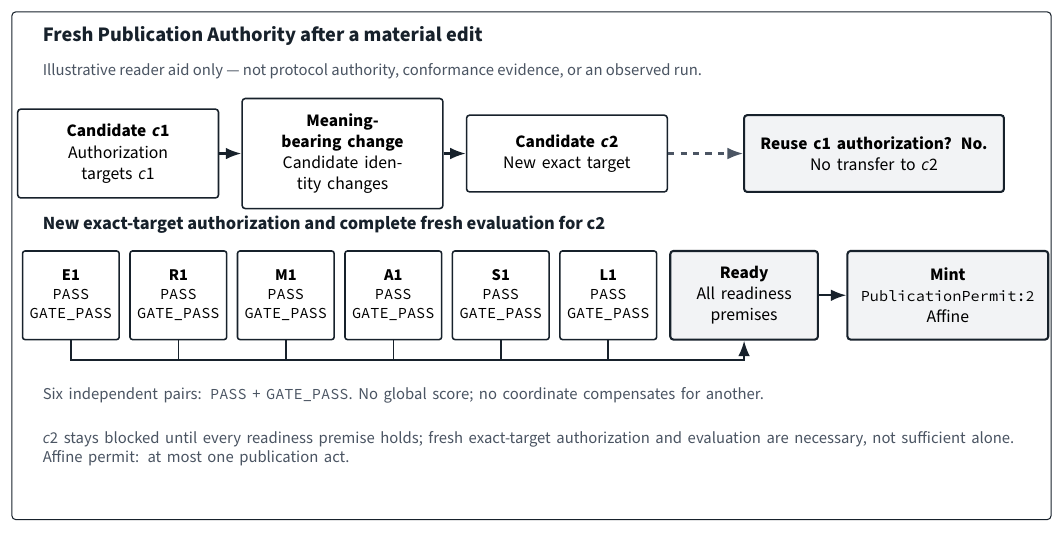}
  \caption{\textbf{Illustrative application of the Publication Authority
  rule.} The upper path shows why predecessor authorization cannot transfer
  after a meaning-bearing edit. The lower path shows the later recovery state:
  all six obligation-local result pairs and every additional readiness premise
  must hold before an affine \texttt{PublicationPermit:2} can be minted. The
  diagram is derived explanatory material, not an observed run, conformance
  evidence, or protocol authority.}
  \Description{A material edit changes candidate c1 into c2 and blocks reuse
  of the predecessor authorization. A later path joins six independent PASS
  and GATE PASS result pairs with all remaining readiness premises before
  minting one affine publication permit.}
  \label{fig:pac-worked-example}
\end{figure*}

Figure~\ref{fig:pac-worked-example} follows one illustrative mutation through
this rule. Once a meaning-bearing edit creates \(c_2\) from \(c_1\), the
authorization targeting \(c_1\) is unusable for \(c_2\). At that intermediate
point, \(\operatorname{Ready}_{\pi_0}(\chi_2,V_2,\nu)\) is false and no permit
exists. The lower path shows the later recovery state, after a new exact-target
authorization and a complete fresh evaluation have been produced for \(c_2\).
Those objects are necessary but not sufficient alone: a permit can be minted
only when every other conjunct in Equation~\eqref{eq:publication-readiness}
also holds. The figure is a reader aid, not an observed evaluation run or a
source of protocol semantics.

\FloatBarrier

The permit is affine: it may be consumed at most once
\citep{walker2005substructural}. One atomic publication act consumes its
digest and nonce and creates exactly one immutable
\texttt{PublicationRecord}, one authenticated \texttt{PUBLISH} event, and
the initial lifecycle prefix. A retry may recover that same record but cannot
create a second publication act. This is per-permit consumption, not a claim
that distinct nonces can never receive distinct permits; stronger global
issuance guarantees remain a production-binding problem.

Before issuance, the lifecycle plan commits to the Core event capability
domain but contains no \texttt{PUBLISH} event. After publication, each event
binds the publication target, event type and time, actor and authority,
authentication, immediate predecessor, and any successor. Current state is
derived from the final authenticated event in the verified ordered prefix,
not accepted from a free-standing status field. Here, ``current'' is relative
to the admitted lifecycle view, not proof of its global completeness. An
authentic prefix ending at \texttt{PUBLISHED} cannot by itself reveal a later
\texttt{CHALLENGE} that was entirely withheld. SF-4 carries no authenticated
latest-head or completeness witness, so that omission does not automatically
produce \texttt{CANNOT\_VERIFY}. A deployment's acquisition and trust
arrangements must establish which lifecycle view it relies on; prefix
integrity alone does not establish that the view is the latest complete one.
The candidate-evidence horizon \(h_e\) remains fixed; a later lifecycle
check authenticates events through that check's own \(t_v\), including events
after \(h_e\). It does not rewrite the original plan or input horizon.

A later edit may restore visible content byte-for-byte to an earlier version.
That restoration neither deletes intervening events nor revives a prior
permit. The candidate is evaluated again at the current \(t_v\), trust, and
lifecycle state, and any publication uses a newly derived permit. An earlier
human authorization record need not be recreated only if A1 finds that it
remains exact, current, and fresh under those inputs; matching visible text
alone does not establish that result. In particular, the new verification
instant must still satisfy \(t_v'\leq t_f\) for that decision record.

\begin{figure*}[!t]
  \centering
  \captionsetup{width=\textwidth}
  \includegraphics[width=\textwidth]{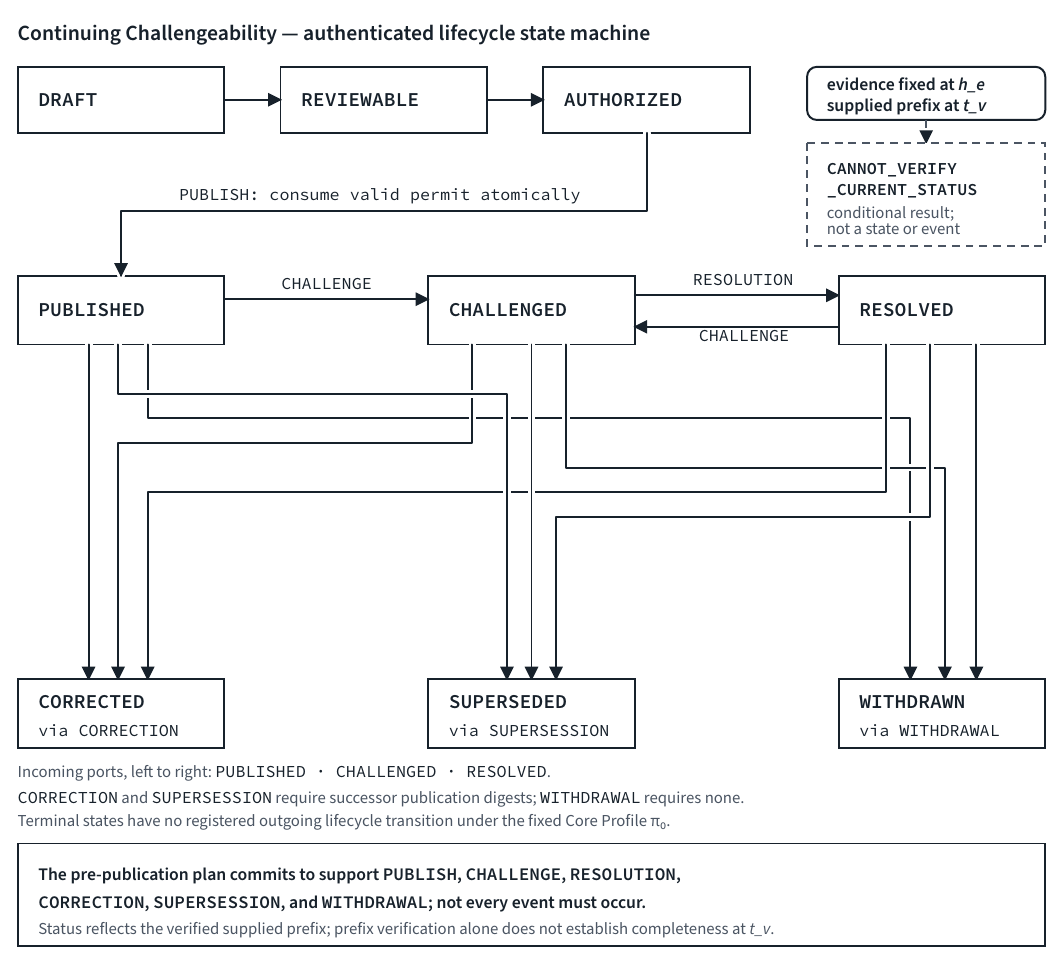}
  \caption{\textbf{Continuing Challengeability as an authenticated lifecycle.}
  Publication and later challenge, resolution, correction, supersession, and
  withdrawal append target-bound events. \texttt{DRAFT}, \texttt{REVIEWABLE},
  and \texttt{AUTHORIZED} are preparation labels; \texttt{AUTHORIZED} records
  only A1 pass, not a permit. With candidate evidence fixed at \(h_e\), current status
  is reconstructed from the supplied authenticated prefix through the later check's
  \(t_v\); prefix verification alone does not establish completeness at that
  instant. \texttt{CANNOT\_VERIFY\_CURRENT\_STATUS} is a conditional visible
  verification result, not a state of the authenticated event machine or a
  rewrite of its history. Recovering a verified notice is nevertheless an
  L-coordinate promotion requiring fresh L1 \texttt{PASS} and
  \texttt{GATE\_PASS}.}
  \Description{A nine-state lifecycle machine with two pre-publication
  transitions, publication, challenges and resolutions, and terminal
  correction, supersession, or withdrawal paths. A separate dashed result
  reports unverifiable current status without altering the event history.}
  \label{fig:pac-lifecycle}
\end{figure*}

\subsection{Falsifiable propositions and conformance boundary}
\label{ssec:formal-boundary}

The design claims are deliberately bounded and directly defeasible.

\paragraph{Proposition 1: bounded permit exactness.}
The conformance rule requires every transition to \texttt{PUBLISHED} to
consume exactly one valid permit bound to the exact candidate and complete
publishable VerificationRecord; each permit may be consumed by at most one
distinct publication act, while an idempotent replay may only return the same
bound record. The evaluated transition-system property is that no
reachable trace violates either condition. A reachable trace, evaluated
against this rule, that
publishes without the exact binding, creates a second publication act from the
same permit, or changes the target without reevaluation falsifies the
proposition.

\paragraph{Proposition 2: coordinate-local promotion safety.}
Every stronger coordinate state has its own target-bound \texttt{PASS}
  witness and separate \texttt{GATE\_PASS}. An observed transition promoted
  by another coordinate's witness or by compensating aggregation falsifies
  the proposition.

\paragraph{Proposition 3: stale non-inheritance and localized non-pass.}
A materially non-equivalent successor cannot inherit predecessor Publication
  Authority. An observed execution that silently accepts such inheritance,
  or collapses a known violation and unavailable determination into the same
  outcome, falsifies the proposition.

\paragraph{Proposition 4: continuing status integrity.}
Within the admitted lifecycle view, authenticated predecessors are not
rewritten, and unverifiable
  current status is not presented as verified. An observed trace that deletes
  or reorders admitted history, or silently retains stale current status after
  prefix verification fails, falsifies the proposition. This property does
  not establish discovery of events absent from the supplied view.

These are protocol contracts assessed through schemas, validators, adversarial
vectors, and finite models; they are not universal metatheorems. Checker
soundness and termination, mutation coverage, cross-model composition,
type preservation, clock authenticity, and common-mode input faults remain
outside the checked boundary. A trusted authority can still authorize a false
claim, and an observer that never executes cannot detect drift.

Conformance attaches to a versioned role and its required inputs, outputs,
and fail-closed behavior, not to a brand or possession of one valid record.
Historical SF-3 conformance says nothing about SF-4; SF-4 says nothing about a
later profile without an explicit successor rule. A technical binding must
show that decoding preserves candidate and profile identity, all three result
maps, authorization invalidation, permit consumption, registered surfaces,
ordered lifecycle effects, and trust and time updates:

\begin{equation}
  \operatorname{decode}_b(\rho_b)\simeq_{\pi_0}\rho .
\label{eq:binding-refinement-target}
\end{equation}

This equation states the general refinement target; it is not a theorem.
Here, \(b\) is a technical binding, \(\rho_b\) its encoded trace, and
\(\rho\) the protocol trace; \(\simeq_{\pi_0}\) means agreement on the
profile-protected observations.
In plain terms, changing the technical transport must not change any protected
protocol meaning. Substrate neutrality is therefore a design objective only.
Later work tests
separately versioned successor bindings, but cannot make SF-4 production-bound
or retroactively establish general refinement. The machine-readable SF-4
registry remains the controlling representation; this chapter explains its
semantic spine without replacing it.
\FloatBarrier

\section{Evaluation and Results}\label{sec:evaluation}

The evaluation asks how far the fixed-profile design satisfies the four
criteria defined in \Cref{sec:method}: semantic coherence, bounded safety,
targeted fault sensitivity, and constructibility. Following FEDS
\citep{venable2016feds}, the first three criteria are evaluated artificially
with executable schemas, canonical vectors, validators, finite models,
and countermodels. The historical rival comparison is reported separately as
method history. Constructibility is evaluated with a
private historical SF-3 reference path and separately identified RM-2025
precursor cases. These objects have different semantic identities and cannot
upgrade one another's claims.

\subsection{Evaluation design and evidence layers}
\label{ssec:eval-boundary}

\Cref{tab:evaluation-layers} separates the three evidence layers before any
result is interpreted. The PAC-2026 layer evaluates the SF-4 specification
artifact. The reference-path layer tests selected predecessor behavior in one
implementation lineage. The precursor layer demonstrates selected lifecycle
and observer behavior that informed the later design. None is a field trial,
user study, certification, or population estimate.

\begin{SpecHeroTable}[t]
\caption{Evaluation objects, supported inferences, and scope boundaries.}
\label{tab:evaluation-layers}
\begin{tabularx}{\textwidth}{@{}P{0.18\textwidth}P{0.29\textwidth}
P{0.25\textwidth}Y@{}}
\toprule
\textbf{Layer} & \textbf{Evaluation object} & \textbf{Licensed inference} &
\textbf{Scope boundary} \\
\midrule
PAC-2026 SF-4 freeze &
Executable registry and vectors, two validation procedures, ten finite
models, and countermodels &
Internal identity and schema coherence, bounded invariant preservation,
and fault localization &
No universal theorem, checker correctness, production refinement,
independent implementation, interoperability, or standard status \\
Historical SF-3 reference path &
Seventeen fixed A1 successor cases and a 373-test source/distribution census &
Constructibility of the tested target-separated A1 partition and selected
record--permit behavior &
No SF-4 implementation, complete conformance, independent replication,
production readiness, or certification \\
RM-2025 precursor &
Controlled lifecycle, three-locus, and separate-consumer cases &
Selected fail-closed, observer-separation, and consumer-contract behavior in
the historical reference architecture &
No PAC-2026 conformance, factual truth, field effectiveness, adoption, or
general interoperability \\
\bottomrule
\end{tabularx}
\end{SpecHeroTable}

Numerical results count declared bytes, configurations, and executions rather
than estimate population prevalence or effectiveness. The vectors, cases,
mutants, and models form the study's versioned, protocol-specific conformance
and fault-sensitivity corpus. \Cref{tab:evaluation-results} links each
observation to its evaluation scope; detailed case and model inventories are
retained in the supplement.

\begin{SpecHeroTable}[t]
\caption{Criterion-led results. Counts are exact only for the declared frozen
corpus, configured finite models, or recorded executions.}
\label{tab:evaluation-results}
\begin{tabularx}{\textwidth}{@{}P{0.17\textwidth}P{0.27\textwidth}
P{0.29\textwidth}Y@{}}
\toprule
\textbf{Criterion} & \textbf{Evaluation procedure} &
\textbf{Bounded result} & \textbf{Claim ceiling} \\
\midrule
Semantic coherence &
Registry validation, separately written digest reconstruction, successor-vector and
exact-key checks &
36 schemas; 53 typed reason codes; all 23 SF-4 temporal cases matched their
declared tuples; all 17 SF-3 cases retained as passing regressions &
Coherence of the frozen identities and cases, not general temporal soundness
or implementation conformance \\
Bounded safety &
Exhaustive TLC exploration of ten configured finite models plus injected
unsafe configurations &
110,764 distinct reachable safe states without invariant violation; 76
expected unsafe witnesses produced the declared violation or blind spot &
Named finite configurations only, not cross-model or unbounded protocol
composition, checker correctness, coverage percentage, or a universal theorem \\
Targeted fault sensitivity &
Negative vectors, countermodels, and complete-record relabeling attempts &
Declared mutations localized; non-pass paths did not derive Publication
Authority &
No exhaustive adversary coverage, independent replication, or novelty
inference \\
Constructibility &
Private historical SF-3 reference path and separately identified RM-2025
precursor cases &
All 17 SF-3 successor cases matched; the same 373 tests passed in source,
extracted-source-distribution, and installed-wheel runs; selected precursor
controls matched &
One implementation lineage and historical cases only; no SF-4 implementation,
independent interoperability, field effectiveness, or production readiness \\
\bottomrule
\end{tabularx}
\end{SpecHeroTable}
 
\subsection{Criterion 1: semantic coherence}
\label{ssec:eval-coherence}

The SF-4 freeze contains 36 JSON Schema 2020-12 definitions, 53 typed
constructor-specific reason codes, and the six fixed-profile obligations. The
primary validator checked schema validity, cross-registry references,
obligation and reason-code assignments, the three result-map domains,
authorization and permit bindings, surface and lifecycle objects, and
positive and adversarial instances. A separately written Node.js procedure
reconstructed the canonical digest vectors. Both procedures reproduced the
frozen Core Registry and Core Profile identities, establishing agreement on
the canonical identities for these bytes.

The successor history supplied two stronger coherence tests. SF-2 failed
closed when authorization did not target the current candidate, but it could
not express an authentic predecessor authorization, a different current
evaluation target, and a portable localized stale result in the same input.
SF-3 separated those coordinates in a typed authorization-evaluation input and
made relation assessment diagnostic rather than authorizing. Its 17 frozen
cases were retained as predecessor regressions; no historical case or schema
was relabeled.

SF-4 addressed a separate temporal-causality counterexample. In the bound
predecessor reproducer, an authentication proof dated before the human
decision still reached \texttt{PASS} and publication under SF-3. Rebinding
that construction to SF-4 yielded localized
\texttt{FAIL/A\_TEMPORAL\_ORDER\_INVALID}. This is a demonstrated correction
relative to the protocol's own predecessor, not a comparison with an external
system. SF-4 preserves the
evidence horizon \(h_e\), adds verification time \(t_v\), and subjects every
authorization evaluation and permit to the causal and half-open validity
constraints defined in Section~\ref{ssec:publication-authority}. Exactly four
incompatible objects receive new \texttt{:2} schemas:
authorization-evaluation input, verification record, publication permit, and
lifecycle-continuity input. An exact-key comparison confirmed the declared
and observed result tuples for all 23 frozen SF-4 temporal cases, while the 17
SF-3 cases remained passing regressions. This demonstrates coherence of the
declared successor corpus and localized treatment of its tested temporal faults.

\subsection{Criterion 2: bounded safety}
\label{ssec:eval-safety}

The ten-model TLA+ suite is deliberately factorized. Within it,
\texttt{ProfileAdmissionPublish} is a bounded integrated admission-to-publish
model: it couples all six Core obligations, their three result maps,
authorization freshness, permit derivation and consumption, and atomic
creation of the \texttt{PublicationRecord}, \texttt{PUBLISH} event, and initial
lifecycle prefix. The remaining models isolate authorization and invalidation,
disclosure--surface correspondence, correction lineage,
witness-or-localized-nonpass behavior, coordinate promotion, one bounded
binding refinement, producer/surface/audit observer separation, SF-3 target
separation, and SF-4 temporal causality. The TLA+ model checker (TLC)
exhaustively explored 110,764 distinct reachable safe
states in the ten configured finite models without an invariant violation.
Separately, 76 deliberately unsafe configurations produced the expected
invariant violation or, in the observer model, the declared blind spot. The 76
runs are expected unsafe witnesses, not additional states within the safe
spaces and not observations of 76 real-world failures.
Six modules are inherited unchanged from SF-2 (80,376 states and 37 unsafe
configurations); four new or rebound modules account for the remaining 30,388
states and 39 unsafe configurations. The inherited modules do not test the
SF-4 verification-time semantics. The supplementary inventory gives each
module's source layer and exact configuration.

The countermodels matter more than the magnitude of the state count. They cover
four fault families: partial or cross-coordinate promotion and outcome
relabeling; stale, transferred, wrong-target, or replayed authority; missing
or reordered lifecycle events; and temporal or observer faults. They also
reproduced four earlier design failures---the missing demotion, under-bound
permit, SF-3 target gap, and SF-4 temporal gap---each of which led to a
recorded correction before the next freeze.
These classifications reuse the same unsafe configurations; they are not
independent additional evidence.

Because the models factor the state space and abstract checker correctness,
the result is bounded to the named configurations. It is not evidence of
cross-model or unbounded composition, termination, general refinement,
arbitrary-checker soundness, or observer correctness.

\subsection{Criterion 3: targeted fault sensitivity}
\label{ssec:eval-fault-sensitivity}

The executable negative corpus covers four fault families: record and
result-map integrity; identity and registry substitution; stale, revoked, or
replayed authority; and surface or lifecycle inconsistency. The observed
outcomes remained localized:
non-pass results were carried as records but did not create Publication
Authority in the tested cases. Such an expected rejection is a successful
negative control: the publication object fails, while the protocol test
passes. This supports sensitivity to the declared mutation grammar. It
does not establish exhaustive attack coverage or the semantic correctness of
future checkers.

Two substitution controls show what the joint binding adds to local checks.
A replacement surface with altered limitations was consistently re-digested
and passed its own correspondence check; a replacement lifecycle plan likewise
passed its local check. Permits re-digested to reference either replacement
were nevertheless rejected because the objects were not those bound into the
accepted \texttt{VerificationRecord}. The controls test individually valid
objects in the wrong publication context, not merely corrupted hashes. A
further control retained a complete non-passing record as evidence while
rejecting a permit derived from it. Together, these frozen controls isolate
the role of joint record--permit binding in publication authorization.

The measurement-disclosure controls test a different distinction. Typed
\texttt{NOT\_MEASURED} and \texttt{CANNOT\_VERIFY} entries with their required
witnesses passed the disclosure checks; removing the respective absence or
unverifiability witness caused schema rejection. Here \texttt{CANNOT\_VERIFY}
is a declared measurement state, not the M1 checker's result: an unverifiable
M1 outcome still blocks authority derivation. The controls establish the
witness requirement for declared absence, not the validity of a measurement.
They do not resolve HC-0024, which concerns evidence binding and resolution in
the separate \texttt{MEASURED} branch. The supplementary artifact index locates
these controls, the substitution controls, and the temporal predecessor
reproducer within the versioned collection~\citep{tiltack2026aijimArtifacts}.

\paragraph{Historical substitution method.}
The 40-source comparison closed on 27 July 2026 returned
\texttt{NO\_COMPLETE\_SUBSTITUTE\_IN\_BOUND\_CORPUS}, but its treatment of
newly written integration rules and later-discovered source omissions prevent
its use as current novelty evidence. The second mapping was input-restricted,
not independently replicated or exactly rerunnable. The supplementary
historical-rival section retains the procedure and its F9 disagreement.
The frozen corpus used C2PA 2.2; C2PA 2.4 was available in April 2026 but first
inspected for this study after the freeze~\citep{c2pa2026spec}. The living
comparison does not rewrite those historical inputs.

\subsection{Criterion 4: bounded constructibility}
\label{ssec:sf3-reference-path}

The private historical SF-3 reference path exercises one exclusive A1
partition: exact-current authorization passes; an authentic but materially
non-equivalent predecessor is stale; missing or unusable relation evidence is
localized as \texttt{CANNOT\_VERIFY}; and checkable target manipulation
fails. All 17 frozen successor cases produced their expected normalized
observations. The same 373-test suite passed from the source tree, an extracted
source distribution, and a separately installed wheel. Complete-record
assembly rejected coherent relabeling from \texttt{FAIL} or
\texttt{CANNOT\_VERIFY} to \texttt{PASS}, and from
\texttt{CANNOT\_VERIFY} to \texttt{FAIL}. Wrong-record use and permit replay
were also rejected in the tested in-memory scope. The test count measures
engineering coverage in that lineage; the path is local, fixed-profile,
and historical. Its evaluated snapshot is now archived in the public
collection~\citep{tiltack2026aijimArtifacts}; it is not an SF-4 implementation.

Three RM-2025 cases provide separate precursor evidence. SF-P0-01 linked one
runtime identity, evidence-bound claim state, consequential human review,
release, and offline mutation detection; changing one signed runtime digest in
an exported copy yielded the expected integrity failure. SF-P0-02 used a clean
control and isolated producer, surface, and audit mutations. Each designated
observer detected its mutation while the other two observed loci remained
hash-identical, supporting pairwise non-substitutability only for those
mutations and declared views.

The historical separate-consumer case tested whether a separate domain
application could retrieve and fail-closed verify one frozen publication
through the then-evaluated version-1 application programming interface (API).
Its positive path and eleven preregistered controls produced their
expected transport, authorization, trust, freshness, and verification
outcomes. A reviewer without a project role reproduced the recorded result
from checksummed artifacts without the live Core, credentials, or private
signing material. The consumer was a separate application, but not an
independently developed AIJIM implementation or an organizationally
independent customer; the offline verifier shared the canonical lineage.
The three deposits used in that historical review were previously public but
are currently \texttt{RESTRICTED}. This article therefore makes no current
public-access claim, and the case remains historical evidence rather than a
presently reproducible external evaluation.

\subsection{Synthesis and post-evaluation status}
\label{ssec:eval-synthesis}

The frozen study connects the design to observable behavior. Its executable
checks preserve the tested record--permit bindings, the finite models preserve
the encoded invariants, and the adversarial cases localize the declared
faults. The predecessor comparison demonstrates a temporal correction; the
substitution controls demonstrate why local validity does not suffice for
publication authority; the absence controls distinguish correct disclosure
from a claim that measurement occurred. The historical implementation layers separately show
that selected target, result, and permit rules can be constructed and
executed. These findings answer the four evaluation criteria within their
stated scopes (\Cref{tab:evaluation-results}); the later studies below add
evidence for separately identified objects and methods.

For traceability, the research programme labels its evidence stages G0--G8.
These are internal research-stage labels, not protocol elements, standard
levels, or maturity ratings. G0--G5 cover only the frozen SF-4 study. The two
historical implementation layers do not extend that result, and the historical
Audit Bundle 2.4.0 format directly binds neither an SF-4 nor SF-5
\texttt{PublicationRecord}.

Subsequent work produced separately versioned records. The binding-refinement study
(G6) matched all 24 expected outcomes for two registered bindings of a later
semantic successor (SF-5): 12/12 in the Core-native path and 12/12 in the
versioned API-to-Core path.
The 22 negative cells were rejected before permit derivation, and no false
acceptance of a \texttt{PublicationPermit} or \texttt{PublicationRecord} was
observed. A disclosed correction to an auxiliary result schema changed no
result cell and required no rerun.
This supports only those exact bindings, source baseline, corpus, and decision
rule. It neither retroactively implements SF-4 nor establishes deployment,
general production readiness, or independent reproduction.

\paragraph{Implementation uptake.}
Separately from this study, an engineering team independently develops an AIJIM implementation
from the protocol specification and documentation. A manuscript-oriented
application integrates this Core through distinct admission,
personal-authorization, publication, export, and verification operations,
translating protocol decision boundaries into application actions.
The supplementary inventory identifies the implementation and connects each
operation to retained local checks or source-inspected assertions. It
distinguishes preparation and synthetic authorization from real publication,
and inner SF-5 verification from outer RFC-021 package qualification.
The inventory supports realization of selected obligations in the later
Core/RFC line, not a completed production demonstration. It is separate from
the frozen SF-4 study and September's G6 corpus; the application integrates
this Core rather than supplying a second semantic implementation for G7.

Independent-implementation conformance is labelled G7. A disclosed-case
calibration matched all 92 expected declarations across 161 role invocations,
but it was not blind because the cases and expected-result structure were
available during development, so the result measures cross-path calibration.
Two earlier hidden-run attempts yielded no usable protocol evidence because
relevant material had been exposed and the apparatus failed; their records
are retained separately in the supplement.

\paragraph{In-house, instance-blind multi-build consistency (G7-B).}
On 12 September 2026, two producers (reference and Node) and two verifier
paths (reference and Rust) were exercised in four pairings against SF-4.
The builds, fixtures, parameter domains, expectations, and analysis were
frozen before a preregistered public randomness-beacon round, whose output
determined three sets of 80 instances. The instance values were unknown at
freeze time; case classes and expected-result rules were known. The two
qualification runs already reached the final endpoint after pre-freeze
withdrawals and reaggregation, without changed generation, journals, or
per-set analyses. Their original 192-scored/9-discrepancy and final
183-scored/0-discrepancy results remain in the supplement. G7-B tests fresh
values of pre-screened classes, not class blindness or an independently
conducted evaluation; \(N_{\mathrm{new}}\) counts new registered discrepancies.

Before freezing, AI-assisted integration changed all three evaluated builds,
partly to align with the reference and known outcomes. Consequently, agreement
tests consistency under instance-blind conditions, not independent convergence
on the specification. The design does not quantify the contribution of shared
integration to agreement or identify a demonstrably unaffected subset.

The 80 archetypes comprised 61 scored, 17 withdrawn, and two specification
ambiguities: 183 scored and 57 unscored instances across the three sets.
The registered endpoint required four complete, identical observations
matching the expectation; a crash could not count as rejection. All 183
scored instances matched: 75 expected acceptances and 108 typed expected
rejections, giving \(N_{\mathrm{new}}=0\) and
\texttt{G7-B\_CONSISTENT}. ``Acceptance'' is the analyzer's operational
category, not publication permission: 66 valid-verification-record cases with
A1 \texttt{PASS}, three equality-boundary cases, and six atomic-publication
cases. Fifteen of the 66 correctly retain localized \texttt{FAIL} or
\texttt{CANNOT\_VERIFY}. The scored set includes 21 instances that confirm the
admission baseline rather than add fault-sensitive coverage. In 36 scored instances, the two
Rust-path terminal classifications came from the bridge---the harness adapter
between implementations---and its operation table,
not independent Rust evidence for that target. These observations retain
their registered score but cannot be counted as native Rust confirmations.
Across all 240 instances, four-way agreement held for 219, whose producer
portable outputs also agreed. All 21 non-agreeing instances belonged to
withdrawn classes. Five registered observations, including ambiguities and
divergence F-01 on inadmissible input, recurred in every set; agreement on an
unscored class is not a conformance result.
The two ambiguities remain unresolved in frozen SF-4: HC-0024 concerns M1
acceptance, whereas HC-0062 concerns authoritative time and rejection
localization. Their interpretations and consequences remain in the
supplementary artifact index. Neither case demonstrates an incorrectly issued
permit.

Later provenance handoffs connect preserved pre-integration source archives to the evaluated
builds through matching imports and documented changes. The registration
describes the reference and Node/Rust lines as independently developed;
the source trail establishes their recoverable pre-integration identities,
not a complete account of original development responsibilities or information
exchange. These handoffs supplement the original package; they do not make
G7-B an independent evaluation or establish general G7 conformance. The fixture
review
also rejected apparent confirmations: HC-0033 initially matched because of an
unintended timing violation, not its declared event swap. Output agreement
alone was therefore insufficient evidence of fault coverage. The supplement
records the pre-freeze disposition history, component limits, and timing proof.

\paragraph{A bounded correction candidate.}
A separately versioned correction makes the M1/R1 commitment requirements
constructive: checkers reconstruct complete sets, validate measurement-evidence
bodies, and bind producing runs and value references to the same candidate.
The value check establishes reference membership, not byte availability.
The candidate also makes intra-evaluation time consistency explicit, including
offset-equivalent instants and rejection of substituted times. Its local
validator completed 49 declared correction cases, including 25 mutated cases,
alongside the 17 SF-3 regressions and 23 SF-4 temporal cases. These additional
checks neither re-score G7-B nor amend its historical ambiguities. The
supplement identifies the unratified candidate, reproduction procedure, and
remaining implementation and evaluation limits.

\paragraph{Reproduction from the combined evidence package.}
A freshly extracted package was executed in a new environment on the same
macOS arm64 host, without a production service or private source repository.
The SF-4 registry, digest, and finite-model checks were re-executed, as were
the historical SF-3 suite (373 tests each from source, a fresh source
distribution, and an installed wheel) and all four G7-B pairings. G7-B
reproduced all 240 archived case observations with no endpoint or signature
differences. This is repeatability of the completed study, not 240 additional
scored cases or another blind experiment. The rival's 40-source/45-search-event
records, the 24 SF-5 binding cells, and the 92 calibration cells were instead
reanalyzed from archived records; those stages did not repeat the original
search, construction, or live executions. All six package stages completed
without a reported failure, and all 2,951 manifest-listed files were unchanged
before and after execution. Exact identities and execution instructions are
in the supplement's artifact index. The unchanged evidence and accompanying
instructions are now publicly archived~\citep{tiltack2026aijimArtifacts}.

G8 denotes separate studies of release governance (G8-A),
institutional use (G8-B), and human or reviewer effects (G8-C). The latter two
are needed for corresponding field and human-effect claims, not for the
bounded semantic result reported by this artifact article.

The combined package makes the SF-4 study, later binding and consistency
results, and local reproduction inspectable as distinct evidence layers.
Their versioned inputs and recorded outcomes support targeted re-execution
and scrutiny. Section~\ref{ssec:limitations} consolidates the remaining
conformance, interoperability, and field-evaluation questions.

\subsection{Supplied protections and transferable repairs}
\label{ssec:native-protection-comparisons}

Three targeted diagnostics identify protection already supplied by unchanged
SF-4 and the integration needed to reproduce it in a constructed alternative.
The alternative uses native in-toto 3.0.0, OPA 1.19.1, and a transactional SQLite
publisher; it is our publication profile, not an unmodified competing protocol.
This author-led, machine-assisted comparison was code-aware: expectations,
scenario definitions, and source closures were frozen before execution, after inspecting
both implementations. It is neither a blind nor an external independent
evaluation. Mutations preceded fresh signatures and approvals; the compared
arms received equivalent task information.

The earlier V2 comparison remains counterevidence to exclusive capability:
each arm matched 21 of 23 locked cases, with two undeclared-origin cases
unassessed rather than failed or passed. Separately, removing the
\texttt{resultSetDigest} relationship at both existing SF-4 comparison sites
allowed a crossed record/permit to commit; restoring the original prevented
it. The valid controls and stale-candidate rejection were preserved.
This ablation isolates the effect of the two-site record--permit binding.

\begin{SpecHeroTable}[t]
\caption{Agreement with the prospectively fixed task, per diagnostic.
Counts describe selected cases, not population accuracy or independent
discoveries. \emph{Prior} means V2, V2 plus the state repair, and V2 plus
state/time repairs, respectively. Each row's final repair restores parity
on that row's case set.}
\label{tab:comparative-protections}
\begin{tabularx}{\textwidth}{@{}P{0.36\textwidth}P{0.12\textwidth}P{0.19\textwidth}Y@{}}
\toprule
\textbf{Protection family} & \textbf{SF-4} &
\textbf{Prior profile} & \textbf{Repaired profile} \\
\midrule
Measurement-state admission & 9/9 & 5/9 & 9/9 \\
Proof validity at verification & 5/5 & 4/5 & 5/5 \\
Required-run omission/substitution & 7/7 & 4/7 & 7/7 \\
\bottomrule
\end{tabularx}
\end{SpecHeroTable}

First, SF-4 refused four malformed measurement-state representations before
M1 evaluation; the prior profile committed them. An explicit state-domain
guard closed this single fault class. Second, SF-4 refused a proof expiring
exactly at verification while allowing publication after a validly checked
proof expired. The alternative needed a separate proof-time rule, in addition
to its existing publication interval.

Third, with both previous repairs retained, the alternative committed an
omitted run entry, an omitted run field, and a well-shaped replacement for
the task's required run. A nonempty shape guard corrected the two omissions
(6/7); exact agreement with the separately admitted run identities corrected
substitution (7/7). SF-4's absent field failed input admission. Empty or
substituted runs stopped at authenticated-result admission, leaving the
substantive result unassessed; only the wrong-target control produced an
admitted R1 \texttt{FAIL}. These are distinct refusal stages.

Two fresh executions per diagnostic reproduced decisions and stored effects
without apparatus errors. Native traces and SQLite readback distinguish
refusal from actual publication. The supplementary record supplies all cases,
identities, and stages. The demonstrated value is an integrated contract
covering these three mechanisms, together with transferable repair rules.
Parity after repair applies to each tested set, not an unexecuted omnibus
regression; no global ranking, field error reduction, or independent
validation follows. These diagnostics neither modify SF-4 nor re-score G7-B.
 \section{Reusable assurance through exact claim-edit regions}
\label{sec:edit-regions}

A publication may revise its selected claims while retaining its evidence.
Which claims must remain so that every permitted selection retains a specified
statistical qualification? We answer with an exact edit-region test, a minimum
retained core, and a complete core-size frontier. This exploratory extension
is separate from the SF-4 evaluation. It uses arithmetic-mean e-closure
\citep[Sections~4--5.1, Equations~(5), (14)]{XuEtAl2026EClosure}:
nonnegative evidence variables with expectation at most one under their
respective null hypotheses support simultaneous control of the expected
false-discovery proportion (FDP). That guarantee is not a truth probability
for individual claims.

For a fixed hypothesis universe \(H\) of size \(m=|H|\), evidence vector
\(e\), and level \(\alpha\), an edit region \([M,U]=\{R:M\subseteq R\subseteq U\}\)
retains core \(M\) while allowing every selection of optional claims.
For example, \(e=(8,1,8)\), \(\alpha=1/2\), and \(U=H\) admit
\(M=\{0\}\): \(\{0\},\{0,1\},\{0,2\},\{0,1,2\}\) all qualify,
whereas the weak-only report \(\{1\}\) does not. Core \(\{2\}\) works
equally well. The result identifies reusable selection flexibility, not the
scientific importance of a mandatory claim.

Our exact criterion checks the entire interval and supplies a violating
report/intersection pair when it fails. If any size-\(h\) core qualifies,
the \(h\) largest evidence values in \(U\) do; the first qualifying size
is cardinality-minimal. For \(U=H\), a shared suffix-minimum calculation
decides all core sizes in \(O(m^2)\) arithmetic work after sorting and
\(O(m)\) compact storage. The supplement proves these results, gives the
exact-rational and witness-output boundaries, and distinguishes this
all-interval problem from individual-report and anchored-closure algorithms
\citep{XuEtAl2026EClosure,Goeman2026ClosedBH,Ramdas2026MaximalEFDR}.

Across 728 snapshot/level problems, all 4,010 frontier decisions agreed with
literal enumeration and the retained solver; all 3,856 negative witnesses
violated the defining inequality. Two fresh executions produced identical
reports. These finite checks validate computation on the declared grid, not
e-value calibration or independent empirical replication.

Native integration compared four complete providers under the same inputs,
with 192 functional combinations across AIJIM and an author-built
in-toto--OPA--SQLite alternative. Both can use the mathematical provider.
The AIJIM backend additionally completed 1,096 timed synthetic publication
attempts. Qualification, including setup, was cheaper than both memoized
controls on the 32-distinct-selection sequence, but more expensive on the
other eleven; full endpoint time remained slower than sweep on that sequence.
The supplement retains every paired cost and the specified algorithmic
separation, without converting either into general protocol superiority.

The practical addition is a checkable boundary for reusing analytical
assurance. A covered selection still needs fresh authorization of its exact
candidate and every other native obligation before permit consumption and
publication. Evidence, hypothesis meaning, universe, or level changes require
reassessment. Removing a false mandatory claim likewise triggers reassessment,
never suppression. The region reuses a mathematical qualification; it does
not reuse an old publication grant.

\section{Discussion}\label{sec:discussion}

\subsection{Answers to the research question and subquestions}
\label{ssec:contributions}

\paragraph{Research question.}
Publication Authority makes the permission to publish an explicit, checkable
consequence of the joint record. It exists
only when one admitted fixed profile binds the candidate, trust state,
evidence horizon, verification time, and requested transition to separate,
localized outcomes for evidence, analytical work, measurement, human
authorization, reader-visible surface, and lifecycle. No success may
compensate for a required non-pass elsewhere. A complete fresh all-pass record
may derive one affine permit; a material non-equivalent successor must be
re-admitted and cannot inherit the predecessor's authority. Registered surface
and lifecycle checks carry the same identity beyond the producer interface.
This answers how the transition can remain challengeable: the authority to
publish is a derived and defeasible property of one recorded state, not an
attribute of an author, interface, signature, or earlier revision.

\paragraph{SQ1: failure classes and requirements.}
SQ1 distinguishes what the literature establishes from what the artifact
hypothesizes. Prior work directly motivates closure for evidence and analytical
work (FC1--FC2) and partly motivates lifecycle continuity (FC5). Stale
authorization and reader-surface drift (FC3--FC4) remain analytically grounded
hypotheses rather than externally measured problems. DR1--DR8 operationalize
those classes through exact candidate identity, separate outcomes, fresh
authorization, non-compensating promotion, surface and lifecycle integrity,
and portable verification. They are a candidate synthesis, not an exhaustive
or minimal set.

\paragraph{SQ2: extent of satisfaction.}
The evaluation connects each criterion to an observable protocol property.
For SF-4 itself, cross-schema identities, 36 schemas, 53 reason codes, canonical
vectors, and the registry validator support semantic coherence. Ten
factorized finite models explored 110,764 distinct reachable safe states and 76
deliberately unsafe configurations, supporting bounded safety and targeted
fault sensitivity for the encoded invariants. Twenty-three observed SF-4
temporal cases and all 17 historical SF-3 predecessor regressions further test
the authorization-successor boundary without relabeling the older cases.
Constructibility is supported indirectly by the historical implementation
lineage and precursor cases. Later SF-5 binding, disclosed-case calibration,
and instance-blind G7-B results address their separately identified roles and
paths. Local package reproduction makes the executed checks repeatable from
identified inputs. The version boundaries and inferential limits of these
layers are set out in \Cref{ssec:eval-synthesis,ssec:limitations}.

The correction candidate adds a concrete instance of specification
self-correction: an ambiguous requirement becomes an explicit check without
recasting its original evaluation as a success. This supports a revisable
design, not completeness. Its value-reference membership check proves neither
byte availability nor scientific validity.

The finding behind that candidate also distinguishes implementation agreement
from specification adequacy. Four producer--verifier pairings across three
builds returned M1 PASS on HC-0024, an unscored case whose \texttt{MEASURED}
evidence-binding and resolution requirement was operationally underdetermined
(\Cref{ssec:eval-synthesis}). Agreement did not resolve that requirement;
the contribution of shared integration and known fixtures to the agreement
was not measured. Knight and Leveson's multiversion experiment likewise
found correlated failures on shared inputs despite separate development
\citep{knight1986independence}; it does not establish the cause of our
observation. SIP-RC separately distinguishes decision recomputation from
implementation independence \citep{liu2026siprc}. A second, specification-only
implementation built independently of the evaluated Core and tested in
exchange with the reference would add
evidence about interpretability and constructibility beyond known-fixture
consistency. The separate-team implementation in \Cref{sec:evaluation} is the
Core integrated by the application, not that second implementation. G7-B's four in-house
pairings after shared integration do not constitute this independent
specification-only conformance and interoperability evaluation.

The native comparisons expose three mechanisms that turn publication rules
into enforced protections (\Cref{ssec:native-protection-comparisons}).
State-domain checks close malformed inputs; verification-time checks close
expired proofs; an independently admitted expectation closes required-run
substitution that shape checks leave open. The archived SF-4 binding already
supplies these protections; adding them to the constructed profile as it was
successively repaired recovers the corresponding outcomes. This yields
transferable integration rules with executable counterexamples. Code-aware case selection
and author-built repairs make this diagnostic evidence, not an estimate of
the prevalence of integration faults or an independent evaluation.

\begin{SpecHeroTable}[!b]
\caption{Claim--evidence--falsifier ledger. Each row states the strongest
claim licensed by its evidence layer and a direct defeater or open gate.}
\label{tab:claim-evidence-ledger}
\begin{tabularx}{\textwidth}{@{}P{0.27\textwidth}P{0.36\textwidth}Y@{}}
\toprule
\textbf{Bounded candidate claim} & \textbf{Current support} &
\textbf{Direct falsifier or open gate} \\
\midrule
Publication Authority and non-compensation form testable design knowledge for
the declared publication transition &
Literature-grounded FC1/FC2, partly grounded FC5, analytic FC3/FC4, and the
DR1--DR8 synthesis &
An equivalent simpler semantics, a safe compensating case, or a missing
material coordinate narrows the claim; FC3/FC4 field relevance remains open.
\\

PAC-2026 SF-4 defines an executable fixed-profile semantics &
Six obligations, 36 schemas, 53 reason codes, canonical vectors, digests,
registry validator, and frozen corpus &
An admissible candidate, result, record, or permit bypasses a mandatory
binding. General profile calculus and production refinement remain open. \\

The fixed corpus supports bounded safety and targeted fault sensitivity &
Ten factorized models, 110,764 reachable safe states, 76 unsafe configurations, 23
SF-4 temporal cases, and 17 historical SF-3 regressions &
An in-scope invariant violation is reachable as safe or a targeted fault
passes. Checker soundness, cross-model or unbounded protocol composition, and
general refinement remain open. \\

Precursor, historical, and post-evaluation paths show bounded constructibility &
Private SF-3 execution of 17 frozen cases and a 373-test suite in three
packaging modes; separate RM-2025 cases; a post-evaluation record reporting
12/12 plus 12/12 G6 outcomes for two SF-5 bindings; a later, separately
scoped application-integration inventory &
Recorded behavior cannot be reproduced or diverges from its frozen objects.
SF-4 production binding and independent G7 remain open; disclosed-case
G7 calibration is not blind confirmation. Selected Product checks are not
a complete production demonstration. \\

G7-B supports in-house instance consistency and local repeatability &
183 scored expectations matched, including 21 baseline confirmations and 36
instances with bridge-derived Rust terminal classifications; 240 archived
observations reproduced &
A scored mismatch defeats the registered endpoint. Qualification had already
produced the endpoint; bridge-derived terminals are not native Rust evidence.
Unscored classes and evaluation independence remain outside the claim. \\

A versioned correction makes selected commitment and time requirements
executable &
49 declared correction cases, including 25 mutated cases, with the retained
17 SF-3 and 23 SF-4 cases &
A commitment bypass or inconsistent evaluation time defeats the selected
rule. Value-reference membership does not establish byte availability;
Core admission and full successor evaluation remain open. \\

Three supplied protections close gaps in the tested constructed profile &
Native state-domain, proof-time, and required-run contrasts; cumulative
repairs; two byte-identical observation runs per study &
A correct commit under a prohibited case or a blocked valid control defeats
the tested rule. Repairs restore parity; code-aware diagnostics do not
establish whole-protocol superiority or fault prevalence. \\

Exact edit regions and the all-core frontier permit qualified selection reuse &
Regional and exchange proofs; 4,010 frontier decisions; 192 native functional
combinations; separately reported workload-specific costs &
A failed defining inequality defeats qualification. The complete alternative
can reuse the provider; calibration, independent replication, and general
endpoint acceleration are not demonstrated. \\

The historical rival comparison documents a superseded method &
40 exact sources, six aggregate records, 45 search events, author mapping, and
an input-restricted model-assisted sensitivity check under a rule withholding
credit from unstandardized integration policy &
Relevant pre-cut-off sources were also missing. The result is not current
novelty evidence; the unexecuted successor challenge permits new integration
rules. An equivalent profile or composition defeats the residual.
\\
\bottomrule
\end{tabularx}
\end{SpecHeroTable}
 
\subsection{Contribution hierarchy and relation to standards}
\label{ssec:comparison}

The primary contribution is design knowledge: Publication Authority as a
derived, non-transferable capability with a per-permit single-use
representation, together with the non-compensation rule and five design laws
that govern its derivation and preservation. One permit authorizes at most one
publication act; this does not claim that distinct nonces can never receive
distinct permits. The second contribution is PAC-2026 SF-4, one executable
fixed-profile instantiation of those ideas. The third is criterion-led
evidence about that instantiation. The three contributions connect a
publication-specific coordination problem to explicit rules and executable
tests.

The edit-region extension adds a constructive answer to a different but
connected question: which analytical assurance can survive a change of
selection? The retained-core theorem makes this reuse exact, and the
quadratic frontier exposes the full retention--flexibility trade-off when
the upper set equals the hypothesis universe (\(U=H\)). Each changed candidate
still requires its own authority:
reusable analytical qualification and permission to publish are different
objects. The supplement separates the named-comparator complexity result from
workload-specific native costs. The construction remains usable by a different
conforming architecture.

PAC makes publication-specific dependencies explicit across established
mechanisms. APP already binds human approval and validation
to a repository release; SCITT already authenticates registered statements
and receipts (Section~\ref{sec:literature}). PAC makes three joint obligations
explicit at the publication boundary: authentic approval for \(c_1\) cannot
authorize a materially different \(c_2\); a valid negative verification record
may remain portable without licensing publication; and the registered surface
must preserve the candidate's declared limitations and the status derived
from its admitted lifecycle view. Each addresses a failure that checking
artifact authenticity alone does not exclude. The tested substitution cases
make the distinction observable: a surface or lifecycle plan can be valid on
its own and still be the wrong object for the accepted record and permit
(Section~\ref{ssec:eval-fault-sensitivity}). These controls make joint
publication validity testable separately from individual-object validity.

The contribution can therefore be assessed through the publication conditions
made explicit, the failures they distinguish, and the evidence for their
operation. The design builds on established components while specifying their
publication-level dependencies. The source comparison in Section~\ref{sec:literature} and the executed,
targeted native comparisons serve different purposes: one locates design
knowledge; the other tests particular integrations and repairs. The separately planned composition
challenge remains unexecuted; these diagnostic studies do not retrospectively
fulfil it.

Replaceable binding remains a design objective, not demonstrated substrate
independence: it must preserve candidate identity, localized outcomes,
authorization effects, surface state, and lifecycle order. The historical
Audit Bundle 2.4.0 format directly binds neither SF-4 nor SF-5. A separately
versioned successor later defined an SF-5 one-writer portable artifact and
offline verifier. Subsequent records report all 24 expected outcomes across
two named successor bindings. That is bounded binding-refinement evidence for
those paths, not a public or externally reviewed standard, a production
binding for SF-4, or a general refinement result.

\subsection{Implications for research and practice}
\label{ssec:implications}

For research, the protocol moves accountability analysis from isolated model
outputs to the transition that grants a claim a stronger recorded status. It
makes interactions among provenance, measurement, authorization, presentation,
and correction independently testable and encourages publication of positive
witnesses, localized counterexamples, unverifiability results, and explicit
defeaters. The next useful studies should test incremental diagnostic value,
not vague trust: can external reviewers detect stale authorization, missing
measurement, surface drift, or supersession more reliably from the record than
from a static article or chat export?

A public successor benchmark could expose disjoint positive and adversarial
cases with exact expected outcomes for independent implementations. Passing it
would show behavior only for the represented faults; it could neither prove
general conformance nor replace the evidence and human authorization required
for a particular publication.

For practice, the immediate value is a localized account of which declared
publication obligations passed, failed, or could not be verified and why.
This makes the basis for publication permission inspectable. It cannot warrant
factual truth, scientific quality, legal compliance, editorial legitimacy, or
appropriate reliance. The concurrent approval-integrity evaluation discussed
in Section~\ref{sec:literature} illustrates why decision quality and faithful
approval use require separate assessment \citep{alpay2026approval}; exact binding
cannot make an unsupported decision correct. Human review remains an attributable governance act,
not a substitute for evidence or for a validated reference construct when a
scientific measurement requires one.

Original reporting illustrates this boundary. A first public story may rely on
confidential interviews, notes, recordings, documents, or observations. The
protocol can bind a digest without publicly disclosing those bytes; access
conditions remain separately governed by the deployment. The commitment
records a claimed byte identity for
later comparison; by itself it proves neither existence at the asserted time,
authenticity, truth, nor public inspectability. A reviewer who later receives
the legitimately disclosed bytes can test whether they match the commitment.
Until profile-required material can be
authenticated, the honest outcome is localized \texttt{CANNOT\_VERIFY}, which
blocks promotion. The story itself cannot bootstrap independent support for
its external factual claims; human authorization records responsibility, not
truth.

\subsection{Limitations and threats to validity}\label{ssec:limitations}

\paragraph{Construct and external relevance.}
FC3 and FC4 are derived design hypotheses, not field-validated publication
problems. FC5 is only partly literature-grounded at PAC's granularity. The
six-coordinate decomposition and non-compensation rule may omit a material
coordinate, separate two coordinates unnecessarily, or impose more structure
than a simpler safe rule requires. A rival profile, a valid compensating case,
or an observer made redundant under authentic views would narrow the claimed
design knowledge. No newsroom or other naturalistic study measured
prevalence, comprehension, diagnostic effort, reliance, organizational value,
or downstream effects.

\paragraph{Literature, corpus, and access.}
The critical integrative search is documented and source-bound but not a
systematic review, and a moving standards landscape makes universal novelty
unavailable. The SF-4 freeze, historical SF-3 implementation snapshot, later
binding-refinement and calibration records, G7-B study, and combined
reproduction material are publicly archived in a versioned
collection~\citep{tiltack2026aijimArtifacts}. Its six declared areas distinguish
actual re-execution from archived reanalysis. Restricted RM-2025/SF-P0 material
and the historical separate-consumer deposits are excluded and were not rerun.
Those precursor observations remain externally uninspectable from this
collection and cannot carry independently reproducible claims. Public access
to the included material removes an access barrier, not its methodological
limits or the need for independent evaluation.
The correction candidate is separately archived as the SF-4 Correction
Research Addendum in version 0.2.0~\citep{tiltack2026aijimCorrectionArtifacts};
publication does not change its research-candidate status or the original
version 0.1.0 evidence.

\paragraph{Formal, implementation, and governance bounds.}
The formal evidence is confined to one factorized profile; it does not prove
type soundness, checker correctness, unbounded composition, cryptographic
adequacy, or general refinement. The originating-lineage SF-3 path uses an
in-memory store, so durable, distributed, concurrent, and crash-consistent
permit consumption remains untested. Neither that path nor the precursor cases
is an independent implementation. The later SF-5 result covers only two
bindings and one corpus, not deployment or general production readiness. The
disclosed-case run remains developer calibration and the earlier hidden
attempts remain invalid. G7-B adds instance-blind consistency, but known case
classes, pre-freeze alignment, and prior qualification at the eventual endpoint
limit its inference. Its 183 scored instances include 21 baseline confirmations
and 36 instances with bridge-derived Rust terminal classifications rather than
native-verifier confirmation. Later provenance handoffs recover the source-archive-to-build
chains, but not complete original development-responsibility or exchange
records. Development origin, shared integration, and instance blindness remain
separate evidence questions. The evaluated interchange
also leaves canonicalization, replay/collision, and published-lifecycle
classes outside the demonstrated coverage. The Rust checker is an Apple
Silicon binary; source-to-binary reproduction and other platforms were not
established. Same-host package reproduction does not establish independent
evaluation. Independently evaluated interoperability, evaluation of release
governance, institutional use, and human or reviewer effects remain
unestablished.

\paragraph{Edit-region scope.}
The new derivation concerns a fixed finite hypothesis universe, valid input
e-values, and a fixed error level. Synthetic nonnegative inputs establish
computational behavior, not statistical calibration of AI-generated claims.
The frontier optimizes retained cardinality within an interval; it does not
optimize scientific relevance or all possible selection families. Two
host-local timing runs cannot establish a population speedup, and the
complete alternative can use the same provider. These are separate
author-led exploratory results, not independent confirmation or new SF-4
requirements. Prior results and the earlier negative amortization study
retain their original identities.

\section{Conclusion}\label{sec:conclusion}

Publication Authority makes the permission to publish a checkable consequence
of the complete, exact state being published. This answers the research
question and defines the primary design contribution. Each mandatory
obligation must earn its own fresh passing result and promotion witness;
success elsewhere cannot replace it. An at-most-once permit represents that
authority for a publication act. Historical approval remains evidence, but a
materially different candidate must earn its own authority.

The second contribution, PAC-2026 SF-4, makes these dependencies executable
within one admitted Core Profile. Six obligations connect evidence, analytical
work, measurement disclosure, human authorization, registered presentation,
and lifecycle continuity through exact identities and temporal rules. This
separates the validity of individual objects from the conditions under which
they may jointly authorize publication.

The third contribution is evidence that these distinctions matter. The frozen
validator rejects valid replacement surfaces outside the accepted record,
accepts witnessed absence without promoting it to measurement, and rejects
the proof-before-decision construction accepted by SF-3. Ten finite models
explored 110,764 distinct reachable safe states; 76 unsafe configurations
produced the expected violation or observer countermodel.
Three native comparisons additionally isolate measurement-state, proof-time,
and required-run protections missing from a constructed in-toto/OPA profile.
Explicit repairs restore parity on the tested cases, identifying the
integration rules responsible for these protections.

The separately tested, unratified correction candidate makes the measurement
commitments and time/localization rules implicated by HC-0024 and HC-0062
explicit while preserving frozen SF-4 and its original observations. Its value
check establishes reference membership, not byte availability.
Later SF-5 and G7-B studies retain their distinct objects; shared integration,
known case classes, and bridge-derived classifications limit G7-B's
independence inference.

The exploratory edit-region construction shows how analytical qualification
can be reused without transferring publication authority. It gives exact
regional decisions and cardinality-minimal retained cores. When the upper
set equals the hypothesis universe (\(U=H\)), a quadratic all-core frontier
provides decisions and explicit counterexamples. These results provide
reusable design and algorithmic knowledge alongside the unchanged SF-4 study.

Together, the rules, counterexamples, and inspectable artifacts explain how
publication acquires authority and remains challengeable. They give subsequent
implementations a concrete basis for independent conformance and field studies.
 
\FloatBarrier

\section*{CRediT authorship contribution statement}
\textbf{Torsten Olivi Tiltack:} Conceptualization, Methodology, Software,
Validation, Formal analysis, Investigation, Data curation, Writing -- original
draft, Writing -- review \& editing, Visualization, Project administration.
\textbf{Yifei Dong:} Conceptualization, Supervision, Writing -- review \& editing,
Project administration.
\textbf{Kun Yu:} Conceptualization, Supervision, Writing -- review \& editing.
\textbf{Xu Wang:} Writing -- review \& editing.
\textbf{Wei Liu:} Writing -- review \& editing.
\textbf{Jianlong Zhou:} Writing -- review \& editing.
\textbf{Ren Ping Liu:} Writing -- review \& editing.
\textbf{Fang Chen:} Supervision.

\section*{Acknowledgements}
The authors acknowledge the engineering team at ViaoraAI S.L. for implementing
AIJIM and integrating it into SHARK.

\section*{Funding}
This research did not receive any specific grant from funding agencies in the
public, commercial, or not-for-profit sectors.

\section*{Declaration of competing interest}
Torsten Olivi Tiltack coordinates the development and scientific evaluation of projects using AIJIM,
and anticipates joining ViaoraAI S.L. The authors declare no other competing
financial or non-financial interests.

\section*{Declaration of generative AI and AI-assisted technologies in the manuscript preparation process}
The authors used Codex and Claude Code under human supervision to support
literature discovery, manuscript drafting and revision, and research-code
development. The lead author reviewed all AI-assisted
material and revised it as needed; the authors take full responsibility for
this article.

\section*{Data and artifact availability}
The versioned research artifact collection is publicly archived on Zenodo
as version 0.1.0~\citep{tiltack2026aijimArtifacts}. It contains the PAC-2026
SF-4 freeze, historical SF-3 implementation snapshot, later binding-refinement,
calibration and G7-B material, and the combined reproduction record. Inputs,
recorded outputs, source snapshots, verification instructions, file digests,
and component-specific reuse terms are included. The source companion is
\url{https://github.com/src01001100/aijim-publication-authority}.
The supplement maps claims to exact artifacts and distinguishes re-execution
from archived reanalysis. Restricted RM-2025/SF-P0 material and the three
historical separate-consumer deposits were not included or rerun; their
earlier public availability is not a current access claim. Public access
enables inspection but does not turn same-host reproduction into external
replication or strengthen the reported results.
The separately reported correction candidate is archived in version 0.2.0
as the SF-4 Correction Research Addendum~\citep{tiltack2026aijimCorrectionArtifacts};
it is not part of the unchanged version 0.1.0 evidence.

The additional edit-region research and three native protection comparisons
are publicly archived in a separate scientific reproduction collection,
version 1.0.0~\citep{tiltack2026aijimExecutableEvidence}. It
includes unchanged computational sources and observations, selected native
dependencies, and labelled scientific projections of contract and result
documents. Original prospective documents remain locally retained; the
supplement identifies the distribution's scope and reproduction levels.
This collection does not alter either published artifact version cited above.
Private Product source and local receipts are not distributed as a publicly
reproducible Product study.
 
\bibliographystyle{elsarticle-num}
\bibliography{references}

\end{document}


\let\WriteBookmarks\relax
\shorttitle{Supplement: \PaperShortTitle}
\shortauthors{\PaperRunAuthor}
\title[mode=title]{Supplementary Material: \PaperTitle}

\ifblindreview
  \author[1]{Anonymous author(s)}
\else
  \author[1]{Torsten Olivi Tiltack}
  \author[1]{Yifei Dong}
  \cormark[1]
  \author[1]{Kun Yu}
  \author[2]{Xu Wang}
  \author[3]{Wei Liu}
  \author[1]{Jianlong Zhou}
  \author[2]{Ren Ping Liu}
  \author[1]{Fang Chen}
\fi

\affiliation[1]{organization={Data Science Institute, University of Technology Sydney},
  city={Ultimo},
  state={NSW},
  country={Australia}}
\affiliation[2]{organization={Global Big Data Technologies Centre, University of Technology Sydney},
  city={Ultimo},
  state={NSW},
  country={Australia}}
\affiliation[3]{organization={School of Computer Science, University of Technology Sydney},
  city={Ultimo},
  state={NSW},
  country={Australia}}

\ifblindreview\else
  \cortext[cor1]{Corresponding author: Yifei.Dong@uts.edu.au}
\fi

\begin{abstract}
This supplement retains detailed crosswalks, bounded counterexamples, model
inventories, historical predecessor cases, and artifact identifiers that are
necessary for audit but would obscure the contribution architecture of the
research article. It does not restate or replace the normative PAC-2026 SF-4
machine-readable freeze.
\end{abstract}
\maketitle
\ApplyPaperPDFMetadata{Supplementary Material: \PaperTitle}

\setcounter{table}{0}
\renewcommand{\thetable}{S\arabic{table}}

\section{Authority and Reading Boundary}
\label{sup:boundary}

The article presents design knowledge and bounded evaluation; this supplement
retains detail required to audit those claims. Neither document is the
normative protocol source. The semantic authority for the evaluated SF-4 layer is the governed
machine-readable \PACFreezeId{} package, version \PACFreezeVersion{}, with the
root and manifest identifiers reported below. SF-2 and SF-3 remain immutable
historical predecessors. Counts, rendered tables, and prose in this supplement
are projections of those identified objects and must not be used to relabel or
modify them.

The full schemas, reason-code registry, canonical vectors, validators, and
model configurations are therefore referenced by package identity rather than
copied into typeset prose. This avoids creating a second source that could
drift from the executable bytes. The tables below state the bounded analytical
or empirical role of each projection and the stronger claim it does not
support.

\section{Formal and Registered Identifier Detail}
\label{sup:formal-detail}

\paragraph{Typed identity construction.}
For auditability, the article-level shorthand for the frozen typed-identity
operation is retained here. For object type \(q\), required context \(x\), and
body \(b\),

\begin{equation}
\begin{aligned}
  \operatorname{Env}(q,x,b)
    &=\{\mathit{domain},\mathit{objectType}=q,\\
    &\qquad \mathit{context}=x,\mathit{body}=b\},\\
  \operatorname{TD}(q,x,b)
    &=H_0\!\left(K_0(\operatorname{Env}(q,x,b))\right).
\end{aligned}
\tag{S1}
\label{eq:sup-typed-digest}
\end{equation}

Here \(K_0\) denotes the admitted canonicalization operation and \(H_0\) the
admitted SHA-256 operation. This projection does not replace the governed
Registry, Profile, schemas, or test vectors.

\paragraph{Registered obligation identifiers.}
The article shorthand expands as follows; the machine-readable Core Registry
remains controlling.
\begin{center}
\footnotesize
\begin{tabularx}{0.94\linewidth}{@{}P{0.13\linewidth}Y@{}}
\toprule
\textbf{Article label} & \textbf{Registered identifier} \\
\midrule
\texttt{E1} & \texttt{PAC\_CORE\_E1\_EVIDENCE\_CLOSURE} \\
\texttt{R1} & \texttt{PAC\_CORE\_R1\_RUN\_ARTIFACT\_CLOSURE} \\
\texttt{M1} & \texttt{PAC\_CORE\_M1\_MEASUREMENT\_DISCLOSURE} \\
\texttt{A1} & \texttt{PAC\_CORE\_A1\_EXACT\_AUTHORIZATION} \\
\texttt{S1} & \texttt{PAC\_CORE\_S1\_SURFACE\_CORRESPONDENCE} \\
\texttt{L1} & \texttt{PAC\_CORE\_L1\_LIFECYCLE\_CONTINUITY} \\
\bottomrule
\end{tabularx}
\end{center}

\section{Versioned Design and Risk Crosswalks}
\label{sup:crosswalks}

\begin{SpecHeroTable}
\caption{Supporting evidence: versioned architectural-family crosswalk. A mapping denotes
conceptual refinement, not identity and not retrospective evidence for
PAC-2026. The six atomic Core Registry obligations remain the executable
conformance authority.}
\label{tab:invariant-crosswalk}
\begin{tabularx}{\textwidth}{@{}P{0.20\textwidth}P{0.22\textwidth}Y@{}}
\toprule
\textbf{Reference Model} & \textbf{PAC-2026 family} &
\textbf{Change in normative meaning} \\
\midrule
\RMInv{1} Evidence-bound outputs &
\PAInv{1} Evidential closure &
Replaces the universal ``at least one item'' rule with policy-declared typed
relations and explicit open/rejected surface semantics. \\
\RMInv{2} Artifact-mandatory runs &
\PAInv{2} Run/artifact and measurement closure &
Narrows the gate to profile-required runs and binds the candidate, profile,
policy, horizon, run identity and state, exact target, authoritative artifact
digests, and typed absence; the Core Registry separates run/artifact closure
from measurement disclosure. \\
\RMInv{3} Audit-safe traceability &
\PAInv{3} Publication lineage and lifecycle &
Adds canonical candidate identity, immutable publication states, and
append-only challenge, resolution, correction, supersession, and withdrawal
links; the Core Registry evaluates lifecycle continuity independently. \\
\RMInv{4} Oversight-preserving control &
\PAInv{4} Revision-exact authorization &
Replaces generic HITL presence with actor-resolved authority over the exact
candidate digest, scope, limitations, intent, evidence horizon, and
verification time. \\
\RMInv{5} Governance separation &
\PAInv{5} Policy/surface correspondence &
Replaces a prescribed layering principle with verifier-visible correspondence
between the evaluated policy, authoritative state, and registered surfaces. \\
\bottomrule
\end{tabularx}
\end{SpecHeroTable}

\begin{SpecHeroTable}
\caption{Illustrative obligation-removal scenarios for the six PAC-2026 Core
coordinates. Each row explains the responsibility lost without the named
check; it does not demonstrate that the other five coordinates remain
conformant. Some omissions are already excluded by schema or profile admission.
These scenarios are design rationale, not executed single-coordinate
ablations or a proof of universal minimality.}
\label{tab:coordinate-necessity}
\begin{tabularx}{\textwidth}{@{}P{0.11\textwidth}P{0.37\textwidth}Y@{}}
\toprule
\textbf{Coordinate} & \textbf{Failure addressed by its responsibility} &
\textbf{Distinction from adjacent responsibilities} \\
\midrule
\(E\) Evidence closure &
The exact claim revision reaches publication with a required source relation
missing or bound to the wrong locus. &
Run reproducibility, disclosure, approval, a corresponding surface, and an
ordered lifecycle do not establish evidential support for the claim. \\

\(R\) Run--artifact closure &
A claim that requires an analysis reaches publication although the named run,
its target, or its authoritative artifact is absent or substituted. &
Evidence presence and human approval do not establish that the declared
run targeted this candidate or produced the cited artifact. \\

\(M\) Measurement disclosure &
A required measurement lacks an explicit state or the supporting disclosure
witness. &
Evidence and run closure do not supply the measurement declaration. S1 checks
correspondence with that declaration, not the adequacy of its support. The
unresolved measured-evidence case HC-0024 is discussed separately; it is not
an established removal counterexample. \\

\(A\) Authorization &
After a material change from candidate \(c_1\) to \(c_2\), the still-valid
signature on the decision for \(c_1\) is reused to publish \(c_2\). &
Closure, measurement, surface, and lifecycle results for \(c_2\) do not show
that an authorized human accepted that exact target and context. \\

\(S\) Surface correspondence &
The record is internally conforming while the registered reader surface omits
a limitation, hides non-measurement, or presents a stronger disposition. &
Correct records and signatures do not establish what a reader can recover
from the rendered publication. \\

\(L\) Lifecycle continuity &
A challenge, correction, supersession, or withdrawal is dropped or reordered,
so a formerly valid publication remains presented as current. &
The pre-publication state cannot establish the horizon-dependent current
state or preserve a later challenge and its resolution. \\
\bottomrule
\end{tabularx}
\end{SpecHeroTable}
 \begin{SpecHeroTable}
\caption{Protocol risk synthesis for the fixed PAC-2026 Core Profile. The
response column states only what the Core can enforce; the residual column
keeps deployment and scientific risks explicit.}
\label{tab:protocol-risk-synthesis}
\begin{tabularx}{\textwidth}{@{}P{0.22\textwidth}P{0.36\textwidth}Y@{}}
\toprule
\textbf{Threat or fault} & \textbf{Protocol response} &
\textbf{Residual risk or nonclaim} \\
\midrule
Omission, substitution, replay, or stale authority &
Typed exact-target identities, fresh authorization, complete result maps,
and affine permit consumption fail closed &
False input accepted by a trusted authority or checker; collusion; common-mode
error \\

Cross-coordinate promotion or missing measurement &
Coordinate-local witnesses, separate binding gates, and typed Honest Absence &
No truth, scientific-validity, or completeness guarantee beyond the admitted
profile \\

Reader-surface drift or lifecycle rewriting &
Registered renderer/extractor correspondence and authenticated ordered event
prefixes at a declared horizon &
Unregistered surfaces, non-executing observers, unrestricted language
equivalence, and reader comprehension \\

Disclosure, linkability, or denial of service &
The Core claims no prevention; if required current status cannot be recovered,
verification fails closed &
Encryption, access control, selective disclosure, retention, availability,
and jurisdiction-specific compliance remain deployment duties \\
\bottomrule
\end{tabularx}
\end{SpecHeroTable}

\begin{SpecHeroTable}[!b]
\caption{Observer-local proof obligations and their boundaries. Each row
identifies a record transition, its divergence class, an observable
falsification witness, and a stronger claim that the obligation does not
establish.}
\label{tab:lifecycle-proof-obligations}
\begin{tabularx}{\textwidth}{@{}P{0.16\textwidth}P{0.24\textwidth}Y Y@{}}
\toprule
\textbf{Observer locus} & \textbf{Record divergence} &
\textbf{Falsification witness} & \textbf{Not established} \\
\midrule
Production & Declared process differs from executed process &
Missing contract artifact, bypassed policy gate, or unbound authorization &
Correctness of the policy or factual truth of the evidence \\
\addlinespace
Surface & Produced record differs from presented disclosure &
Required state omitted, renamed, transformed, or rendered as a reassuring
default &
Reader comprehension or appropriate downstream reliance \\
\addlinespace
Post-publication audit & Released record differs from an externally
reconstructed record &
Hash, signature, schema, chain, or replay mismatch outside the originating
system &
Source authenticity, signing-authority legitimacy, or claim truth \\
\bottomrule
\end{tabularx}
\end{SpecHeroTable}
 \FloatBarrier

\section{Finite-Model and Fault-Injection Detail}
\label{sup:models}

The ten finite models factor the fixed-profile obligations into deliberately
bounded state spaces. Safe and unsafe configurations are separate runs: the
reported unsafe count is the number of expected violations or declared blind
spots, not an addition to the safe reachable-state total. Exhaustion therefore
supports only the encoded configurations and invariants.

\begin{SpecHeroTable}
\caption{Finite-model evidence and actual scope. The source layer identifies the immutable module used in the SF-2/SF-3/SF-4 overlay; a rebound module may retain an earlier safe configuration. Six modules are inherited unchanged from SF-2 and do not test SF-4 verification-time semantics. Reachable states belong to the named safe configuration; unsafe counts are separate specification-mutation runs, each yielding its expected violation or declared observer blind spot.}
\label{tab:tlc-model-inventory}
\begin{tabularx}{\textwidth}{@{}P{0.29\textwidth}P{0.45\textwidth}P{0.11\textwidth}Y@{}}
\toprule
\textbf{Model, source layer, safe configuration} & \textbf{Finite domains and checked abstraction} & \textbf{Reachable states} & \textbf{Unsafe configurations} \\
\midrule
\texttt{\seqsplit{AuthorizationLifecycle}}\newline SF-2 inherited; \newline \texttt{AL\_\allowbreak SAFE.cfg} & 2 digests, 2 publication IDs; exact authorization and invalidation; no verification-time variable. & 160 & 3 \\
\texttt{\seqsplit{DisclosureSurface}}\newline SF-2 inherited; \newline \texttt{DS\_\allowbreak SAFE.cfg} & 2 digests, 1 surface, 3 axes; disclosure and surface correspondence; no verification-time variable. & 46,305 & 4 \\
\texttt{\seqsplit{CorrectionLineage}}\newline SF-2 inherited; \newline \texttt{CL\_\allowbreak SAFE.cfg} & 2 publication IDs, 1 challenge; correction linkage; no verification-time variable. & 18 & 1 \\
\texttt{\seqsplit{WitnessOrCounterexample}}\newline SF-2 inherited; \newline \texttt{WOC\_\allowbreak SAFE.cfg} & 2 candidates, 2 profiles, 2 horizons; 5 obligations on 4 abstract coordinates; no verification-time variable. & 33,210 & 17 \\
\texttt{\seqsplit{CoordinatePromotion}}\newline SF-2 inherited; \newline \texttt{CP\_\allowbreak SAFE.cfg} & 6 coordinates with explicit finite state/edge sets; Boolean witness freshness; no time variable. & 679 & 3 \\
\texttt{\seqsplit{ProfileAdmissionPublish}}\newline SF-4 rebound; \newline \texttt{PAP\_\allowbreak SAFE.cfg} & 2 candidates, 2 nonces (expected N1), 6 obligations, times 0--7; separate horizon/verification time and permit/publication checks. & 21,879 & 18 \\
\texttt{\seqsplit{BindingTemporalRefinement}}\newline SF-4 rebound; \newline \texttt{BTR\_\allowbreak SAFE.cfg} & 2 candidates, times 0--7, trace length at most 10; separate time freshness; publication-window checks remain in ATC/PAP. & 8,369 & 8 \\
\texttt{\seqsplit{TripleFalsifiability}}\newline SF-2 inherited; \newline \texttt{TRIPLE\_\allowbreak SAFE.cfg} & 3 observers, 2 values, 3 characteristic single-locus mutations; shared-fault controls; no time variable. & 4 & 9 \\
\texttt{\seqsplit{AuthorizationTargetSeparation}}\newline SF-4 rebound; \newline \texttt{ATS\_\allowbreak SAFE.cfg} & 2 candidates, 3 relation states, times 0--7; SF-3 target partition with the SF-4 causal guard. & 34 & 4 \\
\texttt{\seqsplit{AuthorizationTemporalCausality}}\newline SF-4 new; \newline \texttt{ATC\_\allowbreak SAFE.cfg} & Times 0--7; initial (horizon, decision, proof, verification, activation, expiry, publication) = (0,1,2,3,4,7,4); scripted boundary cases, not all time tuples. & 106 & 9 \\
\textbf{Total} & Ten factorized finite models & \textbf{110,764} & \textbf{76} \\
\bottomrule
\end{tabularx}
\par\smallskip
\noindent The named files under \texttt{formal/config/} list the exact \texttt{INVARIANT(S)}; the frozen \texttt{formal/run\_tlc\_checks.sh} maps every unsafe configuration to its expected invariant. Model/configuration paths and invariant names are also bound in the paper's evaluation SSOT. These factorized checks do not establish a cross-model refinement proof or unbounded safety.
\end{SpecHeroTable}
 \FloatBarrier

\section{Historical SF-3 Reference-Path Evidence}
\label{sup:sf3}

The following cases retain their SF-3 identities. They show how the private
reference path treated authorization successors and how the corresponding
engineering suite was composed. They are regression evidence for the SF-4
work, not SF-4 conformance cases and not evidence of an independent
implementation.

\begin{SpecHeroTable}
\caption{Historical frozen SF-3 authorization-successor cases (1 of 2). The common \texttt{A1-\allowbreak SF3-\allowbreak } prefix is omitted. Expected observations come from the immutable predecessor vector; the last column states the bounded record/permit consequence tested by the private reference path. These rows are not relabeled as SF-4 cases.}
\label{tab:sf3-cases-a}
\begin{tabularx}{\textwidth}{@{}P{0.16\textwidth}P{0.27\textwidth}P{0.27\textwidth}Y@{}}
\toprule
\textbf{Case suffix} & \textbf{Condition or attack} & \textbf{Frozen normalized observation} & \textbf{Record/permit consequence} \\
\midrule
\texttt{EXACT-\allowbreak CURRENT} & Exact current authorization; relation absent & \texttt{PASS}\, / \,\texttt{A\_\allowbreak AUTHORIZATION\_\allowbreak ACCEPTED} & GATE\_PASS; permit only from the complete fresh six-coordinate all-pass record \\
\texttt{EXACT-\allowbreak HORIZON-\allowbreak REPLAY} & Replay exact-target authorization at another horizon & \texttt{FAIL}\, / \,\texttt{A\_\allowbreak AUTHORIZATION\_\allowbreak STALE}\, / \,\texttt{HORIZON\_\allowbreak CHANGED} & GATE\_BLOCKED; no permit \\
\texttt{MATERIAL-\allowbreak PREDECESSOR} & Authentic materially non-equivalent predecessor & \texttt{FAIL}\, / \,\texttt{A\_\allowbreak AUTHORIZATION\_\allowbreak STALE}\, / \,\texttt{CANDIDATE\_\allowbreak MUTATION} & GATE\_BLOCKED; no permit \\
\texttt{MISSING-\allowbreak RELATION} & Predecessor with no relation assessment & \texttt{CANNOT\_\allowbreak VERIFY}\, / \,\texttt{A\_\allowbreak CANDIDATE\_\allowbreak RELATION\_\allowbreak UNVERIFIABLE} & GATE\_UNVERIFIABLE; no permit \\
\texttt{UNVERIFIABLE-\allowbreak RELATION} & Predecessor with unusable relation evidence & \texttt{CANNOT\_\allowbreak VERIFY}\, / \,\texttt{A\_\allowbreak CANDIDATE\_\allowbreak RELATION\_\allowbreak UNVERIFIABLE} & GATE\_UNVERIFIABLE; no permit \\
\texttt{BASELINE-\allowbreak SUBSTITUTION} & Substitute relation baseline and re-digest & \texttt{FAIL}\, / \,\texttt{A\_\allowbreak TARGET\_\allowbreak MISMATCH} & GATE\_BLOCKED; no permit \\
\texttt{TARGET-\allowbreak SUBSTITUTION} & Substitute relation target and re-digest & \texttt{FAIL}\, / \,\texttt{A\_\allowbreak TARGET\_\allowbreak MISMATCH} & GATE\_BLOCKED; no permit \\
\texttt{TARGET-\allowbreak BODY-\allowbreak SUBSTITUTION} & Substitute target body and re-digest & \texttt{FAIL}\, / \,\texttt{A\_\allowbreak TARGET\_\allowbreak MISMATCH} & GATE\_BLOCKED; no permit \\
\texttt{RULE-\allowbreak CHECKER-\allowbreak SUBSTITUTION} & Replace the registered relation rule or checker & \texttt{INPUT\_\allowbreak INVALID} & No substantive result, gate, valid record, or permit \\
\bottomrule
\end{tabularx}
\end{SpecHeroTable}
 \begin{SpecHeroTable}
\caption{Historical frozen SF-3 authorization-successor cases (2 of 2). The common \texttt{A1-\allowbreak SF3-\allowbreak } prefix is omitted. Expected observations come from the immutable predecessor vector; the last column states the bounded record/permit consequence tested by the private reference path. These rows are not relabeled as SF-4 cases.}
\label{tab:sf3-cases-b}
\begin{tabularx}{\textwidth}{@{}P{0.16\textwidth}P{0.27\textwidth}P{0.27\textwidth}Y@{}}
\toprule
\textbf{Case suffix} & \textbf{Condition or attack} & \textbf{Frozen normalized observation} & \textbf{Record/permit consequence} \\
\midrule
\texttt{ASSESSMENT-\allowbreak DIGEST-\allowbreak FORGERY} & Forge the relation-assessment digest & \texttt{FAIL}\, / \,\texttt{A\_\allowbreak TARGET\_\allowbreak MISMATCH} & GATE\_BLOCKED; no permit \\
\texttt{RELATION-\allowbreak DETAIL-\allowbreak DIGEST-\allowbreak FORGERY} & Forge detail digest and re-digest the assessment & \texttt{FAIL}\, / \,\texttt{A\_\allowbreak TARGET\_\allowbreak MISMATCH} & GATE\_BLOCKED; no permit \\
\texttt{CHANGED-\allowbreak LOCI-\allowbreak SUBSTITUTION} & Replace changed loci and re-digest the relation & \texttt{FAIL}\, / \,\texttt{A\_\allowbreak TARGET\_\allowbreak MISMATCH} & GATE\_BLOCKED; no permit \\
\texttt{RELATION-\allowbreak ON-\allowbreak EXACT-\allowbreak TARGET} & Attach a relation to exact-current authorization & \texttt{FAIL}\, / \,\texttt{A\_\allowbreak TARGET\_\allowbreak MISMATCH} & GATE\_BLOCKED; no permit \\
\texttt{MATERIAL-\allowbreak PROMOTED} & Relabel a material predecessor as PASS & \texttt{VERIFICATION\_\allowbreak RECORD\_\allowbreak INVALID} & Coherent promotion rejected; no permit \\
\texttt{UNVERIFIABLE-\allowbreak PROMOTED} & Relabel unverifiability as PASS or FAIL & \texttt{VERIFICATION\_\allowbreak RECORD\_\allowbreak INVALID} & Both coherent relabels rejected; no permit \\
\texttt{STALE-\allowbreak PERMIT} & Attempt permit derivation from a stale record & \texttt{PERMIT\_\allowbreak INVALID} & Permit rejected; no publication \\
\texttt{C1-\allowbreak PRESENTATION-\allowbreak ONLY} & Change presentation metadata outside candidate closure & \texttt{CANDIDATE\_\allowbreak DIGEST\_\allowbreak UNCHANGED} & Identity non-regression only; no authorization or permit inference \\
\bottomrule
\end{tabularx}
\end{SpecHeroTable}
 \begin{SpecHeroTable}
\caption{Historical SF-3 private-path composition of the 373-node suite at checkpoint \texttt{58944d72}. The same collected suite passed in the recorded source, extracted-sdist, and isolated-wheel runs. This implementation evidence is not the SF-4 conformance corpus; node counts are engineering coverage, not independent scientific properties.}
\label{tab:sf3-test-census}
\begin{tabularx}{\textwidth}{@{}P{0.39\textwidth}Y P{0.10\textwidth}@{}}
\toprule
\textbf{Test group} & \textbf{Bounded scope} & \textbf{Nodes} \\
\midrule
Shared and SF-2 architecture/regression modules & 15 test files & 242 \\
SF-3 A1 classifier & Substantive A1-v2 partition & 20 \\
SF-3 authority admission & Exact private corpus and authority chain & 17 \\
SF-3 candidate relation & Typed non-authorizing relation semantics & 27 \\
SF-3 object admission & K0/H0 identities and schema admission & 21 \\
SF-3 record and permit & Six-coordinate record and affine permit boundary & 11 \\
SF-3 successor conformance & 17 frozen cases plus 18 closure/relabel controls & 35 \\
\textbf{Total} & Exact collected suite & \textbf{373} \\
\bottomrule
\end{tabularx}
\end{SpecHeroTable}
 \FloatBarrier

\section{Historical, Superseded Rival Procedure}
\label{sup:rival}

The rival analysis used a version-pinned corpus closed on 27 July 2026. It
withheld credit for newly written integration rules, and a later source audit
found relevant pre-cut-off omissions. The procedure is retained for historical
traceability, not as current novelty evidence. Its recorded result does not
answer the stronger, still unexecuted composition challenge, which allows a
newly written profile to win. Later sources and rules do not retroactively
change the frozen comparison.

\begin{SpecHeroTable}
\caption{Historical, superseded rival procedure for the 27 July corpus.
Its recorded result is not current novelty evidence and does not decide the
later, stronger composition challenge.}
\label{tab:rival-method}
\begin{tabularx}{\textwidth}{@{}P{0.18\textwidth}P{0.31\textwidth}
P{0.27\textwidth}Y@{}}
\toprule
\textbf{Stage} & \textbf{Procedure} & \textbf{Control} &
\textbf{Bounded output} \\
\midrule
Corpus construction &
Collect exact dated sources that materially specify at least one candidate
function &
Stable source identity; source type and status recorded; search events
retained &
40 exact sources, six aggregate records, 45 search events \\

Source-native reconstruction &
Assign each component its strongest documented semantics before mapping to
AIJIM functions &
No capability is weakened merely to preserve the candidate claim &
Rival-native stack and explicit native non-guarantees \\

Input-restricted second mapping &
Compare the author mapping with the recorded A1 reconstruction generated on
27 July &
Same bound corpus; permitted/prohibited inputs fixed in the manifest;
manuscript and author mapping prohibited; date and agent identity retained &
Function-level concordance with one retained F9 downgrade; model/runtime,
prompt, and sampling were not externally attested or preserved for exact rerun \\

Substitution decision &
Test whether the corpus-documented composition supplies equivalent publication
semantics &
Newly written integration rules were not credited; this restriction is
superseded in the later challenge &
\texttt{NO\_COMPLETE\_SUBSTITUTE\_IN\_BOUND\_CORPUS} \\
\bottomrule
\end{tabularx}
\end{SpecHeroTable}
 \FloatBarrier

\section{Conformance and Refinement Boundary}
\label{sup:conformance}

The classes below decompose responsibilities under the fixed profile. A
conforming class licenses only the stated responsibility. Complete protocol
conformance, independent interoperability, production readiness, and standard
status require additional gates that remain open.

\begin{SpecHeroTable}[!b]
\caption{Conformance classes for the fixed PAC-2026 Core Profile. Each class
is fail-closed and licenses only the responsibility stated in its row.}
\label{tab:conformance-classes}
\begin{tabularx}{\textwidth}{@{}P{0.18\textwidth}P{0.31\textwidth}
P{0.28\textwidth}Y@{}}
\toprule
\textbf{Class} & \textbf{Required contract} & \textbf{Fail-closed condition} &
\textbf{Does not establish} \\
\midrule
Candidate constructor &
Canonical \texttt{SurfaceIntent} and exact
\texttt{PublicationCandidate} under the admitted Core/Profile &
Incomplete closure, unknown type, or identity mismatch &
Evidence quality or publication authority \\

Authorization authority &
Authenticated \texttt{AuthorizationRecord} binding the exact candidate,
scope, trust snapshot, \(h_e\), decision time, and validity interval;
freshly evaluated by A1 at \(t_v\) &
Wrong target, stale/revoked authority, invalid proof, or scope mismatch &
Measurement, truth, or Core verification \\

Verification engine &
One complete ordered \texttt{VerificationRecord} with applicability,
substantive, and promotion-gate results for all six obligations &
Silence, partial/reordered maps, unregistered checker, stale input, or
cross-coordinate witness &
Scientific correctness of arbitrary checkers \\

Surface implementation &
Candidate-bound manifest plus registered renderer/extractor whose Core
projection corresponds exactly &
Omission, promotion, substitution, or unrecoverable required observable &
Unrestricted natural-language equivalence or comprehension \\

Lifecycle custodian &
Authenticated publication and challenge-event prefix with target,
predecessor, evidence-horizon, verification-time, and successor continuity &
Missing/reordered event, broken predecessor, stale current state, or
unverifiable prefix &
Availability or legitimacy of every challenge \\

Technical binding &
Encode/decode mapping that preserves identities, results, effects, and order;
a bundle binding embeds required SF-4 re-admission and shares the atomic Core
publication identity &
Round-trip or trace refinement changes any protected semantic fact, or relies
on a sidecar, dual write, or post-hoc join &
Semantic conformance merely from a signature, receipt, or historical Audit
Bundle 2.4.0 package \\

Complete implementation &
All required classes plus fresh affine permit issuance and atomic
single-consumption publication &
Any class fails, or a permit is partial, stale, replayed, or multiply consumed &
Independent interoperability, field efficacy, or standard status \\
\bottomrule
\end{tabularx}
\end{SpecHeroTable}
 \FloatBarrier

\section{Artifact and Evidence Index}
\label{sup:artifacts}

\begin{SpecHeroTable}
\caption{Complete identifiers of the governed PAC-2026 SF-4 package used by this article. Remaining package- and per-file digests are retained in the governed artifact inventory.}
\label{tab:pac-package-identifiers}
\begin{tabular}{@{}ll@{}}
\toprule
\textbf{Object} & \textbf{Complete SHA-256 identifier} \\
\midrule
Package root & \mbox{\PACPackageRoot{}} \\
Artifact manifest & \mbox{\PACArtifactManifest{}} \\
\bottomrule
\end{tabular}
\end{SpecHeroTable}
 \FloatBarrier

\hypertarget{artifact-index}{}

This index separates the PAC-2026 semantic artifact and implementation
evidence now archived in the public research collection
(\href{https://doi.org/10.5281/zenodo.22727750}{doi:10.5281/zenodo.22727750})
from restricted historical precursor deposits. Identifiers anchor provenance;
public access does not imply public-standard status, implementation
conformance, or continued operation of a live service.

\paragraph{PAC-2026 bounded semantic freeze.}
The originally private semantic-freeze package contains \PACFreezeId{}, version
\PACFreezeVersion{}, disposition \PACFreezeDisposition{}. Its package-root
SHA-256 is
\texttt{\seqsplit{0484c8c9cac04dc02d9288d4c7081e849b4ba7d3ffd0ae6f300199a437cd262a}}.
The effective semantics bind the immutable SF-2 predecessor to the SF-3 delta
closure and then to the immutable SF-4 successor snapshot; neither historical
freeze is relabeled. The final SF-4 package contains 35 artifacts totaling
681,815 bytes and a self-contained verifier for exact path, byte-length,
digest, file-set, predecessor, and symlink closure. These are SF-4 package
counts, not a relabeling of either predecessor. The Core Registry typed digest
is \PACCoreRegistryDigest{}; the Core Profile typed digest is
\PACCoreProfileDigest{}.
Its original custody classification is preserved in the frozen record; the
unchanged package bytes are now included in the public collection. This does
not make the freeze an accepted public standard or authorize a production rollout.

\paragraph{Executable protocol corpus.}
The effective corpus contains the bootstrap, fixed Core Registry and Profile,
36 executable schemas, 53 typed reason codes, Unicode and canonical digest
vectors, the Python registry validator, and a separately implemented Node.js
digest reconstruction. The registered identities include the exact
authorization-statement projection and the sole
\texttt{publicationTargetDigest} lifecycle context. It also contains the
target-separated authorization evaluation objects, the SF-4
evidence-horizon/verification-time separation, 23 temporal vectors, ten TLA+
models, all 76 expected unsafe configurations, the pinned toolchain report,
and the exact claim--evidence--falsifier ledger. The model report records
110,764 distinct reachable states
across the safe configurations and the expected violation for each unsafe
configuration.

\paragraph{Locations of the three illustrative controls.}
Within the SF-4 snapshot, \nolinkurl{formal/validate_pac_registry.py} contains
the control labeled ``Permit with a valid substitute surface outside the
VerificationRecord'' and its adjacent lifecycle-plan control. Both first
validate the replacement locally and then require rejection at permit binding.
The same file's \nolinkurl{validate_measurement_schema} function includes
``NOT\_MEASURED without witness'' and ``CANNOT\_VERIFY without unverifiability
witness''; \nolinkurl{expect_invalid} requires a schema error, and the main
validator invokes this function. These controls address required disclosure
fields, not HC-0024's unresolved \texttt{MEASURED}-branch evidence resolution.
The predecessor fixture is
\nolinkurl{protocol/successors/PAC-2026-SF-4/SF3_TEMPORAL_COUNTEREXAMPLE_FIXTURE.json},
with identifier
\nolinkurl{PAC-2026-SF-3-TEMPORAL-CAUSALITY-COUNTEREXAMPLE-1}; the freeze-level
\nolinkurl{VALIDATION_RESULTS.md} records SF-3 publication and the rebound
SF-4 \texttt{FAIL/A\_TEMPORAL\_ORDER\_INVALID} result. These are selected
explanations of existing controls, not new cases or additional observations.

\paragraph{Historical SF-3 reference path.}
The implementation evidence is bound to semantic-code checkpoint
\texttt{\seqsplit{f62e65459617030716d8906a89919234478ad1ff}} and final
bounded-review checkpoint
\texttt{\seqsplit{58944d727ff9b2f228c6114bb4b95af8d084f291}}.
Its exact 4,202-byte authorization-successor vector has SHA-256
\texttt{\seqsplit{c046125a4a30edf161751084c4384ff18a112336cf6e3cff06f90e70351ff9db}}.
All 17 cases produced their expected normalized outcomes; recorded 373-test
suites passed from source, an extracted source distribution, and an isolated
installed wheel. The maximum disposition is
\nolinkurl{REFERENCE_IMPLEMENTATION_TEST_PROVEN_LOCAL_BOUNDED}. The branch
was private and unmerged at that checkpoint, with SF-2 as the public/default
package surface. Its evaluated snapshot is now archived; the historical
evidence establishes neither complete conformance, reproducible build,
general or production refinement, independent implementation,
interoperability, certification, nor production release.
The rendered case, finite-model, and test-census tables are generated from
\texttt{paper/ssot/evaluation\_evidence.v1.json}. The generator rejects source
hash drift and incomplete case or count closure; this is article-projection
QA, not independent evidence or semantic authority.

\paragraph{Post-freeze binding and cross-path evidence.}
The bounded binding-refinement package records source baseline
\texttt{\seqsplit{5ea3f2312de847cdc1aa3acb28ae7c7978471c7f}}, Core version
0.5.0a4, two registered successor bindings, a frozen 12-case corpus, and 24
unchanged result cells. The 22 negative cells were rejected before permit
derivation, with no false \texttt{PublicationPermit} or
\texttt{PublicationRecord} acceptance. A disclosed three-field correction to
the auxiliary result schema changed no result cell and required no rerun. Its
evidence commit is
\texttt{\seqsplit{869f4c642743117e082d409be51b8ef703e68a38}}, its closure commit is
\texttt{\seqsplit{6c9694418aed2697a1e41ea03ca3a52f2d972c4d}}, and the result
SHA-256 is
\texttt{\seqsplit{8ed0cfa4a4308ef2b13c19b7527f4bb8d61918fbdab5e69ddf9684b638d770d4}}.
Its disposition is \texttt{PASS\_BOUNDED} for that exact boundary.

The developer-calibration package crosses two producer and two verifier paths
over 23 public cases, 92 declared cell projections, and 161 recorded role
invocations, with no run- or case-level error. The bound bridge commit is
\texttt{\seqsplit{67306dde71ebfd139777caa1ac1c6fedb4613f6a}}. Its run is
\nolinkurl{PAC-2026-SF-4-G7-V6-PUBLIC-CROSS-RUNNER-SUCCESSOR-02}; the result
record has SHA-256
\texttt{\seqsplit{f8980bc5223429fbc80677577a34c69cc33ca41411d46f6296fb14b4dc048593}}.
The 37,945,599-byte archive has SHA-256
\texttt{\seqsplit{94a4051850d91e12fc7007e05fcedd7af783b5198014349b416ae0684518207f}},
and all 923 internal inventory entries passed extraction readback. Because the
corpus and expected projections were visible during development, this is
cross-path calibration rather than an interoperability claim or fresh blind
confirmation. Both unchanged packages are included in the public collection.

\paragraph{G7-B: evaluated builds, blindness, and retained findings.}
The completed 12 September study binds apparatus-freeze commit
\texttt{\seqsplit{aad6a9642bb1ed8c562ba76f245d08259a3f1daa}} and the later
packaging commit
\texttt{\seqsplit{f39ecb344ff65f78b0e823cd4d4d17048b3c41a0}}. The unchanged
delivered archive, \texttt{g7b-evidence-f39ecb344ff6.tar.gz}, has SHA-256
\texttt{\seqsplit{09d6c4061879e482d7cf8ea8766c8d1810f728f8508ed4fd077ea837a3fa49fe}}.
The registered semantic target is PAC-2026 SF-4 \texttt{0.4.0-sf.4}.
The evaluated reference producer/verifier is A12 at \texttt{024f14c}; the
Node producer is \texttt{1e60967}; the Rust checker is 1.1.7 at
\texttt{d686065}. The bridge \texttt{cb9d586} is apparatus, not another
implementation. Full registration, exact vendored bytes, integration history,
raw role outputs, and results are retained in the sub-package.

The original study package does not contain a complete upstream development
history. Two later, separately bound provenance handoffs recover the following
source chains: reference A4 \texttt{160de3d4} to matching import
\texttt{4bc3816c} and seven integration commits ending at A12
\texttt{024f14c}; Node \texttt{56a4dfc9} to byte-identical import
\texttt{9029d572} and 11 commits ending at \texttt{1e60967}; and Rust
\texttt{c3184016} to byte-identical import \texttt{e8b93a20} and 12 commits
ending at \texttt{d686065}. Full identities, archive digests, and the two
source-report digests are bound in
\nolinkurl{paper/ssot/evaluation_evidence.v1.json}, under
\texttt{postEvaluationPackage.implementationProvenance}. Only the provenance
addendum of the second report supports this account; its earlier proposals
are not study results or new submission requirements.

Implementation A contains its Python semantic Core and delegation-only
adapters. Implementation B contains local Node producer and Rust verifier
semantics, with no declared reference-runtime dependency in the inspected
manifests. The registration classifies these as independently developed
lines. The recovered archives and import histories document distinct
pre-integration source snapshots; they do not supply a complete original
developer/component assignment, information-exchange history, or code-copy
audit. These limits neither prove dependence nor establish independently
conducted evaluation. The new handoffs do not alter the frozen inputs or
results.

The original record also documents subsequent integration: one AI coding
agent changed all three evaluated builds after import, partly against the reference and known
expected results. A second AI coding agent built the G7-B apparatus in the
same integration environment. These are tool roles, not additional human
reviewers. The case archetypes, expectation rules, parameter domains, and
earlier corpora were known. Each qualification run comprised three index-seeded
sets. Its original \nolinkurl{endpoints.json} reported 192 scored instances
and \(N_{\mathrm{new}}=9\). After the F-01 lifecycle trio was withdrawn,
the pre-freeze \nolinkurl{endpoints.final-registration.json} reported
183 scored instances and \(N_{\mathrm{new}}=0\), with unchanged generation,
journal, and per-set analysis bindings. Both runs already showed the final
219/240 four-way agreement and 219/219 portable agreement. The apparatus
operator saw these outcomes; registered signatures came from all six
qualification sets. Only the final instance values were drawn after the
freeze. Instance blindness alone establishes neither implementation nor
evaluation independence.
The older phrase ``common-origin
lineage'' in the frozen aggregator is preserved as historical wording, not
used as a finding about development origin.

Four RFC~3161 tokens bind the freeze statement before drand round 6459109
(12 September 2026, 09:11:30 UTC); the latest token precedes the round by
316 seconds. Seed derivation uses the round randomness, not the freeze
commit. The complementary OpenTimestamps Bitcoin anchor is later than the
round and therefore does not establish pre-round existence. The three sets
contain 240 instances, of which 183 were scored. All 75 expected acceptances
and 108 typed expected rejections matched in all four pairings:
\(N_{\mathrm{new}}=0\), \texttt{G7-B\_CONSISTENT}. Four-way complete agreement
held for 219/240 instances, and portable outputs agreed for those 219/219.
The 240-instance replay count is not the scored denominator. There are
61 scored archetypes (25 acceptance and 36 rejection), 17 withdrawn, and two
specification ambiguities; three instances per archetype give 183 scored
and 57 unscored observations. All 21 instances without four-way agreement
belong to seven withdrawn archetypes: HC-0033, HC-0034, HC-0044,
HC-0073--HC-0075, and HC-0080. The other 36 unscored instances show agreement
but do not support a conformance inference.

``Acceptance'' uses \nolinkurl{tools/analyze.py}'s operational predicate:
66 \texttt{VERIFICATION\_RECORD\_VALID} cases with A1 \texttt{PASS}, three
\texttt{EQUALITY\_BOUNDARIES\_ACCEPTED} cases, and six
\texttt{ATOMIC\_PUBLICATION\_ACCEPTED} cases. Fifteen of the 66 retain a
localized negative or indeterminate result: E1 \texttt{FAIL} (HC-0018), E1
\texttt{CANNOT\_VERIFY} (HC-0019), R1 \texttt{CANNOT\_VERIFY} (HC-0022), M1
\texttt{CANNOT\_VERIFY} (HC-0025), and S1 \texttt{FAIL} (HC-0030), each in
three sets. Thus 75 is neither an all-coordinate-pass count nor a count of
publication permissions.

Three quantities answer different questions. The raw analyzer reports
210/240 complete matches across its applicable checks; four-way equality
alone holds for 219/240; the registered endpoint matches 183/183 scored
instances. HC-0021 is withdrawn, HC-0024 is ambiguous, and HC-0027 is scored
with its context component excluded. The aggregator applies those registered
dispositions; the raw match count is not an alternative scored denominator
or an additional sensitivity result.

Two qualifications apply within the scored set. Seven schema/profile
archetypes (HC-0009, HC-0010, HC-0012--HC-0016; 21 instances) use the admission
baseline with case-derived identifiers: they are positive confirmations, not
distinct fault challenges. Thirteen archetypes have terminal-only A1 targets;
HC-0066 is withdrawn, leaving 12 scored archetypes and 36 scored instances.
For both Rust-verifier pairings in those instances, the bridge supplies the
terminal classification from an operation table. The native Rust execution
class and localizations are retained, but the compared terminal is not
independent Rust evidence for the target. This coupling is registered as
\texttt{RUST\_TERMINAL\_FROM\_BRIDGE\_OPERATION\_TABLE} in
\nolinkurl{docs/EXPECTATION_SOURCES.md} and \nolinkurl{tools/analyze.py}.
These descriptions do not rescore the registered endpoint.

The pre-freeze disposition history is mixed. An outcome-blind fixture audit
identified nine of the 17 withdrawn archetypes; the other eight had
mismatch-triggered repair or adjudication histories. The repaired lifecycle
trio was ultimately withdrawn after an outcome-blind admissibility review.
Six repaired archetypes remained scored. The records preserve both the
qualification exposure and these review boundaries rather than describing
all decisions as outcome-blind. A concrete example is HC-0033: its initial
match arose from an unintended event-time violation while the declared event
swap was absent. Its apparent confirmation was withdrawn. No disposition or
expectation was changed after the evidence seed was revealed.

The source record retains 17 withdrawn archetypes, six specification
clarification findings, and five registered observations reproduced across
all three sets. HC-0024 and HC-0062 are specification ambiguities, not scored
confirmations. F-01 records divergence on inadmissible input in HC-0033,
HC-0034, and HC-0080. HC-0027 was scored without its context component.
Six canonicalization properties and replay/collision rejection cannot be
carried by the evaluated interchange; published-phase lifecycle cases cannot
be constructed under the single-evaluation schedule. Anchors outside
2001--2199 were not drawn. Some lifecycle outputs failed harness conversion
without a retained typed observation; these are apparatus limitations, never
successful semantic rejections. The observation class was derived from the
adapter result type rather than the addendum's prescribed method; that
deviation is reported, not scored. The files
\nolinkurl{docs/RESULT.md},
\nolinkurl{docs/IMPLEMENTATIONS_AND_BLINDNESS.md}, and
\nolinkurl{docs/ADJUDICATION_NONEVIDENCE.md} retain the complete allocations.

\paragraph{Consequences of the two specification ambiguities.}
\label{sup:g7b-ambiguities}
HC-0024 concerns what counts as target-bound measurement evidence. Its
\texttt{MEASURED} entry carries a syntactically valid digest that differs from
the candidate's measurement-disclosure commitment and resolves to no carried
measurement bytes. The archived adjudication identifies three readings:
disclosure binding alone permits M1 \texttt{PASS}; requiring the candidate
commitment yields \texttt{FAIL}; requiring resolved evidence bytes also calls
the baseline expectations into question. SF-4 does not operationally settle
that commitment or resolution requirement. All four pairings returned M1
\texttt{PASS} and \texttt{VERIFICATION\_RECORD\_VALID}. This is a substantive
uncertainty about M1 acceptance, not merely diagnostic wording. Since permit
derivation requires every hard obligation to pass, the interpretation can
affect which records qualify for derivation; this evaluation case records
neither permit issuance nor publication.

HC-0062 concerns the authoritative verification time and rejection layer.
The A1 input's time is changed and lower-level digests are rebuilt, while
other evaluation-time references remain unchanged. SF-4 requires substitution
to be detected at the next bound layer, but the interchange does not
independently name the authoritative expected time. The archived assessment
therefore considers rejection at record assembly and adoption of the A1 time
as competing readings; it does not entail the expected A1 rejection at that
particular layer. All four pairings actually rejected with
\texttt{AUTHORIZATION\_EVALUATION\_INPUT\_REJECTED} and
\texttt{FAIL/A\_TARGET\_MISMATCH}. The ambiguity concerns admission and
rejection localization, with no observed incorrectly issued permit.
Both findings remain unscored and unresolved. Their full arguments are in
\nolinkurl{docs/ADJUDICATION_NONEVIDENCE.md} (A-07),
\nolinkurl{docs/adjudication/spec-only-adjudication-2026-09-12.json}
(HC-0024), and
\nolinkurl{docs/adjudication/fixture-audit-2026-09-12/H.json}
(HC-0062), under the archived G7-B source directory. Explaining them here
neither changes SF-4 nor decides a successor rule.

\paragraph{Combined CSI evidence package and completed local reproduction.}
The versioned internal package
\nolinkurl{CSI_EVIDENCE_PACKAGE_2026-09-12_RC1.zip}
contains 206,236,847 bytes and 2,951 manifest-listed files. Its SHA-256 is
\texttt{\seqsplit{cdf6684b71cbe8abc6b24cab48e772e072713db72ed0345e2b153105b968cd32}};
the package-manifest SHA-256 is
\texttt{\seqsplit{c86e28b489aae3726fd25f24244434a0e8b65b363b96d623d673a52d2af9df53}}.
Its paper-source snapshot is the pre-integration commit
\texttt{\seqsplit{2a4172214725c5cbdb7c4c56c83a2156657498fa}}, not the revised
manuscript containing this paragraph. The accompanying acceptance record
\nolinkurl{CSI_EVIDENCE_ACCEPTANCE_2026-09-12_RC1/RESULTS.json}
has SHA-256
\texttt{\seqsplit{2f907abc9aa58bb1f00adc47d03aa3793e3c94929ca709348a2f4494a0b90074}}.
It records \texttt{REPRODUCED}, an empty failure list, and unchanged inventory
before and after execution. The record and raw outputs are separate from the
input ZIP and must accompany any release that cites this reproduction.

After the setup in \texttt{README.md}, select OpenSSL~3 explicitly and a
new results directory outside the extracted package:
\begin{quote}\small\ttfamily\raggedright
.venv/bin/python reproduce.py \textbackslash\newline
\quad \verb|--openssl| /path/to/openssl \textbackslash\newline
\quad \verb|--output| /path/to/new-results
\end{quote}
Replace the two example paths with the installed OpenSSL~3 binary and an
unused output directory. \texttt{CLAIM\_MAP.md} specifies the command's scope.
The completed run used a fresh
extraction, a new Python environment populated from the packaged wheelhouse,
and a cleared inherited environment on the same macOS arm64 host. It used
CPython 3.14.6, Node 26.5.0, and OpenSSL 3.6.3; G7-B used cryptography
45.0.3, while G6's archived reanalysis installed cryptography 50.0.0 in its
separate environment. The raw TLC log records Homebrew Java 21.0.11 and
literally includes both host \texttt{aarch64} and Java \texttt{x86\_64}
strings; no unrecorded binary architecture is inferred from them.
No production service or private source repository
was needed. Runtime was 968.615 seconds, from 10:48:51 to 11:04:59 UTC on
12 September; this is one observed duration, not a performance benchmark.

\begin{SpecHeroTable}
\caption{What the combined package actually reproduced. Re-execution and
archived reanalysis are different evidentiary operations.}
\label{tab:package-reproduction}
\begin{tabularx}{\textwidth}{@{}P{0.19\textwidth}P{0.23\textwidth}Y@{}}
\toprule
\textbf{Area} & \textbf{Operation} & \textbf{Observed result and boundary} \\
\midrule
SF-4 formal & Re-execution & Registry/digest checks; ten models, 110,764 safe
states, 76 exact expected negative controls; bounded configurations only. \\
Historical SF-3 & Re-execution & 373/373 tests in source, fresh sdist, and
installed wheel; not recovery of historical distribution bytes. \\
Rival corpus & Archived reanalysis & 40 source records and 45 search events
validated; no new search or reconstruction judgment. \\
SF-5 G6 & Archived reanalysis & 24 cells, two positive and 22 negative, no
false permits in the stored result; no new live execution. \\
Disclosed calibration & Archived reanalysis & 92 stored cells for 23 cases
match their declarations; no new producer/verifier execution. \\
G7-B & Re-execution & 240/240 archived observations identical, with unchanged
endpoints and signatures; scored denominator remains 183. \\
\bottomrule
\end{tabularx}
\end{SpecHeroTable}

G7-B passed all eight required reproduction checks. The optional online
beacon re-fetch alone was skipped in offline mode; stored beacon and time
proofs were checked. Fifty focused packaging tests also passed in the source
and freshly extracted environments; they test the packaging apparatus, not
50 additional protocol properties. The Rust checker is the evaluated Apple
Silicon binary: source-to-binary reproducibility and other platforms remain
unestablished. This same-host reproduction is neither an external replication
nor another blind study. The restricted precursor deposits below were not
included or rerun.

RC2 and RC3 are internal documentation successors, not new experiments.
The RC3 reader-companion checkpoint,
\nolinkurl{aijim-publication-authority-evidence-2026-09-12-rc3.zip}, has
SHA-256
\texttt{\seqsplit{b2d9be564b295b74e2ef1aed54906c69aeac5e5292cbe5144fcd8b7426e248eb}}.
Its 2,970-file manifest binds claim-to-file/result-key instructions, unchanged
RC1 result copies, and an inventory of the separately required raw-output
companion (10,094 files and 12 symlinks). Sixty packaging tests passed from
source and fresh extraction, but no full reproduction was performed for RC2
or RC3. The subsequently prepared RC4 documentation successor is identified
separately in the release-collection metadata; its packaging checks are not a
new full-study reproduction. The six-stage result belongs only to RC1 and its
acceptance record.

\paragraph{Public research collection, version 0.1.0.}
The 12 September release is archived under the version-specific DOI
\href{https://doi.org/10.5281/zenodo.22727750}{10.5281/zenodo.22727750}.
The 592,538,269-byte collection ZIP has SHA-256
\texttt{\seqsplit{03c7ccd4f0eade1b22ef27d17d4239ce24423bbf94d1d672714c231a579ff8d9}};
its collection manifest has SHA-256
\texttt{\seqsplit{f2c254475df641ea652a21badfed4fc7f3b4fa249869a44788a339b552a45b9a}}.
The source companion at
\url{https://github.com/src01001100/aijim-publication-authority}, tag
\texttt{v0.1.0}, binds commit
\texttt{\seqsplit{e0ae8013295f7574bf492a0c288d521357e28d8f}}.
The archive preserves RC1, RC4, original outputs and provenance reports.
An added Rust source snapshot does not establish source-to-binary reproduction.
Its article checkpoint \texttt{7f34e131} predates this DOI insertion; original
records and build-time reservation labels remain unchanged.

\paragraph{Public correction addendum, version 0.2.0.}
The separately versioned SF-4 Correction Research Addendum is archived at
\href{https://doi.org/10.5281/zenodo.22763325}{10.5281/zenodo.22763325}.
Section~\ref{sup:sf4-correction} identifies its research-candidate scope and
reproduction procedure; the original version 0.1.0 evidence remains unchanged.

\paragraph{Strongest-rival record.}
The original version 0.1.0 package contains the exact 40-source corpus, six aggregate source records,
45 search events, the author-constructed source-native rival, the
function-by-function mapping, the model-assisted input-restricted
reconstruction, and the
retained F9 disagreement. These files support only the
\texttt{NO\_COMPLETE\_SUBSTITUTE\_IN\_BOUND\_CORPUS} result. They are not a
systematic or human-independent absence proof.

\paragraph{Finalized lifecycle case (SF-P0-01).}
The reviewer material contains the 37-event, 76-file evidence record and its
self-contained verification procedure. The author build identifies the
publication as \FinalPublicationIdentityAnchor{}; the blind build anonymizes
it. Runtime manifest, evidence manifest, checksum inventory, and final Audit
Bundle bind the reported source revision, build, image, publication, and file
identities. No live Core, database, credential, or private signing key is
required for the packaged checks. The governed package is retained in private
research custody and is not distributed with this preprint; the publication
identifier is not linked to the unrelated Flow-A DOI.

\paragraph{Matched three-locus experiment (SF-P0-02).}
The reviewer material contains the frozen clean control, Producer, Surface,
and Audit mutations, observer inputs and outputs, unchanged-locus proofs, 19
hash-chained events, 98 checksummed files, and the preregistration. In the
author build, the source snapshot is bound to
\MatchedExperimentCommit{}; the blind build anonymizes it.

\paragraph{Historical separate-consumer evidence.}
The finalized \FlowAExperimentLabel{} evidence package is identified by
\mbox{\FlowAEvidenceAnchor{}}, its version-matched offline verifier by
\mbox{\FlowAVerifierAnchor{}}, and its checksummed reviewer-input package by
\mbox{\FlowAReviewerInputAnchor{}}. These deposits were publicly accessible for the
historical consumer experiment and are currently \texttt{RESTRICTED}. The
DOIs preserve their historical identities but do not represent present public
access. The files do not constitute a PAC-2026 implementation or an
independently developed verifier lineage.

\paragraph{Frozen precursor implementation.}
The anonymized review artifact contains the \emph{RM-2025} implementation
snapshot, historical conformance vectors, expected verifier outcomes, and a
SHA-256 manifest. Its implementation snapshot binds the evidence to
\BlindableImplementationSnapshotCommit{}. Historical records must be replayed
with their archived version-matched verifier, dependencies, trust material,
and corpus. They are not rewritten into PAC-2026 records.

\paragraph{Availability boundary.}
Collection documentation identifies sources, environments, instructions,
dependencies, digests, and component-specific reuse terms. Manuscripts are
supplied for inspection, not blanket-relicensed. RM-2025/SF-P0 material and
the three Flow-A deposits remain restricted and excluded. Neither publication
nor anonymous download closes G6--G8 or strengthens experimental results.
 
\section{Bounded Correction Candidate After the SF-4 Evaluation}
\label{sup:sf4-correction}

\paragraph{Object and access boundary.}
The separately examined candidate
\nolinkurl{PAC-2026-SF-4-CORRECTION}, semantic version
\nolinkurl{0.4.1-correction-candidate}, is a research proposal, not a ratified
freeze or current Core implementation. Its source is
\nolinkurl{protocol/successors/PAC-2026-SF-4-CORRECTION/};
\nolinkurl{SEMANTIC_DELTA_CONTRACT.md} states the proposed rules and
\nolinkurl{VALIDATION_RESULTS.md} records the local checks and object
identities; \nolinkurl{RELEASE_IDENTITY.json} binds the research-candidate
version and status. The correction archive is publicly available as the
SF-4 Correction Research Addendum, version 0.2.0
(\href{https://doi.org/10.5281/zenodo.22763325}{doi:10.5281/zenodo.22763325}). It is separate from the
unchanged \nolinkurl{v0.1.0} evidence described above. The evaluated SF-4
objects and their historical adjudications remain unchanged.

\paragraph{Constructive commitment binding.}
For M1 and R1, the candidate defines canonical, policy-complete disclosure and
run--artifact sets. Their typed commitments include profile and policy
identities but exclude the candidate digest, avoiding a circular definition
when the candidate body carries those commitments. M1 validates the shape
and measurement assignment of each measured-evidence body. R1 receives the
same committed disclosure set, checks it against the candidate, and derives
each measured entry's producer from its bound evidence body. The producer
must be a declared \texttt{RAN} run, and the value reference must belong to
that run's committed artifact set. Omitting an entry, substituting its
commitment, or omitting the value reference yields localized failure;
unavailable mandatory evidence remains distinct from a passing result.
This establishes reference membership, not availability of the referenced
bytes: consistently bound references can pass without supplied value bytes.

\paragraph{Time and adjacent requirements.}
The candidate explicitly binds the evaluation's time-bearing objects to one
parsed instant; equivalent UTC offsets denote the same instant, while a
substituted instant is rejected at the bound evaluation layer. Later
evaluations retain their own contexts. It also makes the selected context
encoding and lifecycle-payload requirements executable. These corrections
preserve the distinction between HC-0024's acceptance-relevant M1 ambiguity
and HC-0062's authoritative-time and rejection-localization ambiguity.
Neither original case demonstrated an incorrectly issued permit.

\paragraph{Evidence and reproduction.}
The local validator completed 49 declared correction cases, including 25
mutated cases, alongside the retained 17 SF-3 regression and 23 SF-4 temporal
cases. This is additional local specification evidence, not a regenerated
SF-4 corpus, a new blind experiment, or a re-scoring of G7-B's 183 scored
instances among 240 generated instances. The original finite-model totals
and G7-B methodological qualifications remain unchanged. From an extracted
correction archive containing the unchanged predecessor freezes,
reconstruction and validation use:
\begin{quote}
\footnotesize
\texttt{cd protocol/successors/PAC-2026-SF-4-CORRECTION}\\
\texttt{python -B APPLY\_DELTA.py}\\
\texttt{cd candidate/formal}\\
\texttt{python -B validate\_pac\_registry.py}
\end{quote}
The accompanying validation report states the environment and exact file
identities. Porting the correction to Core and evaluating a regenerated
successor corpus have not been completed. E1's commitment preimage and the
priority of coexisting L1 violations (S-03(a)) remain beyond this selected
correction. The result is therefore a bounded, inspectable design repair,
not complete conformance or field validation.
 \section{Exact Claim-Edit Regions and Retained-Core Frontiers}
\label{sup:edit-regions}

This post-SF-4 extension asks which selections of a fixed claim universe
preserve one statistical qualification. It supplies an exact interval test,
a minimum retained core, and a shared calculation for all core sizes when
the upper set is the entire universe. Arithmetic-mean e-closure supplies
the statistical rule
\citep[Sections~4--5.1, Equations~(5), (14)]{XuEtAl2026EClosure}; this extension does not
change the frozen protocol.

\subsection{Statistical qualification and deterministic inputs}
\label{sup:edit-region-definitions}

Let \(H=\{0,\ldots,m-1\}\) index fixed null hypotheses. Let \(E_i\geq0\)
be random variables satisfying \(\mathbb E_P[E_i]\leq1\) under every
distribution \(P\) for which hypothesis \(i\) is true. Write \(e_i\) for
their realised values and fix \(\alpha\in(0,1]\). For nonempty \(S\subseteq H\),
define \(\bar e_S=|S|^{-1}\sum_{i\in S}e_i\) and
\(\mathrm{FDP}_S(R)=|R\cap S|/(|R|\vee1)\). The closure family is
\begin{equation}
 \mathcal C_\alpha(e)=
 \left\{R\subseteq H:\mathrm{FDP}_S(R)\leq\alpha\bar e_S
 \text{ for every }\varnothing\neq S\subseteq H\right\}.
 \label{eq:sup-region-closure}
\end{equation}
The empty report qualifies. If \(H_0\neq\varnothing\) is the actual set of
true nulls, every qualified selection obeys
\(\mathrm{FDP}_{H_0}(R)\leq\alpha\bar e_{H_0}\). Taking expectations
gives the inherited simultaneous expected-FDP guarantee, without requiring
independence between the \(E_i\). If \(H_0\) is empty the loss is zero.
This is not a high-probability bound on realised FDP or a truth probability
for each claim.

The algorithms accept deterministic nonnegative rational values. Their
fixtures check exact computation, not distributional calibration. In the
examples, \(B(R)=\max_{\varnothing\neq S\subseteq H}
\mathrm{FDP}_S(R)/\bar e_S\), with \(0/0=0\), positive-over-zero equal
to infinity, and \(B(\varnothing)=0\). Thus \(B(R)\leq\alpha\) denotes
qualification, not a confidence statement about the observed report.
A negative witness satisfies
\(\alpha|R|\sum_{i\in S}e_i<|R\cap S||S|\): it verifies a violation of
the qualification inequality, rather than satisfying that qualification.

\subsection{Interval qualification and minimum retained core}
\label{sup:edit-region-core}

For \(M\subseteq U\subseteq H\), let
\([M,U]=\{R:M\subseteq R\subseteq U\}\). Partition \(H\) into mandatory
core \(M\), optional set \(O=U\setminus M\), and outside set \(X=H\setminus U\),
with sizes \(h,p,q\). Sort each group in nondecreasing evidence order and
let \(A_r,B_t,C_s\) be its prefix sums; zero-length prefixes are zero.

\paragraph{Proposition 1: exact interval qualification.}
Every report in \([M,U]\) belongs to \(\mathcal C_\alpha(e)\) if and only if
\begin{equation}
 \alpha(h+t)(A_r+B_t+C_s)\geq(r+t)(r+t+s)
 \label{eq:sup-region-prefix}
\end{equation}
for all \(0\leq r\leq h\), \(0\leq t\leq p\), \(0\leq s\leq q\)
with \(r+t>0\).

\paragraph{Proof and negative witness.}
Fix an intersection \(S\) with these group counts. If \(r+t>0\),
\begin{equation}
 \max_{R\in[M,U]}\mathrm{FDP}_S(R)=\frac{r+t}{h+t},
 \qquad R^*=M\cup(S\cap O).
 \label{eq:sup-region-worst-report}
\end{equation}
Adding optional members of \(S\) cannot decrease the ratio; adding other
optional members cannot increase it. At fixed counts, the three prefixes
minimise \(\sum_{i\in S}e_i\). Both extrema are attained together by the
prefix intersection \(S^*\) and \(R^*=M\cup(S^*\cap O)\). Substitution
into Equation~\ref{eq:sup-region-closure} proves necessity and sufficiency.
A failed prefix inequality yields this concrete violating pair. Counts
with \(r+t=0\), including the empty report, have zero loss.

For fixed \((r,t)\), adding outside value \(c\) changes the left-minus-right
side of Equation~\ref{eq:sup-region-prefix} by
\(\alpha(h+t)c-(r+t)\). A minimising outside prefix contains precisely
values strictly below \((r+t)/(\alpha(h+t))\); ties can be omitted.
The threshold is nondecreasing in either \(r\) or \(t\). Choosing the
shorter group dimension as the outer loop yields
\(O((h+1)(p+1)+q\min(h+1,p+1))\) arithmetic work after ordering,
plus partitioning and prefixes. This is a fixed-level decision bound,
not a bound for the uncapped maximum-ratio problem.

\paragraph{Proposition 2: strongest-core reduction and minimality.}
If some size-\(h\) core qualifies \([M,U]\), the \(h\) largest values
in \(U\) also form a qualifying core. The first qualifying size among these
nested strongest cores is therefore cardinality-minimal for this fixed \(U\).

\paragraph{Proof.}
Suppose a qualified report \(R\) contains \(i\) but not \(j\), with
\(e_i\leq e_j\), and replace \(i\) by \(j\) to obtain \(R'\).
For an intersection containing \(j\) but not \(i\), also replace \(j\)
by \(i\) in the intersection. Its cardinality and overlap with the
corresponding report are unchanged, while its evidence sum cannot increase.
The original inequality therefore implies the new one. In the remaining
membership cases the overlap is unchanged or smaller, so \(R'\) qualifies.
Now replace a lower-valued mandatory member by a higher-valued optional
member. Each report of the new interval either still contains the old member
and belonged to the old interval, or is a replacement of an old qualified
report. Repeated exchanges produce a strongest core. Its intervals shrink
as \(h\) increases, proving monotonicity. If \(U\) fails, no core works,
because every interval contains \(U\). Otherwise a positive size-\(h\)
region and, for \(h>0\), a counterexample for the strongest size-\(h-1\)
core establish minimality. Ties may yield several optimal cores.

The retained implementation uses endpoint checks and binary search,
requiring \(O(\log(|U|+1))\) regional calls. This maximises
\(2^{|U|-|M|}\) selectable subsets within one interval, not scientific
importance or all possible safe families. For \(e=(8,1,8)\),
\(\alpha=1/2\), \(U=H\), core \(\{0\}\) admits all four containing
reports, including \(\{0,1\}\) with \(B=1/2\). The empty core fails
because \(B(\{1\})=1\); \(\{2\}\) is another minimum core.
Qualified endpoints alone are insufficient: for \(e=(4,4/5,4)\),
\(B(\{0\})=B(H)=5/12\), but \(B(\{0,1\})=5/8\).

\paragraph{Empty cores and the known FWER implication.}
An empty-core region includes every singleton in its upper set. Consider
a measurable procedure that returns \(\widehat U(E)\) only when
\([\varnothing,\widehat U(E)]\subseteq\mathcal C_\alpha(E)\), and
returns \(\varnothing\) otherwise. Under the stated e-value assumptions,
\[
 \Pr\{\widehat U(E)\cap H_0\neq\varnothing\}
 \leq \mathbb E\!\left[\max_{R\in\mathcal C_\alpha(E)}
                     \mathrm{FDP}_{H_0}(R)\right]\leq\alpha.
\]
A false rejection supplies a singleton with FDP one. This is the known
procedure-level FWER consequence of simultaneous closure
\citep[Section~8.2, Theorem~39]{XuEtAl2026EClosure}, not a new error-control
result and not conditional error control given an observed certificate.
The minimum-core question makes the resulting retention--flexibility
trade-off explicit without changing the inherited guarantee.

\subsection{All-core frontier for the special case \texorpdfstring{\(U=H\)}{U=H}}
\label{sup:edit-region-frontier}

\paragraph{Proposition 3: exact shared frontier.}
When \(U=H\), all strongest size-\(h\) core decisions, \(h=0,\ldots,m\),
can be computed together in \(O(m^2)\) arithmetic work after sorting,
with \(O(m)\) compact workspace and output.

\paragraph{Proof and reconstruction.}
Use one-based sorted positions \(e_1\leq\cdots\leq e_m\),
\(P_j=\sum_{i=1}^j e_i\), and \(P_0=0\); retain original identities
separately. The strongest core occupies the final \(h\) positions.
Put \(p=m-h\). With no outside group, the regional condition becomes
\begin{equation}
 \alpha(h+t)(P_{p+r}-P_p+P_t)\geq(r+t)^2,
 \quad0\leq r\leq h,\quad0\leq t\leq p,\quad r+t>0.
 \label{eq:sup-frontier-original}
\end{equation}
The substitution \(d=p-t\), \(j=p+r\) is bijective onto
\(0\leq d<m\), \(d\leq p\leq m\), \(\max(p,d+1)\leq j\leq m\),
and gives
\begin{equation}
\begin{aligned}
 F_d(j)&=\alpha(m-d)P_j-(j-d)^2,\\
 F_d(j)&\geq\alpha(m-d)(P_p-P_{p-d}).
\end{aligned}
 \label{eq:sup-frontier-transform}
\end{equation}
For each \(d\), scan \(p\) downward from \(m\) to \(d\), maintaining
the minimum \(F_d(j)\) on the specified suffix. Add \(j=p\) only when
\(p>d\), excluding the empty intersection. Comparing that minimum with
the right-hand side checks all original inequalities for this \((d,p)\);
all core sizes are decided in the same scan.

When an inequality fails, retain \((d,j)\) for \(h=m-p\). Put \(t=p-d\),
\(R=M_h\cup\{1,\ldots,t\}\), and
\(S=\{1,\ldots,t\}\cup\{p+1,\ldots,j\}\) in sorted positions.
Then \(|R|=m-d\), \(|S|=j-d>0\), and \(S\subseteq R\), so the
failed transformed inequality is exactly the original closure violation.
Empty position ranges contribute no members. At \(m=0\) the empty core passes.

The implementation performs \(m(m+1)/2\) suffix-value evaluations,
\(m(m-1)/2\) suffix comparisons, \(m(m+3)/2\) regional comparisons,
and \(m\) prefix additions. Order, prefixes, decisions and one compact
pair per failing size use \(O(m)\) elements. Expanding one witness costs
\(O(m)\); materialising all witnesses may require \(O(m^2)\) output.
Sorting adds \(O(m\log m)\) comparison work. These bounds exclude
exact-rational bit complexity and are not elapsed-time guarantees.

\paragraph{A separating family and its counterfamily.}
Let \(m=2^k\geq4\), \(h^*=m/2+1\), one value be
\(1/[\alpha(h^*+1)]\), and all others be \(m/\alpha\).
For \(h<h^*\), add the weak claim to a strongest core; its singleton
intersection fails because
\(\alpha|R|e_{\rm weak}=(h+1)/(h^*+1)<1\).
For \(h\geq h^*\), weak-only intersections pass. An intersection with
\(z\geq1\) strong values in the worst-report reduction satisfies
\(\alpha|R|\sum_{i\in S}e_i\geq |R|zm\geq|S|^2\), since
\(|S|\leq|R|\leq m\). Thus the minimum is exactly \(h^*\).
After its failed midpoint, retained binary search evaluates
\(\Theta(\log m)\) positive core sizes with both \(h\) and \(m-h\)
proportional to \(m\). Each fully evaluates
\((h+1)(m-h+1)-1\) regional inequalities, yielding
\(\Theta(m^2\log m)\) regional work versus the quadratic frontier.
This separates the specified algorithms, not every possible algorithm.
With the same construction but \(h^*=1\), positive binary queries shrink
geometrically and their regional work sums to \(O(m^2)\); a full frontier
need not help an easy single query.

\subsection{Executed checks and native qualification costs}
\label{sup:edit-region-evidence}

Exact checks used Python~3.14.6 and rational arithmetic. The fixed grid uses
\(m=0,\ldots,5\), every vector in \(\{0,1,40\}^m\), and
\(\alpha\in\{1/20,1/2\}\). The regional experiment examined 132,860
inputs, including every assignment to \(M/O/X\): 8,552 qualified and
124,308 failed. All 398,580 decisions across three complete methods agreed
with literal set enumeration; all 372,924 negative witnesses violated the
defining qualification inequality. The minimum-core experiment examined 18,662 fixed-\(U\)
optimisation problems and checked 37,324 binary/linear outputs against
literal minimum cardinality.

The subsequent \(U=H\) frontier experiment covered 728 snapshot/level
problems, 4,010 size decisions, 18,662 arbitrary core regions and 506,310
defining inequalities. All 3,856 negative compact witnesses expanded
correctly; all 728 minima agreed with the retained solver. Two fresh
executions of each mathematical program produced byte-identical reports.
These denominators overlap and are not independent samples or external
replication. The general results rest on the proofs, with finite tests
checking their implementations. Core construction followed the initial
regional results; the frontier followed a negative native amortisation
result. Both were identified as exploratory extensions before their own runs.

The native study used four complete providers: frontier (F), retained
region (R), memoised sweep (S), and memoised warm-binary (B). All share the
scientific inputs and decisions. Twelve functional families across four
providers, two backends and two fresh processes yielded 192 combinations:
176 non-race outputs paired exactly and 16 races with one legitimate durable
successor each. Timing used the AIJIM native backend only: 12 sequences,
137 steps per provider/process, four providers and two serial fresh
processes yielded 1,096 timed native publication attempts. The 137-step
collection includes four intentionally unavailable responses; an attempt
is not necessarily a successful commit. Provider order rotated by sequence
and was reversed in run~2. An interrupted first timing attempt, excluded
because startup overlap with a short analysis self-test could not be ruled
out, remains separately preserved and unscored.

Qualification cost includes registry setup and query work, including
construction, conversion and storage, failed construction, fallback and
cache handling. Full endpoint time additionally includes database work,
evidence changes, synthetic fresh approvals and native consumption/commit.
Warmup/imports, final audit and reporting are excluded.
Table~\ref{tab:sup-frontier-costs} retains every qualification workload.

\begin{SpecHeroTable}
\caption{Complete qualification cost in milliseconds per synthetic sequence.
Each cell gives run~1/run~2; construction is already included.
Display values are rounded from retained nanoseconds.}
\label{tab:sup-frontier-costs}
\begin{tabularx}{\linewidth}{@{}Yrrrr@{}}
\toprule
\textbf{Sequence} & \textbf{F} & \textbf{S} & \textbf{B} & \textbf{R} \\
\midrule
Unique short & 1.056/1.068 & 0.121/0.146 & 0.171/0.141 & 1.643/1.621 \\
Unique medium & 1.290/1.375 & 0.701/0.638 & 0.644/0.888 & 1.954/2.312 \\
Unique long & 2.231/2.465 & 2.389/2.840 & 2.603/3.441 & 2.744/3.404 \\
Repeated & 1.789/1.926 & 0.759/0.746 & 0.815/0.971 & 2.544/2.262 \\
Evidence churn & 9.090/9.383 & 1.257/1.735 & 1.099/1.373 & 8.137/10.772 \\
Region exits & 1.292/2.391 & 0.428/0.636 & 0.543/0.512 & 2.542/3.299 \\
Availability changes & 1.273/1.406 & 0.445/0.370 & 0.394/0.593 & 2.014/2.236 \\
Level changes & 2.442/4.522 & 0.708/0.998 & 0.805/0.748 & 3.892/3.606 \\
Hard, \(m=32\) & 1.511/2.478 & 1.089/1.030 & 1.277/1.145 & 3.972/3.481 \\
Easy, \(m=32\) & 1.438/1.627 & 0.649/0.898 & 0.736/0.940 & 1.917/4.297 \\
Hard, \(m=128\) & 15.136/15.917 & 2.424/2.742 & 2.777/2.650 & 39.039/55.109 \\
Easy, \(m=128\) & 14.418/15.079 & 1.163/2.211 & 1.363/3.546 & 22.881/25.172 \\
\bottomrule
\end{tabularx}
\end{SpecHeroTable}

Frontier qualification costs less than both memo controls on the 32-distinct-
selection sequence, but more on every other sequence in both runs. It
improves on retained region in ten sequences; evidence churn and level changes
are mixed. Hard-\(m=128\) construction falls from 38.247/53.773~ms to
14.269/15.047~ms, but eight edits do not amortise it against memo controls.
Full endpoint time for the 32-selection sequence is slower than sweep in
both runs. These observations support workload-specific qualification-cost
benefit, not significance, causal endpoint acceleration or human effort savings.

\paragraph{Scope, prior art and reproduction levels.}
Xu et al.'s computational result finds a largest individually qualified
report in the arithmetic-mean closure
\citep[Section~10, Equation~(61)]{XuEtAl2026EClosure}; here the quantified
object is every report in an interval. The preceding logarithmic-factor
comparison concerns our retained binary/regional implementation, not Xu's
algorithm merely because its stated complexity has the same order.

Goeman's different local e-collection admits nonempty reports of one
cardinality \citep[Section~3, Equation~(9); Section~6, Theorem~19]{Goeman2026ClosedBH}.
Its adjusted-p-value algorithm uses sliding-window minima maintained by a
deque, followed by reverse cumulative minima
\citep[Section~7; Appendix~B, Algorithm~2]{Goeman2026ClosedBH}.
This is relevant computational prior art; its output and collection differ
from the retained-core frontier studied here.

Ramdas's anchored e-BH uses
\(E_S=(\sum_{i\in S}e_i+m-|S|)/m\), rather than our unchanged
arithmetic-mean collection
\citep[Section~7, Equation~(39)]{Ramdas2026MaximalEFDR}.
The ranked-exchange argument has a close precedent in Proposition~A.1;
Theorem~A.2 gives a quadratic scan under its common-order and profile
conditions \citep[Appendices~A.1--A.2]{Ramdas2026MaximalEFDR}.
Candidate-prefix feasibility there need not be monotone. Our monotonicity
instead follows from shrinking the entire interval as its strongest core
grows. Amortisation over statistical query paths also has precedents
\citep{Durand2025FastJointError}.

The specific construction is exact interval qualification, its minimum
retained cardinality, and a shared all-core frontier under the stated
collection. The difference in the quantified object identifies the
contribution; it does not establish global priority or new error-control
theory.
A complete alternative can use the provider. Its region is bound to fixed
evidence, hypothesis meanings, universe and level; arbitrary changes to
model-generated text are not merely subset edits. Changed inputs require
reassessment. The output is a trusted local mathematical result, not an
authenticated remote proof or publication permit. Fresh human authorisation
and all native obligations remain necessary. A necessary correction can
remove a mandatory claim; this triggers reassessment, not suppression.

These new materials are archived in a separate version-1.0.0
collection~\citep{tiltack2026aijimExecutableEvidence}, not in the
published v0.1.0 or v0.2.0 DOI packages. Under
\nolinkurl{research/experiments/closure-edit-region-v1/},
the labelled scientific projections \nolinkurl{CONTRACT.md} (including
exploratory Amendment~A) and \nolinkurl{FRONTIER_CONTRACT.md} retain the
derivations; the unchanged implementations
are \nolinkurl{region_check.py}, \nolinkurl{minimum_core.py}, and
\nolinkurl{core_frontier.py}. The mathematical frontier source has SHA-256
\nolinkurl{a8f4c39e79e3a1a7f93b116a4d83e6b69a4efc72d6bc5bc5415e692c4d6c748a};
its two reports share
\nolinkurl{f7f5f5d1f5cd286f1b35eb9f4066066376aca86c1e8e0f5c16b313365488d951}.
The historical native execution freeze is
\nolinkurl{e68a734fe8af12816554397292e0a45563e47948b8c29f50c4454708f91e7c90};
\nolinkurl{frontier_native_comparison.json} is
\nolinkurl{789df4b0bc74ac3e8f70aef278068a87fa2c513caf8fb93d5e09a94d584e04f8}.

The package-root \nolinkurl{REPRODUCTION.md} gives the supported commands
for mathematical checks and archived native reanalysis, using fresh outputs
outside the package. Scientific fields are compared with retained reports;
documentary metadata need not reproduce the historical report bytes.
The original prospective documents remain local, with exact and derivative
identities distinguished in \nolinkurl{SOURCE_IDENTITIES.json}.
Raw phase costs and the full endpoint table remain in the labelled
\nolinkurl{FRONTIER_NATIVE_RESULTS.md} projection
and the bound \nolinkurl{frontier_cost_run1/results.json} and
\nolinkurl{frontier_cost_run2/results.json} files. These operations do not
regenerate or rescore the frozen SF-4 evaluation.
 \section{Executable Comparison of Three Supplied Protection Families}
\label{sup:comparative-protections}

These diagnostics examine how three existing SF-4 protections behave at the
publication boundary and how a constructed profile reproduces them.
The comparator comprises in-toto 3.0.0, OPA 1.19.1, and a transactional SQLite
publisher with explicitly authored publication semantics. It is not an
unmodified third-party publication protocol. The SF-4 arm uses the unchanged
archived \path{0.4.0-sf.4} reference binding, not the separately versioned
commitment-correction candidate.

\subsection{Design, authority, and observation points}

All studies were author-led and machine-assisted. Codex and Claude Code
supported source analysis, research-code development, and internal review;
the assistance did not constitute an independent evaluator.
Implementation inspection
preceded scenario selection; code-aware hypotheses and expected outcomes were
frozen with the experiment sources before execution. This is prospective
execution of targeted diagnostics, not blind selection, a random sample of
deployments, or independent external evaluation. Each complete matrix was
executed twice from fresh processes and SQLite state without intervening
source repairs. Repetition checks repeatability, not additional independent
case coverage. The input and temporal readback auditors were written after execution.
These audits and the internal machine-assisted review are quality checks,
not independent predeclared scientific oracles or a second human evaluation.

The studies retain cumulative repairs: the temporal baseline includes the
state-domain guard, and the required-run baseline includes both state and
proof-time guards. Every new guard executes in the actual OPA admission and
consumption policy; its bytes enter the bound policy identity. No old
safeguard is removed. Native signatures and approvals are freshly formed over
the perturbed objects, so the contrasts do not depend on stale digests or
deliberately broken signatures. Equivalent authoritative facts are available
to both arms, although their native encodings differ. Expected verdicts are
used for scoring, not supplied to the checkers.

A task-matching refusal requires no durable publication and a retained native
reason. Schema or authentication refusal is not relabelled as a substantive
checker \texttt{FAIL}. Unexpected exceptions remain apparatus failures.
Native decisions, native in-toto verification, signed inputs, and stored
SQLite payloads are reconciled rather than treating a runner label or policy
decision as a commit. All three studies completed without apparatus errors
or failed database integrity checks.

\subsection{Measurement-state admission: nine cases}

Let I0 denote the existing V2 comparator and I1 its state-domain repair.
The required measurement states are \texttt{MEASURED},
\texttt{NOT\_MEASURED}, and \texttt{CANNOT\_VERIFY}; the latter two require
their corresponding typed witnesses.

\begin{SpecHeroTable}
\caption{Input-domain diagnostic. Entries are durable publication counts
within each listed group, per execution.}
\label{tab:sup-input-contract}
\begin{tabularx}{\textwidth}{@{}YP{0.10\textwidth}P{0.10\textwidth}P{0.10\textwidth}@{}}
\toprule
\textbf{Group} & \textbf{SF-4} & \textbf{I0} & \textbf{I1} \\
\midrule
Three valid declared-state controls (3 cases) & 3 & 3 & 3 \\
Missing absence/unverifiability witness (2 cases) & 0 & 0 & 0 \\
State missing, null, numeric 7, or \texttt{UNKNOWN} (4 cases) & 0 & 4 & 0 \\
\bottomrule
\end{tabularx}
\end{SpecHeroTable}

Agreement with the fixed task was 9/9, 5/9, and 9/9 for SF-4, I0, and I1.
The four malformed variants are representations of one uncovered fault class,
not four independent discoveries. SF-4's valid fixture contains all three
states, so its three positive executions repeat that complete fixture; the
comparator's single measurement varies across the three states.

SF-4 stops at \path{pac2026_schema_instance_invalid} at
\path{$.measurements[1].state}, before substantive M1 evaluation.
I0 verifies the in-toto chain, returns OPA acceptance, and stores the actual
malformed object. I1 evaluates those same object bytes but returns
\path{MEASUREMENT_STATE_MISSING_OR_INVALID:measurement.state}; it stores
nothing. The original omission was a missing domain check before
state-specific branches. This demonstrates the value of supplying that
admission contract, not invention of schemas, an OPA language defect, or a
resolution of HC-0024. Measurement-byte availability and truth were not tested.

\subsection{Two validity intervals: five cases}

T0 retains I1; T1 adds only
\(\mathit{proofIssuedAt}\leq t_v<\mathit{proofExpiresAt}\).
Existing grant activation/end checks remain. Both arms receive the same
declared proof and publication intervals. In the comparator, proof-time
fields belong to the actual signed evaluation artifact and its approval-bound
digest; the existing grant fields continue to denote publication validity.

\begin{SpecHeroTable}
\caption{Temporal diagnostic. Each entry is a durable publication count.}
\label{tab:sup-temporal-contract}
\begin{tabularx}{\textwidth}{@{}YP{0.10\textwidth}P{0.10\textwidth}P{0.10\textwidth}@{}}
\toprule
\textbf{Case} & \textbf{SF-4} & \textbf{T0} & \textbf{T1} \\
\midrule
Proof valid at verification, expired by publication & 1 & 1 & 1 \\
Proof expires exactly at verification & 0 & 1 & 0 \\
Proof expires one nanosecond after verification & 1 & 1 & 1 \\
Publication before grant activation & 0 & 0 & 0 \\
Publication exactly at grant end & 0 & 0 & 0 \\
\bottomrule
\end{tabularx}
\end{SpecHeroTable}

Task agreement was 5/5, 4/5, and 5/5, with respectively two, three, and two
durable records. The central difference is one temporal fault class.
SF-4's original-proof refusal is
\path{pac2026_sf4_proof_expired}, before permit issuance. The fixture invokes
the native private A1 classifier; only its specific
\path{_Pac2026Sf4A1Error} expiry outcome is normalized. This is authentication
refusal, not an admitted A1 \texttt{FAIL}. Publication-boundary refusals occur
at serialized consumption with
\path{pac2026_sf4_publication_permit_not_current}.
T1 returns \path{PROOF_NOT_VALID_AT_VERIFICATION:temporal}; retained interval
rules diagnose not-yet-valid or expired publication grants.

Post-run verification checked all 20 in-toto chains and all 10 original
SF-4 proof signatures across the two executions, including authentic but
expired proofs. The useful distinction is between validity at verification
and a separately bounded later publication; neither signature authenticity
nor current publication permission alone supplies it. These are trusted
synthetic times and public fixture identities, not a measurement of external
clock truth or real human authorization. SF-4 additionally restricts
\(\mathit{verificationTime}\leq\mathit{validFrom}\); this source-inspected
restriction was not a sixth tested control or an availability advantage.

\subsection{Required-run identity: seven cases}

Comparator arm B retains both previous repairs. C adds a nonempty,
well-shaped run-array guard, but no particular run identity. D adds exact
agreement between expected and observed run identities.

The task independently fixes \path{run-1} for the admitted candidate.
SF-4 obtains this expectation from its authenticated PolicySnapshot.
Every comparator receives the same owner-signed in-toto task envelope,
separately admitted task digest, and candidate binding. Thus the producer
cannot redefine what is required by changing what it supplies.
Three apparatus controls verified valid task admission, rejection of altered
contents under the old signature, and rejection of a re-signed but unadmitted
task. They are not extra scientific comparison cases.

\begin{SpecHeroTable}
\caption{Required-run diagnostic. Each entry is a durable publication count;
the state and time controls verify that earlier repairs remain active.}
\label{tab:sup-required-run}
\begin{tabularx}{\textwidth}{@{}YP{0.08\textwidth}P{0.08\textwidth}P{0.08\textwidth}P{0.08\textwidth}@{}}
\toprule
\textbf{Case} & \textbf{SF-4} & \textbf{B} & \textbf{C} & \textbf{D} \\
\midrule
Expected run, correct candidate & 1 & 1 & 1 & 1 \\
Expected run, wrong candidate & 0 & 0 & 0 & 0 \\
Required entry omitted & 0 & 1 & 0 & 0 \\
Entire run field omitted & 0 & 1 & 0 & 0 \\
Well-shaped, correct-target \path{run-other} substituted & 0 & 1 & 1 & 0 \\
Measurement state absent & 0 & 0 & 0 & 0 \\
Proof expires at verification & 0 & 0 & 0 & 0 \\
\bottomrule
\end{tabularx}
\end{SpecHeroTable}

Task agreement was 7/7, 4/7, 6/7, and 7/7. Omitting the sole run erases its
former target: deletion from a correct-target and wrong-target input yields
identical objects. Those duplicates were checked and executed once per
distinct omission form. The result does not claim to detect an erased target
or constitute a 28-field admission census.

The wrong-target SF-4 control yields an admitted R1 \texttt{FAIL},
\path{R_ARTIFACT_TARGET_MISMATCH}, and prevents permit issuance.
For an empty list or substituted identity, the native checker proposes
\path{R_REQUIRED_RUN_MISSING}, but authenticated-result admission stops with
\path{pac2026_policy_observation_domain_incomplete}. The completed substantive
outcome is therefore \texttt{UNASSESSED}, not admitted R1 \texttt{FAIL}.
A missing field instead fails checker-input admission with
\path{pac2026_schema_instance_invalid}. No later consumption is pretended.

The shape repair returns \path{RUN_PRODUCER_SHAPE_INVALID} for the two
omissions but admits the well-shaped substitute. The identity repair returns
\path{EXPECTED_RUN_MISSING:run-1} and
\path{UNEXPECTED_RUN_PRESENT:run-other}. Native readback confirms that B/C's
substitution publications actually contain the supplied substitute while
retaining the signed task requiring \path{run-1}. This distinguishes correct
shape and target from fulfillment of an independently admitted task identity.
A signed requirement does not prove that the real-world task was complete
or scientifically sound.

\subsection{Retained comparison and mechanism evidence}

The preceding V2 locked comparison yielded 21 expected outcomes and two
unassessed undeclared-origin T8 cases per arm, in each of two executions.
Its 46 case/arm observations reproduced decisions, native codes, record
counts, lifecycle states, and normalized effects. Neither the unassessed
cases nor repeated observations become additional passes. Both arms handled
the assessed candidate, binding, freshness, replay, witness, availability,
and lifecycle contrasts. Later lifecycle tests assess a supplied prefix;
they do not demonstrate transactional append of later events.

A separate retained V1 mechanism experiment jointly removed the
\path{resultSetDigest} comparison at its two existing SF-4 enforcement sites.
Original/ablated/restored configurations each allowed a valid single
publication (one record), allowed two matching packages (two records), and
refused a stale authorization after candidate change (zero). Only the crossed
record/permit changed: zero/one/zero records. The byte-identical restoration
and second fresh execution reproduced all 12 configuration/case outcomes.
This identifies the tested relationship's causal role under joint ablation;
it does not establish each site's individual sufficiency or exclusive
capability of the architecture.

\subsection{Artifact identities, repeatability, and inference}

The complete original contracts, source freezes, inputs, native traces,
databases, observation manifests, and post-run audits remain locally retained
under \path{research/experiments/}. The historical records are identified by:
\begin{itemize}
\item \path{publication-input-contract-v1}: \path{CONTRACT.md},
\path{RESULTS.md}, \path{run-01/}, \path{run-02/}, and
\path{reproduction.json};
\item \path{publication-temporal-contract-v1}: the same contract/result/run
structure and its post-run audit;
\item \path{admission-silence-census-v1}: \path{CONTRACT.md},
\path{RESULTS.md}, \path{freeze-01.json}, both run directories,
\path{apparatus-01/summary.json}, and \path{AUDIT.json};
\item \path{publication-authority-comparison-v2}: \path{RESULTS.md},
\path{confirmation-freeze.json}, and \path{results/confirmation-01/},
\path{results/confirmation-02/}, \path{results/reproduction.json};
\item \path{publication-authority-comparison-v1}: the exact ablation patch,
restoration manifest, and \path{results/ablation-01/} /
\path{results/ablation-02/}.
\end{itemize}

The archived collection~\citep{tiltack2026aijimExecutableEvidence} distributes
unchanged computational sources and raw
observations, with contracts and reports as labelled scientific projections
and freeze manifests as \path{identity/*.scientific.json} identity projections,
each recording the original SHA-256. \path{SOURCE_IDENTITIES.json}
distinguishes original and distributed identities. The three original
diagnostic freeze SHA-256 values, in input/time/run order, are:
\begin{quote}\small
\path{979d25b6bd7dd5e269ba327d9f44db87e4b50c6d64bc16051d744da35d531fdf}\\
\path{60fd32f4d6a2cc7c149c823e4580bba1e4fae1cae405797b07439835c6f0aa0d}\\
\path{bc2898906471c1a4963208ba1d91f23534786b1103a5a898fd94c5ff27412ea6}
\end{quote}
Their paired observation files are byte-identical within each study, with
SHA-256 values respectively:
\begin{quote}\small
\path{ee75c4c9689374f11ff72189534476d66818db7e50e1022324dfaabf5eb05b76}\\
\path{093cb92ab6e02063552ec76b0ce73e9106e16f4785ddfa4e51c282853f82d1f7}\\
\path{d853c143f4dfd69a9ee87e1f132353783a3841052439ff89b3e3893cc45a1024}
\end{quote}
The input, temporal, and run studies retain respectively 27, 15, and 28
case/arm observations per execution, with 501, 292, and 937 manifested raw
files per run. These denominators are not pooled into a comparative score.
All protected source closures remained unchanged during the executions.
The required-run freeze records Python 3.14.6, cryptography 50.0.0 for
SF-4 and 50.0.2 for the comparator, securesystemslib 1.5.1, OPA 1.19.1,
and SQLite 3.53.4. Each study freezes its exact executable identities.
The package-root \path{REPRODUCTION.md} defines supported integrity checks,
mathematical checks, archived reanalysis, and a separate diagnostic-child
replay via \path{reproduction/native_replay.py}. Its macOS arm64 profile in
\path{reproduction/PROFILE.md} requires separate pinned Python environments
and an external OPA executable. Selected native dependencies and required
authority metadata are bundled unchanged; these child paths need no older
archive download. The metadata records historical authority decisions, not
a new review or independent evaluation. The fixed smoke selection compares retained
decisions, native codes, and commit counts and checks package integrity.
Full historical parent-runner/documentary-freeze replay is not claimed.
Original receipts retain their historical host identities; the unchanged
v0.1.0/v0.2.0 DOI deposits do not contain these later diagnostics.

The results establish three concrete supplied protections relative to the
named pre-repair profiles and executable repair parity on each respective
case set. The final cumulative profile was not re-run as an omnibus suite
over every earlier matrix. No lower implementation cost, field failure rate,
general safety guarantee, first invention of these design principles, or
superiority over the fully augmented alternative follows. SIP-RC, APP, and
other source-reviewed publication protocols were not native executable arms.
The comparison supplies mechanism-level design knowledge and inspectable
counterexamples; it does not replace independent external evaluation.
 \section{Separate-Team Implementation and Application Integration}
\label{sup:shark-integration}

Separately from this study, ViaoraAI S.L.'s engineering team independently develops the implementation from
the AIJIM specification and documentation. This development attribution is reported by the author;
the technical inventory below establishes selected implementation paths, not
the organisational independence of an evaluation. SHARK integrates the same
Core rather than providing a second semantic implementation.

The inventory was inspected on 18 October 2026 against Product source
\path{8efd123db08f1f9394419b1b23d85d5756e7511c}, Core development version
\path{0.5.0a4}, and the SF-5/RFC-021 line. The retained Source-B
(personal-authorization integration) and Source-C (publication and read-original
recovery) implementation/review packets identify their own earlier
implementation candidates. The dated preparation
receipts do not embed a source commit; they are not attributed retrospectively
to the inspected checkout. No new product test was executed for this inventory.

\begin{SpecHeroTable}
\caption{Reported SHARK integration scope. A retained observation is distinct
from an assertion found in test source and from a real user publication.}
\label{tab:sup-shark-integration}
\begin{tabularx}{\textwidth}{@{}P{0.18\textwidth}P{0.41\textwidth}Y@{}}
\toprule
\textbf{Operation} & \textbf{Implementation and retained support} &
\textbf{What this establishes} \\
\midrule
Candidate admission &
The API separates authenticated submission and read-original recovery.
The 17 October receipt records selected preparation checks. &
Admission preparation and lost-reply handling, not a completed real Stage-A
admission. \\
Personal authorization &
Source-B binds actor and candidate and requires separate personal
confirmation. A retained API log reports controlled local checks. &
Locally exercised authorization integration with synthetic identities, not
an actual human authorization or independent human evaluation. \\
Single-use publication &
Source-C connects preparation, publication, and read-original recovery.
SQL test source asserts one signer call, one commit, and read-only reopening. &
A connected Core-owned consumption path and explicit test assertions.
The specific SQL execution receipt was not established by this inventory. \\
Export preparation &
The 17 October retained-plan receipt covers selected retained-material,
mutation, and missing-witness cases. &
Selected export-preparation observations, explicitly without claiming
publication. \\
Offline verification &
The RFC-021 entry accepts separately selected trust and distribution inputs.
Test source distinguishes inner SF-5 \texttt{PASS} from outer
\texttt{CANNOT\_VERIFY} without distribution qualification. &
An implemented verification boundary; no complete, independently set-up
outer-package \texttt{PASS} is established here. \\
\bottomrule
\end{tabularx}
\end{SpecHeroTable}

The earlier public G6 observation is separate: on 9 September 2026, twelve
cases matched their expectations on each of the Core-native and
API-to-Core bindings (12/12 plus 12/12). It is neither twenty-four API cases
nor a test of this later RFC-021 inventory. The current integration therefore
adds a concrete application mapping of protocol operations; it does not
extend the frozen SF-4 evaluation, close G7, or combine selective receipts
into a completed production demonstration.

The \path{README.md} in the archived
collection~\citep{tiltack2026aijimExecutableEvidence} identifies
the included earlier public G6 readback and closure as implementation context.
Private Product source and local
receipts are not distributed as a publicly reproducible Product study.
Their identities and inspection status remain in the local research inventory;
the public G6 evidence also remains in the previously archived collection.
 
\bibliographystyle{elsarticle-num}
\bibliography{references}